\RequirePackage{fix-cm}
\documentclass[twocolumn]{svjour3public}          % twocolumn
\smartqed  % flush right qed marks, e.g. at end of proof
\usepackage{graphicx} % for pdf, bitmapped graphics files
\usepackage{epsfig} % for postscript graphics files
\usepackage{verbatim} % For the comment environment.
\usepackage{color}
\usepackage{soul}
\usepackage[small]{caption}
\usepackage{hyperref}
\usepackage{amssymb}
\usepackage{amsmath}
\usepackage[english]{babel}
\usepackage[nolist,nohyperlinks]{acronym}
\usepackage{epstopdf}
\usepackage{bm}  % bold math
\usepackage{hyphenat}
\usepackage{pdfpages}
\usepackage{subfig}
\usepackage{xspace}
\usepackage{mwe}
\usepackage{multirow,bigdelim}
\usepackage{mathtools}
\usepackage[linesnumbered]{algorithm2e} 
\usepackage{mathrsfs}
\usepackage{paralist}
\usepackage{float}

\usepackage{enumitem}
\newcommand*{\Scale}[2][4]{\scalebox{#1}{$#2$}}%

\makeatletter
\newcommand{\removelatexerror}{\let\@latex@error\@gobble}
\makeatother

\makeatletter
\newcommand*{\centerfloat}{%
  \parindent \z@
  \leftskip \z@ \@plus 1fil \@minus \textwidth
  \rightskip\leftskip
  \parfillskip \z@skip}
\makeatother

\newlength\algowd

\let\oldnl\nl% Store \nl in \oldnl
\newcommand{\nonl}{\renewcommand{\nl}{\let\nl\oldnl}}% Remove line number for one line

\newcommand{\codecomment}[1]{\,\,{\ttfamily\scriptsize // #1}}

\def\PerformancesHeight{4.3cm}
\def\PerformancesFormat{pdf}

\DeclareMathSymbol{\Gamma}{\mathord}{operators}{"00}
\DeclareMathSymbol{\Delta}{\mathord}{operators}{"01}
\DeclareMathSymbol{\Theta}{\mathord}{operators}{"02}
\DeclareMathSymbol{\Lambda}{\mathord}{operators}{"03}
\DeclareMathSymbol{\Xi}{\mathord}{operators}{"04}
\DeclareMathSymbol{\Pi}{\mathord}{operators}{"05}
\DeclareMathSymbol{\Sigma}{\mathord}{operators}{"06}
\DeclareMathSymbol{\Upsilon}{\mathord}{operators}{"07}
\DeclareMathSymbol{\Phi}{\mathord}{operators}{"08}
\DeclareMathSymbol{\Psi}{\mathord}{operators}{"09}
\DeclareMathSymbol{\Omega}{\mathord}{operators}{"0A}
\journalname{Autonomous Robots}
\begin{document}

\title{3D Multi-Robot Patrolling with a Two-Level Coordination Strategy %\thanks{Grants or other notes
}
%\subtitle{Simulations and Experiments with UGVs}


\author{Luigi Freda \and
        Mario Gianni \and
        Fiora Pirri \\
        Abel Gawel \and 
        Renaud Dub\'e \and 
        Roland Siegwart \and
        Cesar Cadena 
}


\institute{L. Freda, M. Gianni and F.Pirri \at
ALCOR Lab, DIAG - Sapienza University of Rome, Italy\\
              \email{freda,gianni,pirri@diag.uniroma1.it}           %  \\
           \and
           A. Gawel, R. Dub\'e, R. Siegwart and C. Cadena \at
           Autonomous Systems Lab - ETH Zurich, Switzerland \\                               		   \email{gawela, rdube, rsiegwart, cesarc@ethz.ch}          
%  \\           
}

\date{Received: \today / Accepted: date}

\maketitle

\begin{abstract}
%==================================================================
%==================================================================
%\input{abstract}

Teams of UGVs patrolling harsh and complex 3D environments can experience interference and spatial conflicts with one another. %
Neglecting the occurrence of these events crucially hinders both soundness and reliability of a patrolling process.
This work presents a distributed multi-robot patrolling technique, which uses a two-level coordination strategy to minimize and explicitly manage the occurrence of conflicts and interference. 
The first level guides the agents to single out exclusive target nodes on a topological map. This target selection relies on a shared idleness representation and a coordination mechanism preventing topological conflicts. 
The second level hosts coordination strategies based on a metric representation of space and is supported by a 3D SLAM system. Here, each robot path planner negotiates spatial conflicts by applying a multi-robot traversability function. 
Continuous interactions between these two levels ensure coordination and conflicts resolution. 
Both simulations and real-world experiments are presented to validate the performances of the proposed patrolling strategy in 3D environments. 
Results show this is a promising solution for managing spatial conflicts and preventing deadlocks.

%==================================================================
%==================================================================

\keywords{3D patrolling \and 3D multi-robot systems  \and distributed multi-robot coordination \and UGVs}
\end{abstract}

\section{Introduction}\label{Sect:Intro}
%==================================================================
%==================================================================
%\input{introduction}

\def\enableKeepAspect{1} % enable/disable figures (kept aspect ratio) 
\def\enableMosaicTxF{0} % enable/disable mosaic 2x4
\def\enableMosaicTxT{0} % enable/disable mosaic 2x2 (first page)

\newcommand{\gridimageheight}{2.85cm}
\newcommand{\gridimagewidth}{1.65in}

\if\enableKeepAspect1
\begin{figure*}[ht]
\centerfloat
%\resizebox{1.2\textwidth}{!}{
%\begin{center}
%
\subfloat
{
\includegraphics[height=\gridimageheight, keepaspectratio]{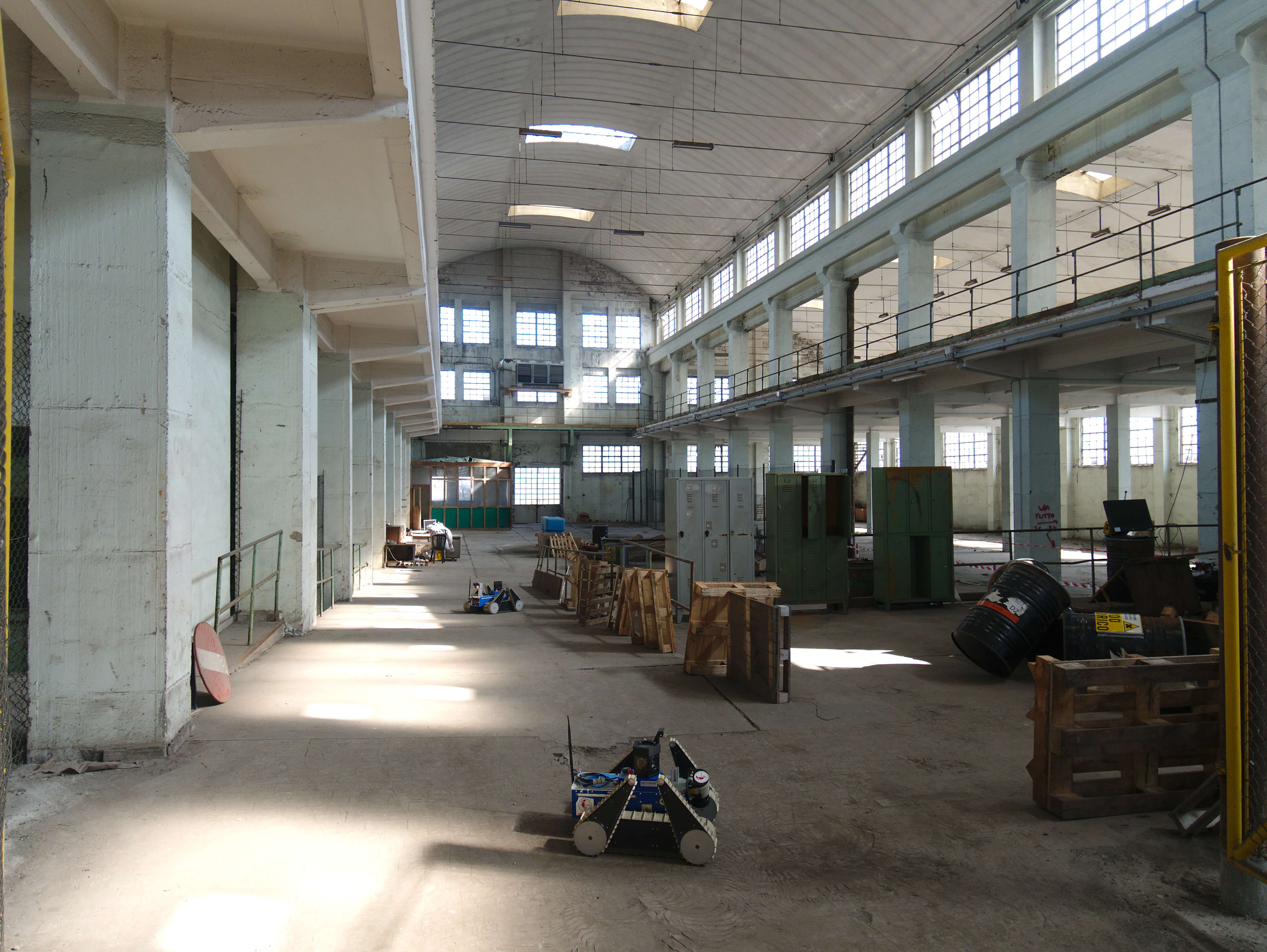}
\includegraphics[height=\gridimageheight,keepaspectratio]{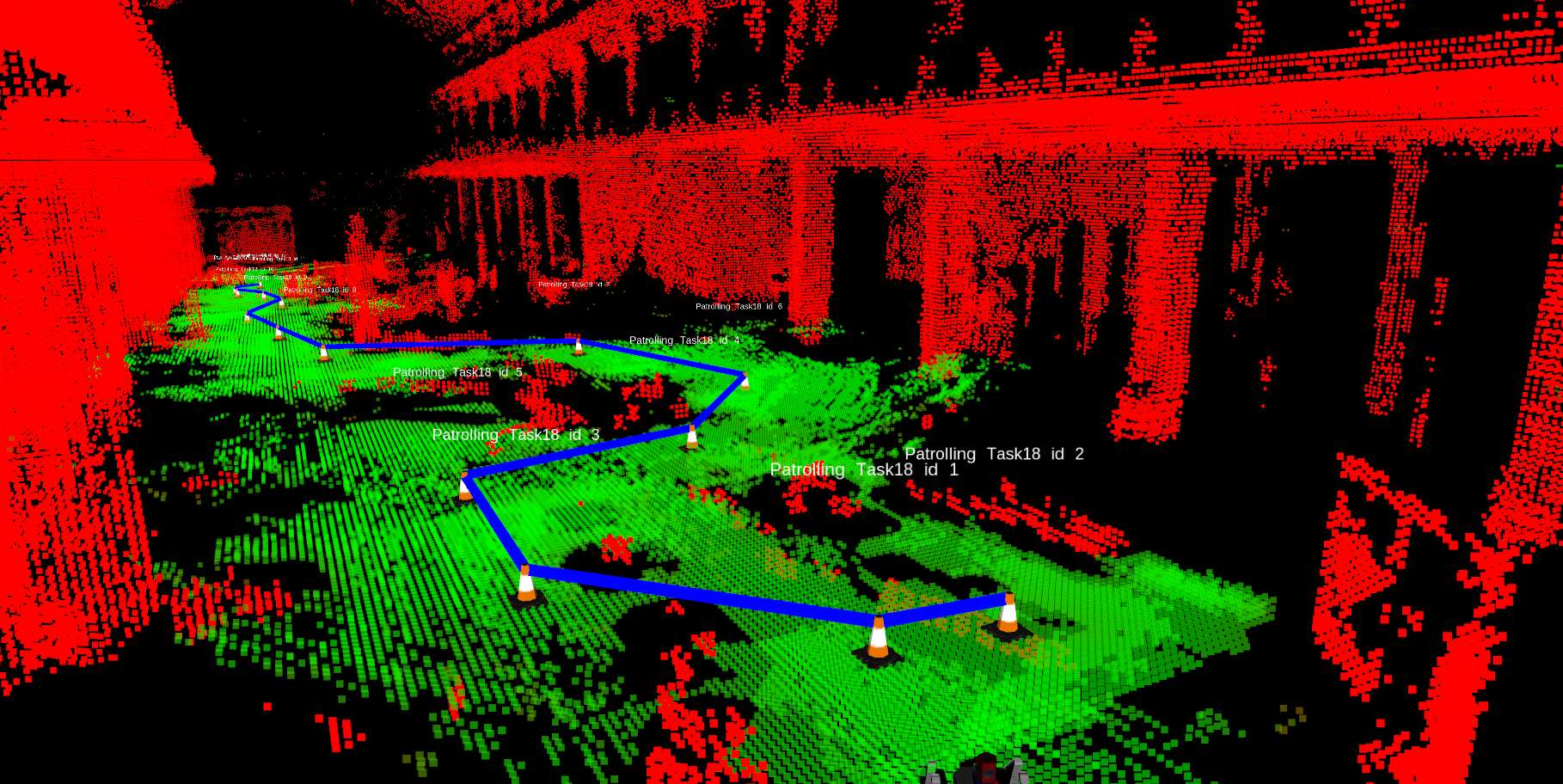}
\,\,
\includegraphics[height=\gridimageheight,keepaspectratio]{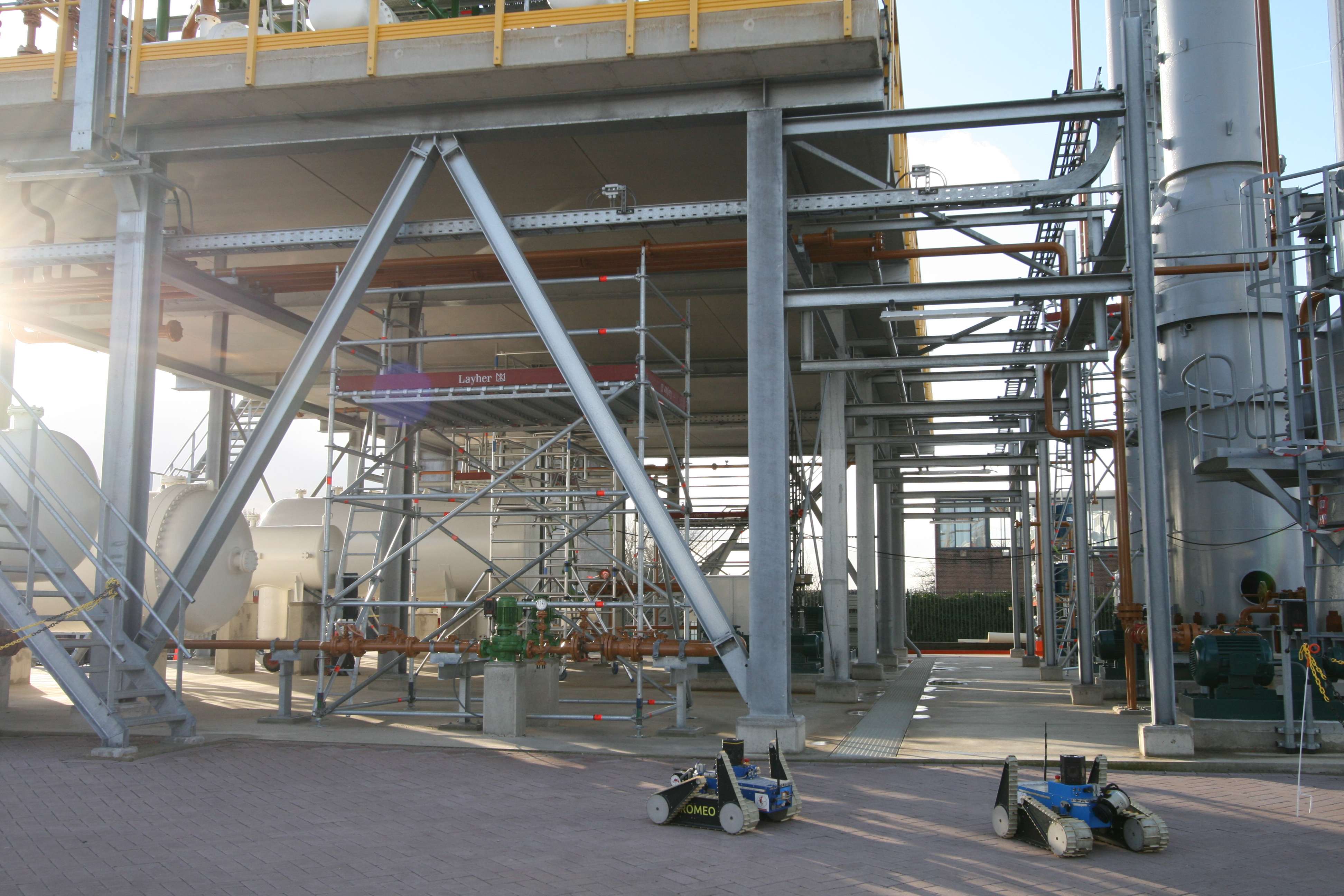}
\includegraphics[height=\gridimageheight,keepaspectratio]{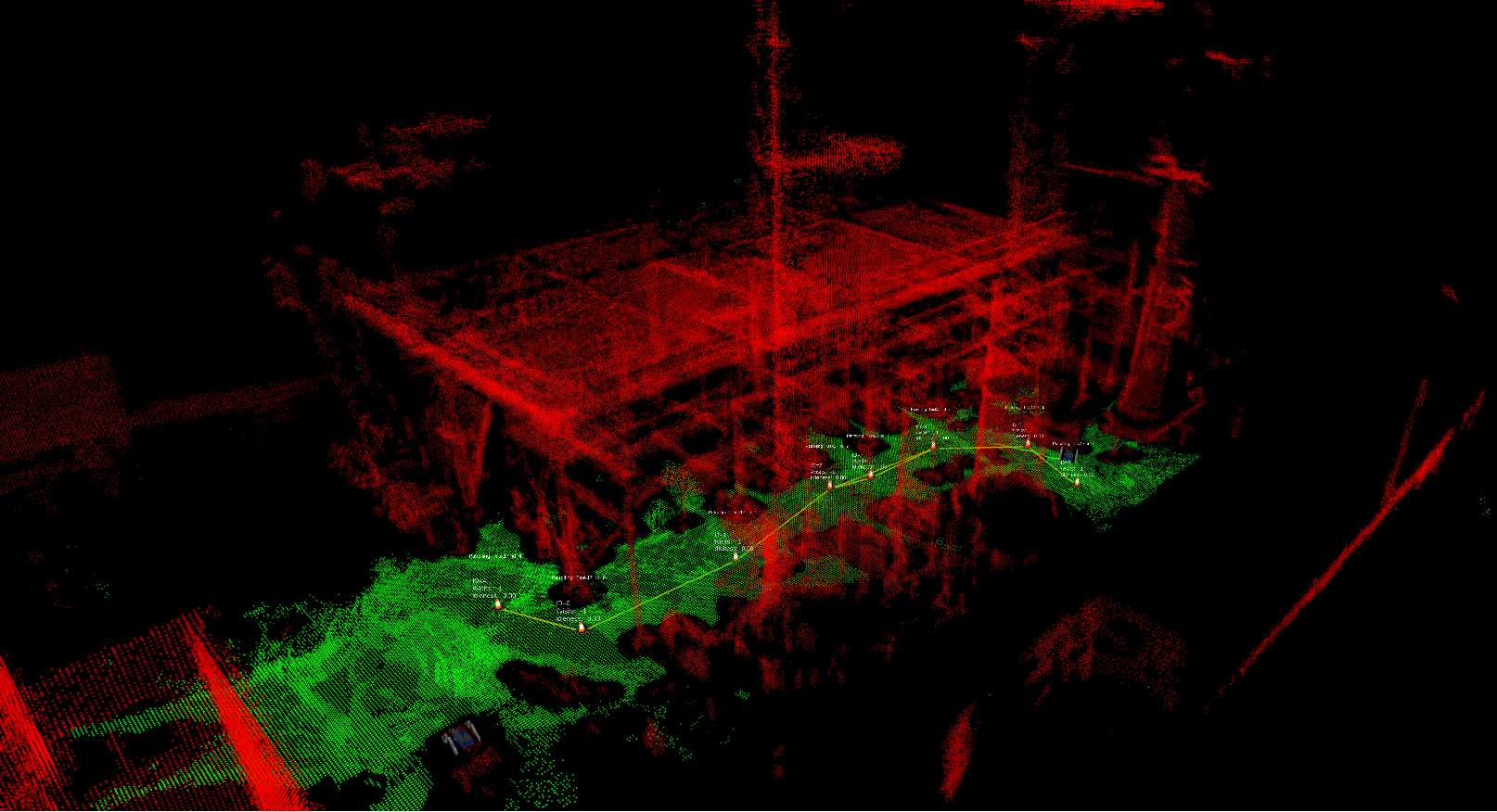}
}
\,
\subfloat
{
\includegraphics[height=\gridimageheight, keepaspectratio]{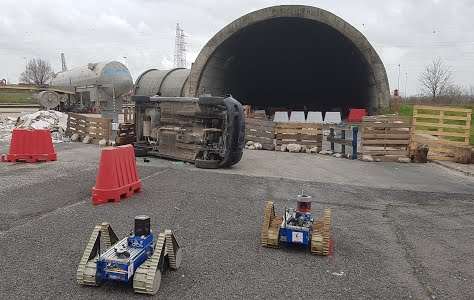}
\includegraphics[height=\gridimageheight, keepaspectratio]{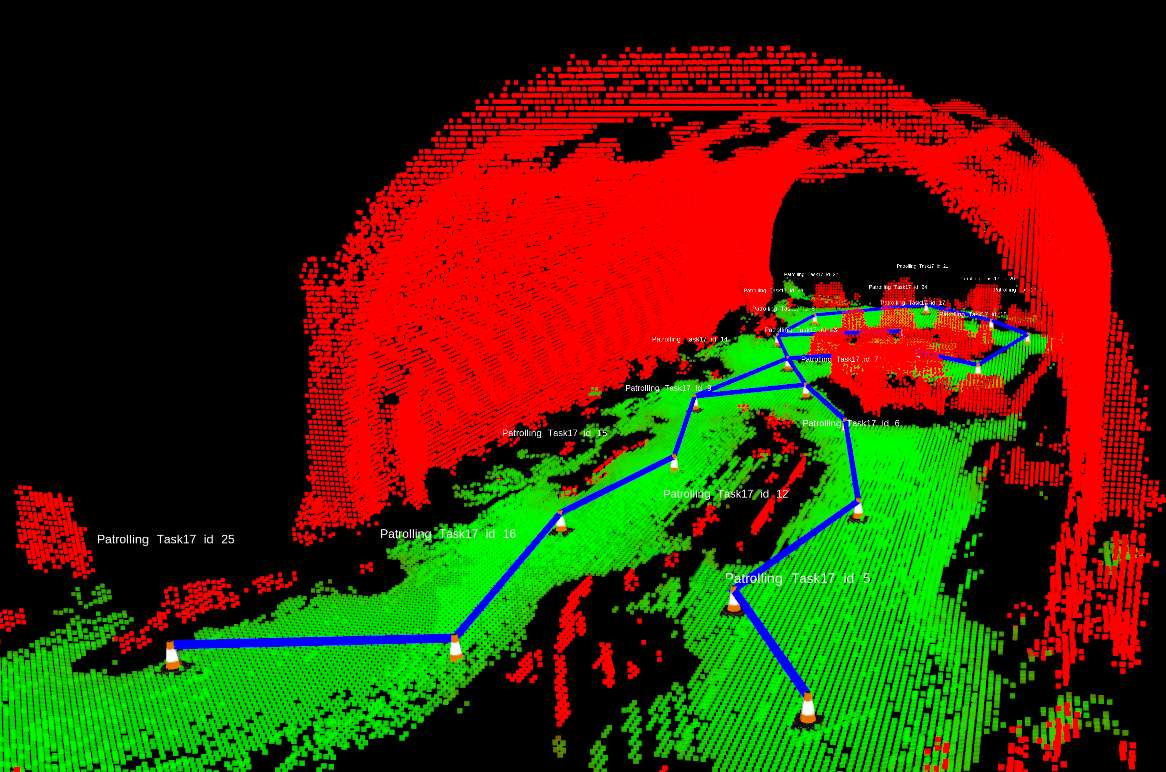}
\,\,
\includegraphics[height=\gridimageheight,keepaspectratio]{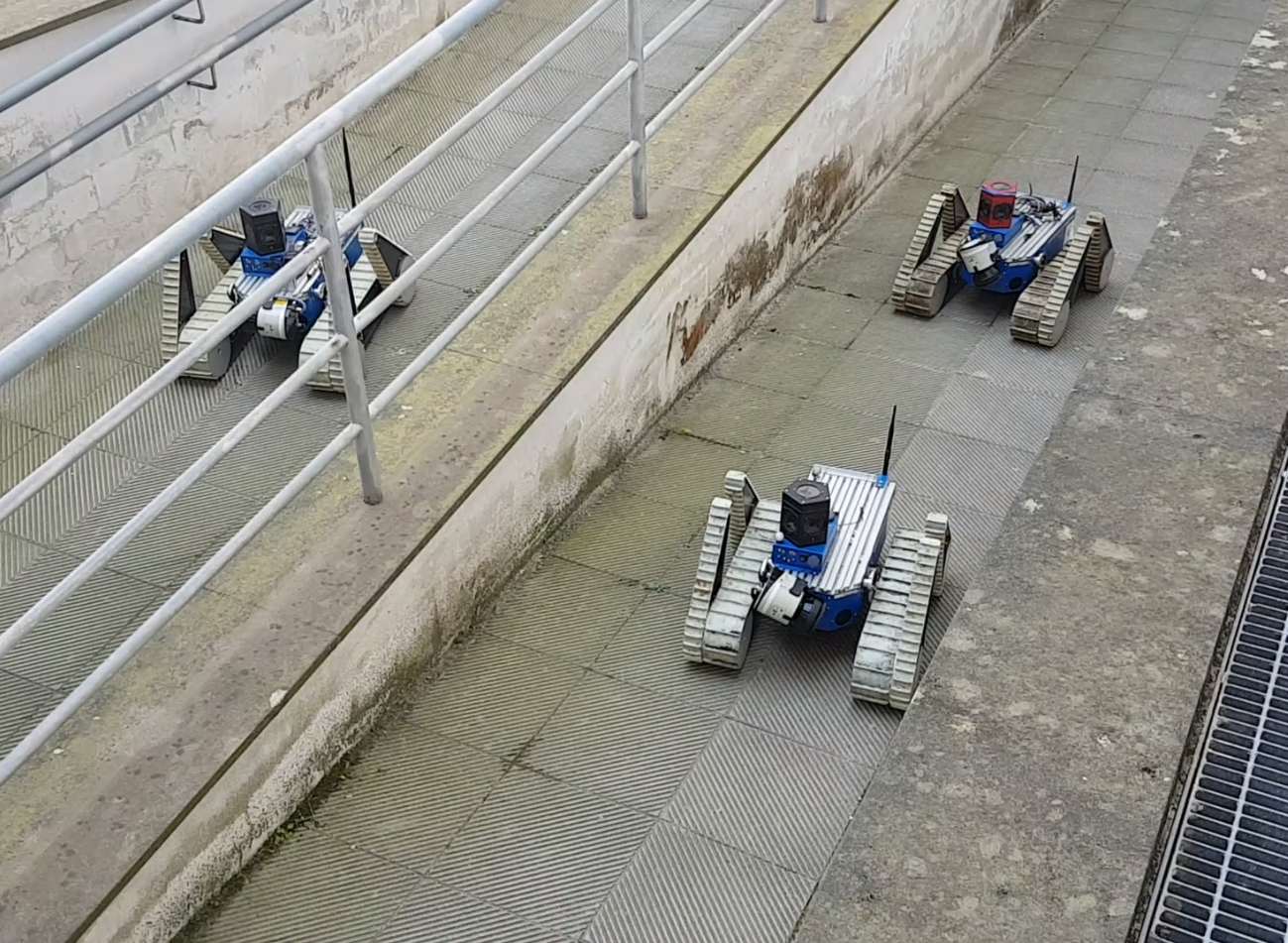}
\includegraphics[height=\gridimageheight,keepaspectratio]{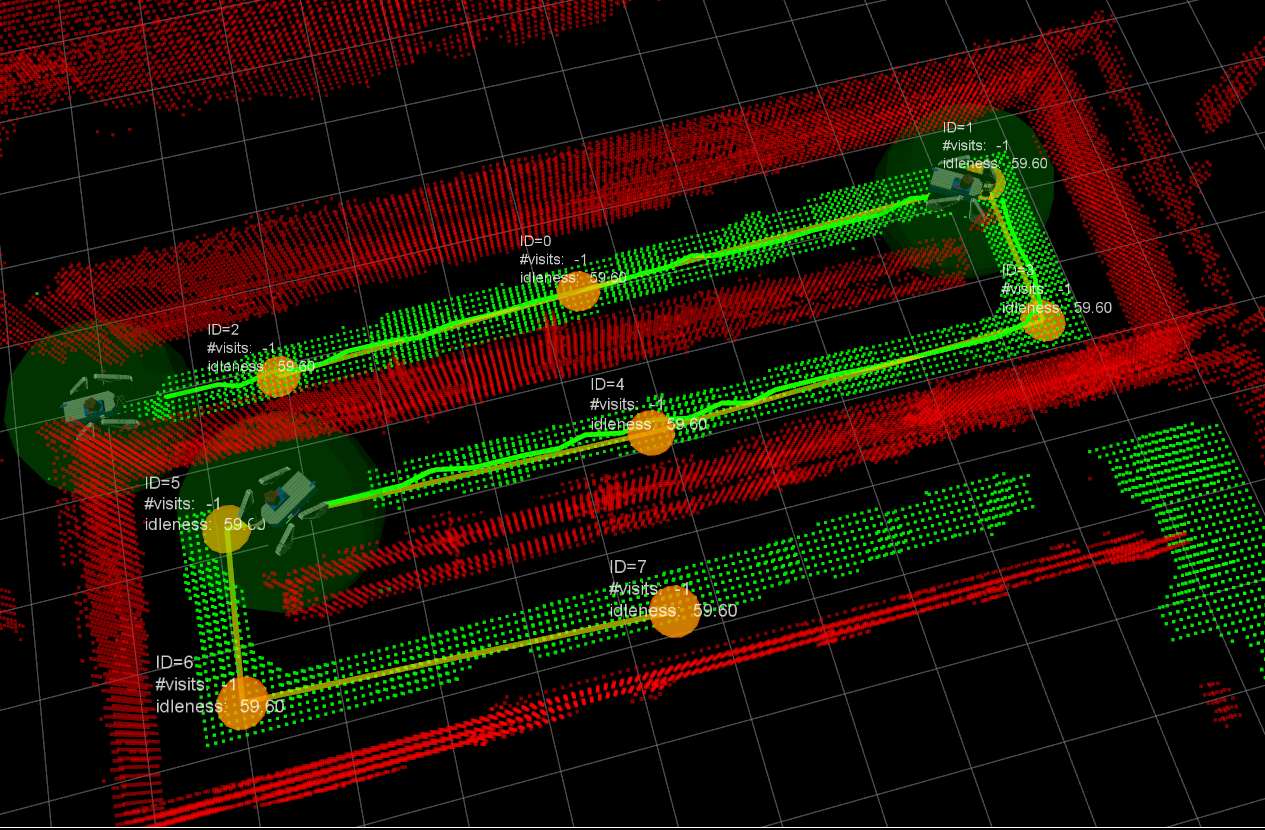}
}
%
%\end{center}
%}
\caption{Patrolling scenarios with their 3D maps and patrolling graphs. We refer the reader to the paper webpage \href{https://sites.google.com/a/dis.uniroma1.it/3d-cc-patrolling/}{ https://sites.google.com/a/dis.uniroma1.it/3d-cc-patrolling/} for videos and further details.}
\label{Fig:PatrollingScenarios}
\end{figure*}
\fi

\if\enableMosaicTxF1
\begin{figure*}[ht]
\centerfloat
\setlength{\tabcolsep}{0.2em}
\def\arraystretch{0.} % for lines spacing 
\begin{tabular}{cccc}
\subfloat{\includegraphics[width = \gridimagewidth, height=\gridimageheight]{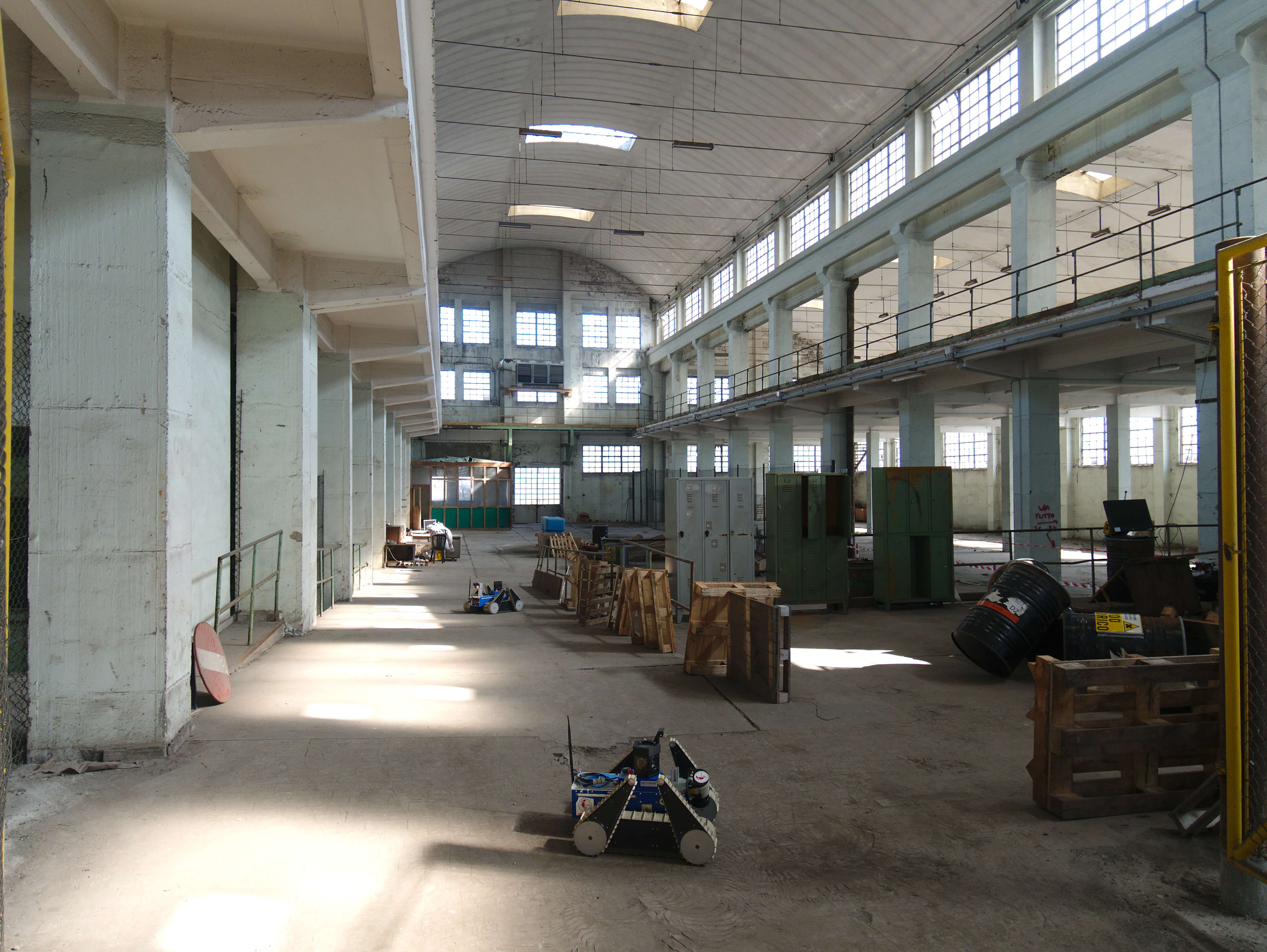}}
&
\subfloat{\includegraphics[width = \gridimagewidth, height=\gridimageheight]{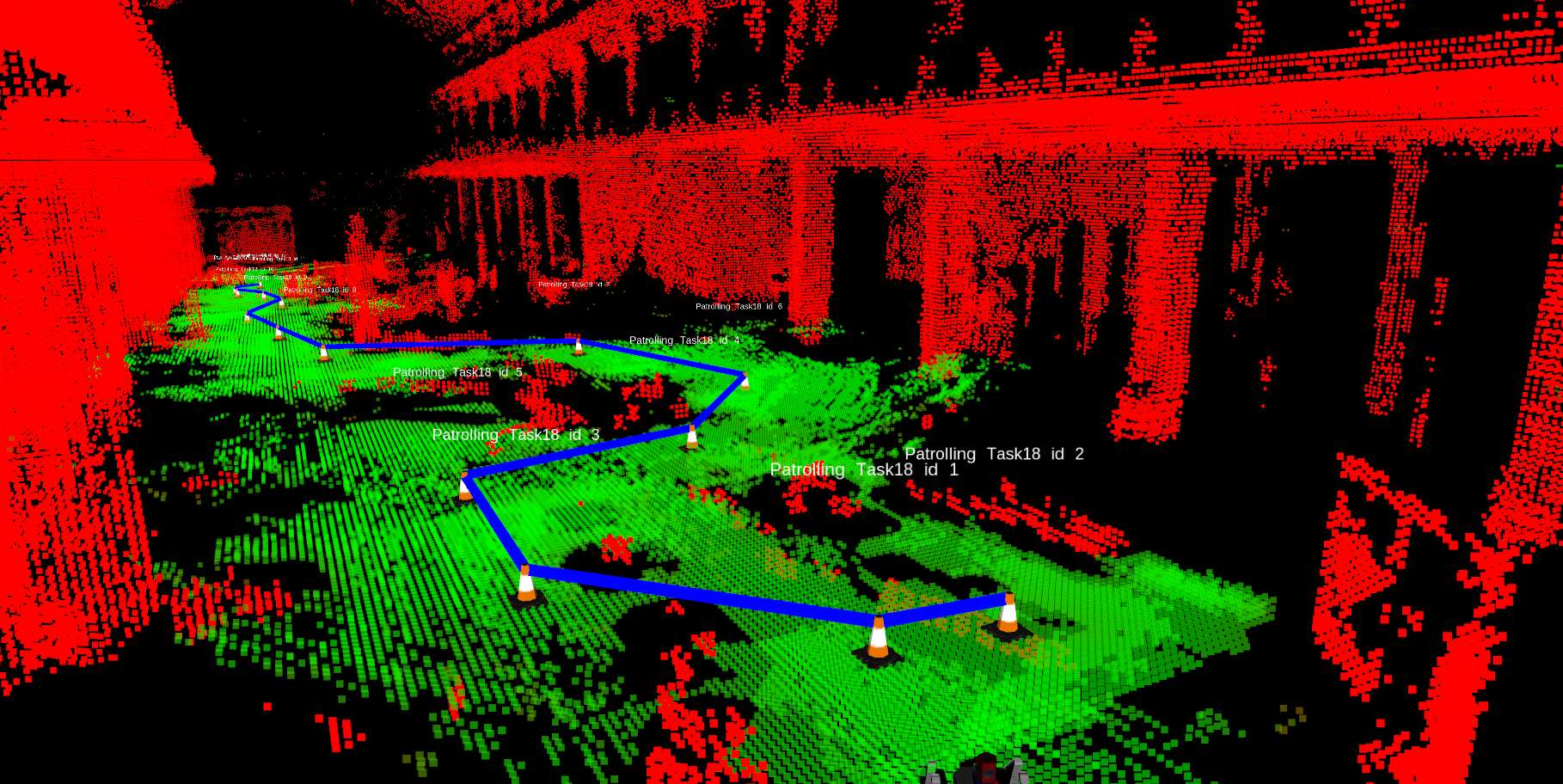}}
&
\,
\subfloat{\includegraphics[width = \gridimagewidth, height=\gridimageheight]{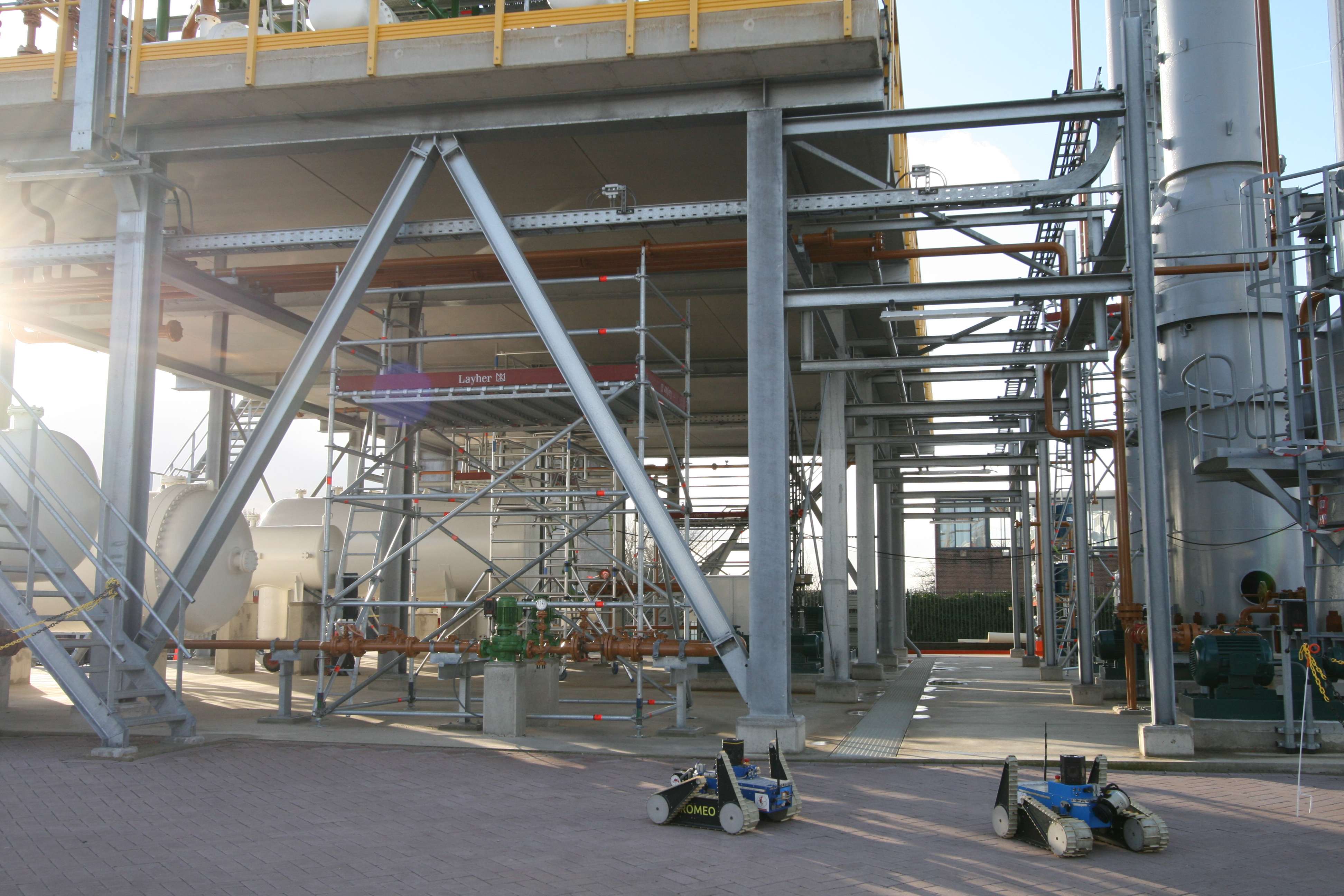}}
&
\subfloat{\includegraphics[width = \gridimagewidth, height=\gridimageheight]{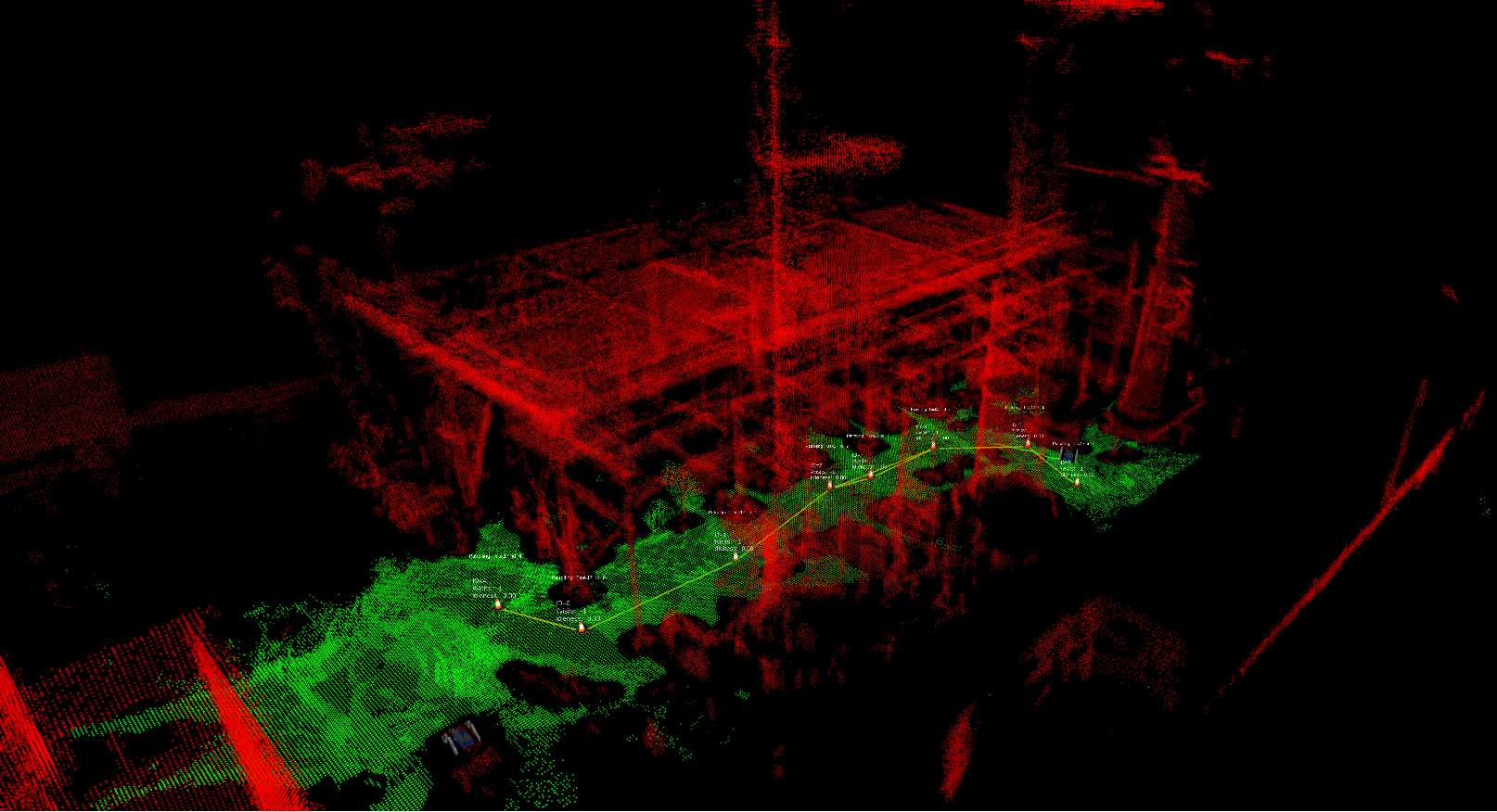}}
\\
\subfloat{\includegraphics[width = \gridimagewidth, height=\gridimageheight]{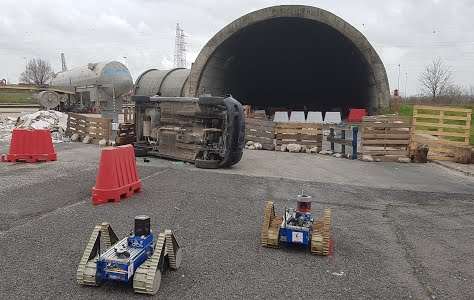}}
&
\subfloat{\includegraphics[width = \gridimagewidth, height=\gridimageheight]{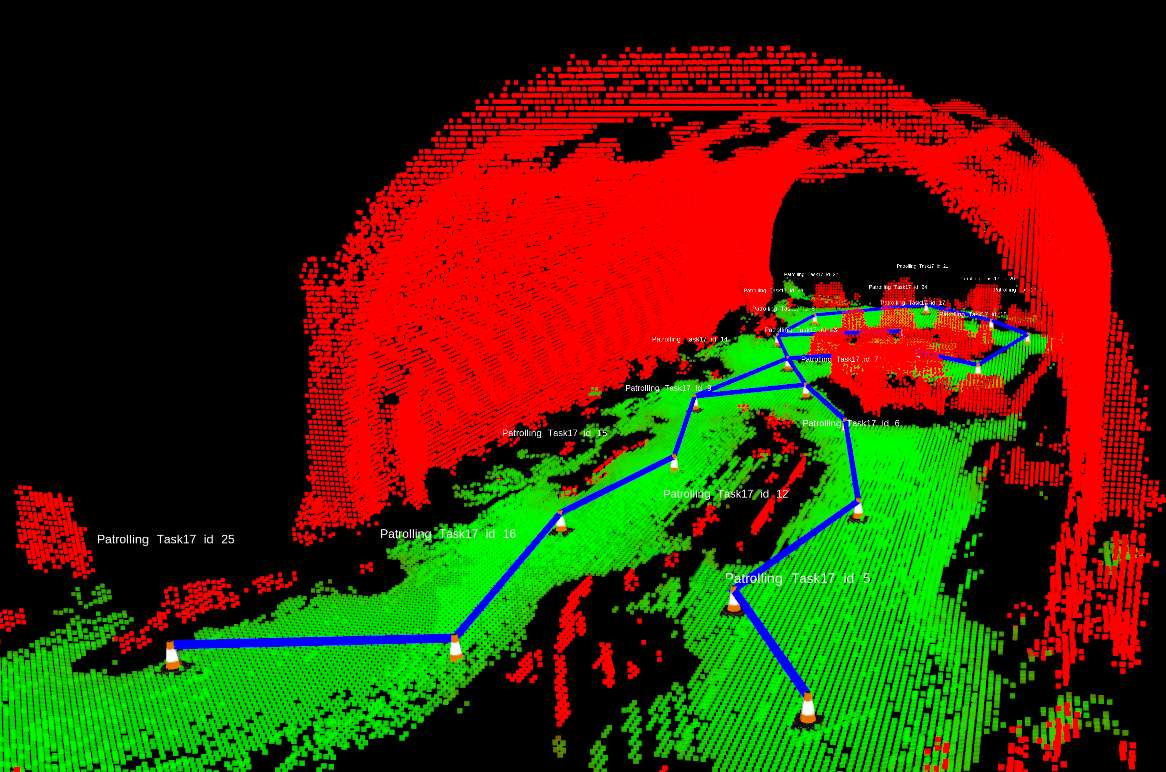}}
&
\,
\subfloat{\includegraphics[width = \gridimagewidth, height=\gridimageheight]{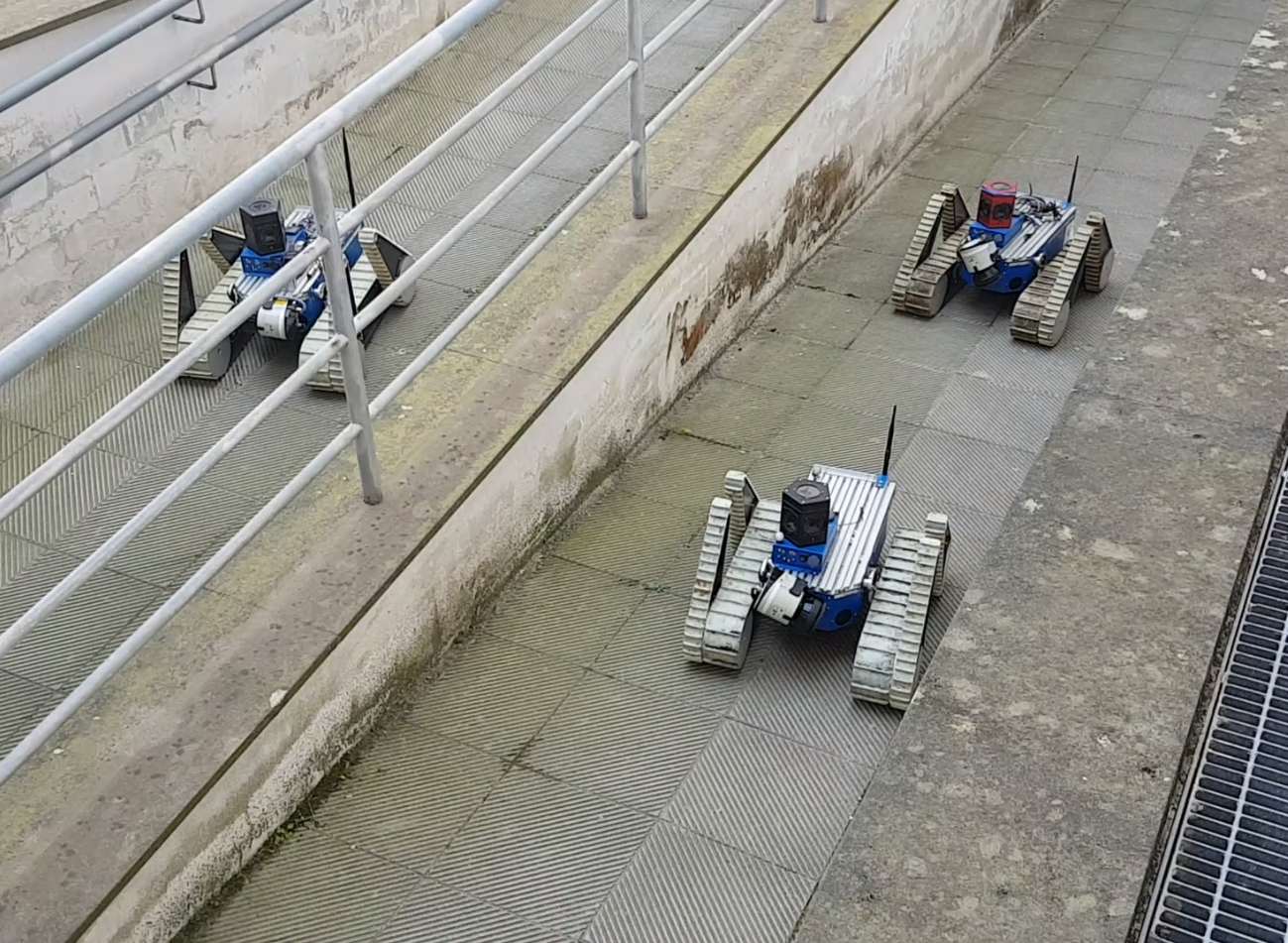}}
&
\subfloat{\includegraphics[width = \gridimagewidth, height=\gridimageheight]{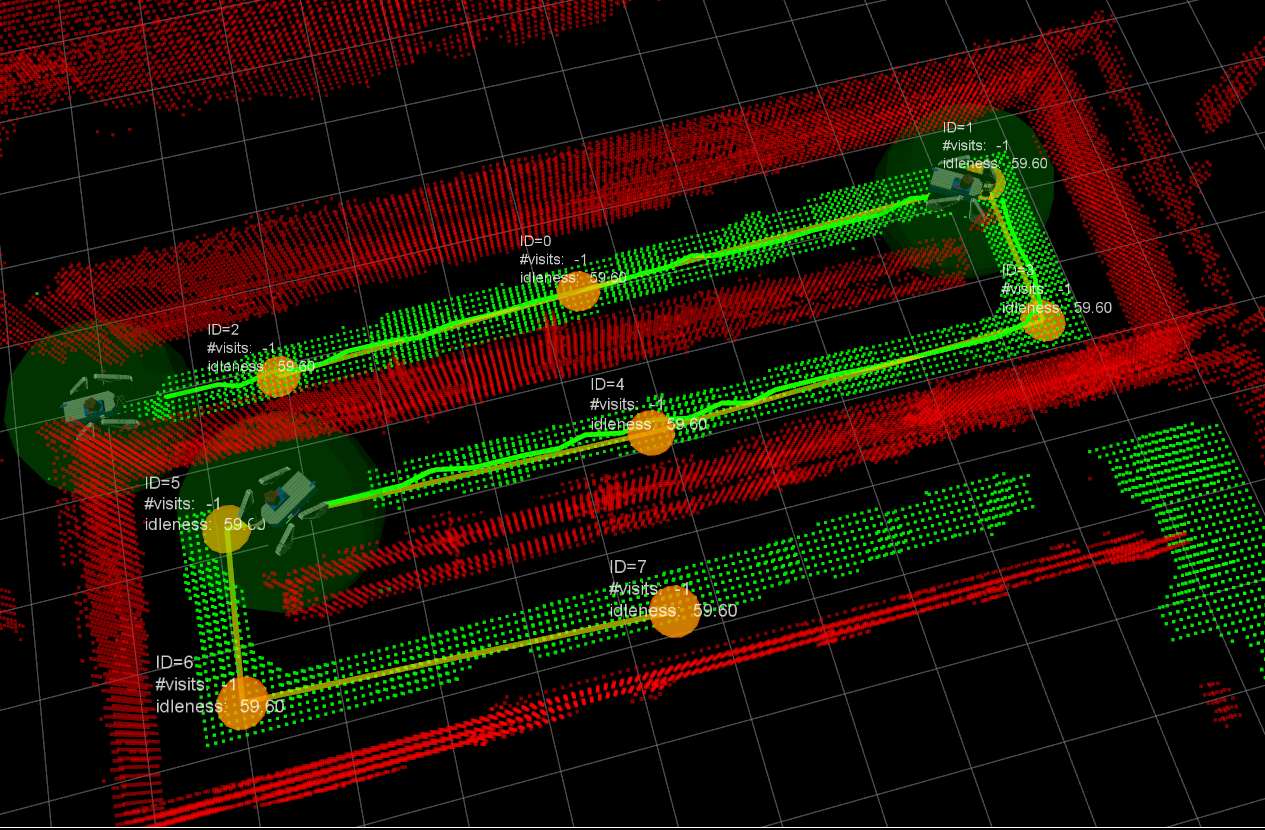}}
\end{tabular}
\caption{Some patrolling scenarios with their 3D maps and patrolling graphs.}
\label{Fig:ExperimentScenarios}
\end{figure*}
\fi

\if\enableMosaicTxT1
\begin{figure}
\centerfloat
\setlength{\tabcolsep}{0.2em}
\begin{tabular}{cc}
\subfloat{\includegraphics[width = \gridimagewidth, height=\gridimageheight]{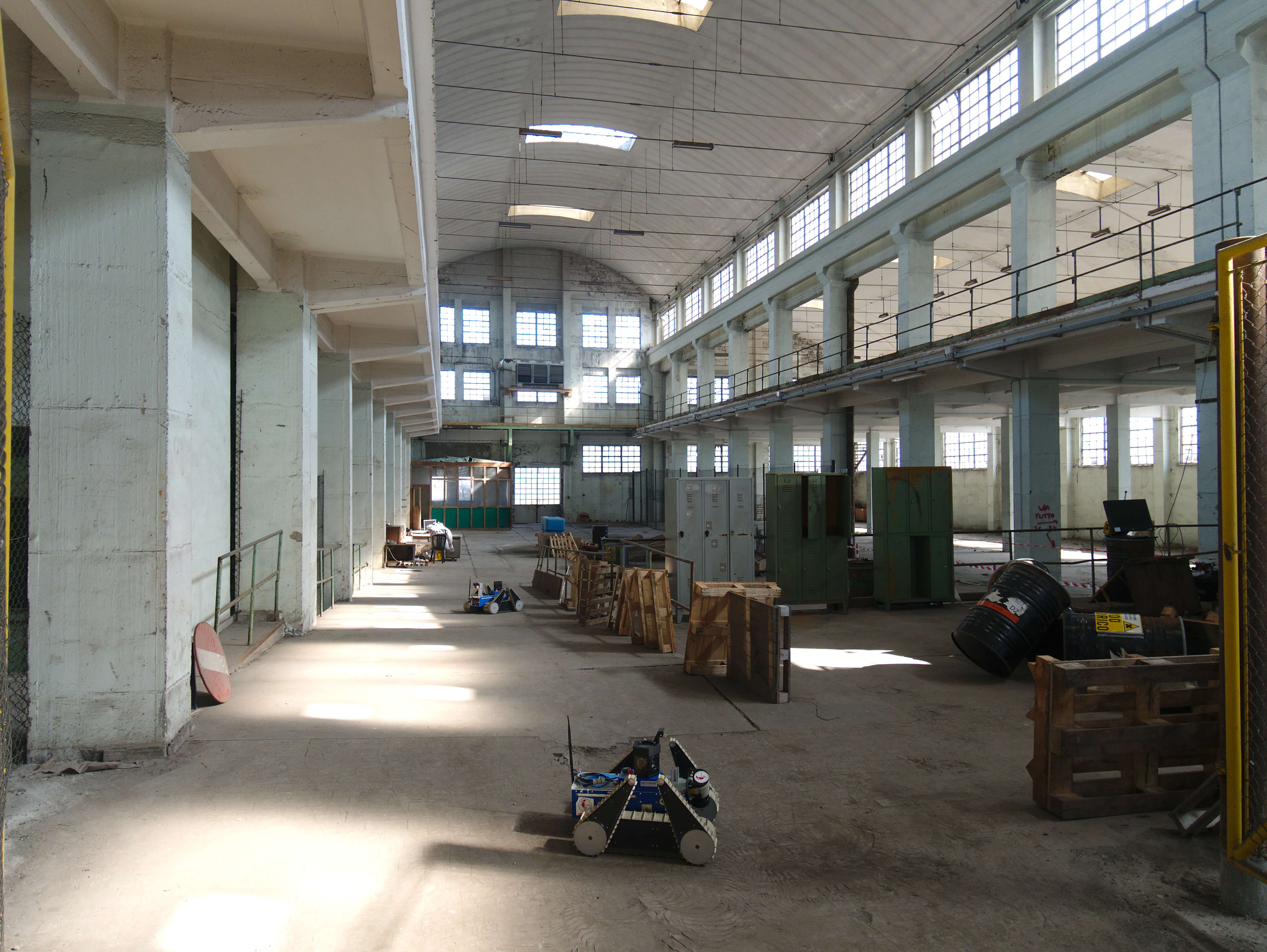}} &
\subfloat{\includegraphics[width = \gridimagewidth, height=\gridimageheight]{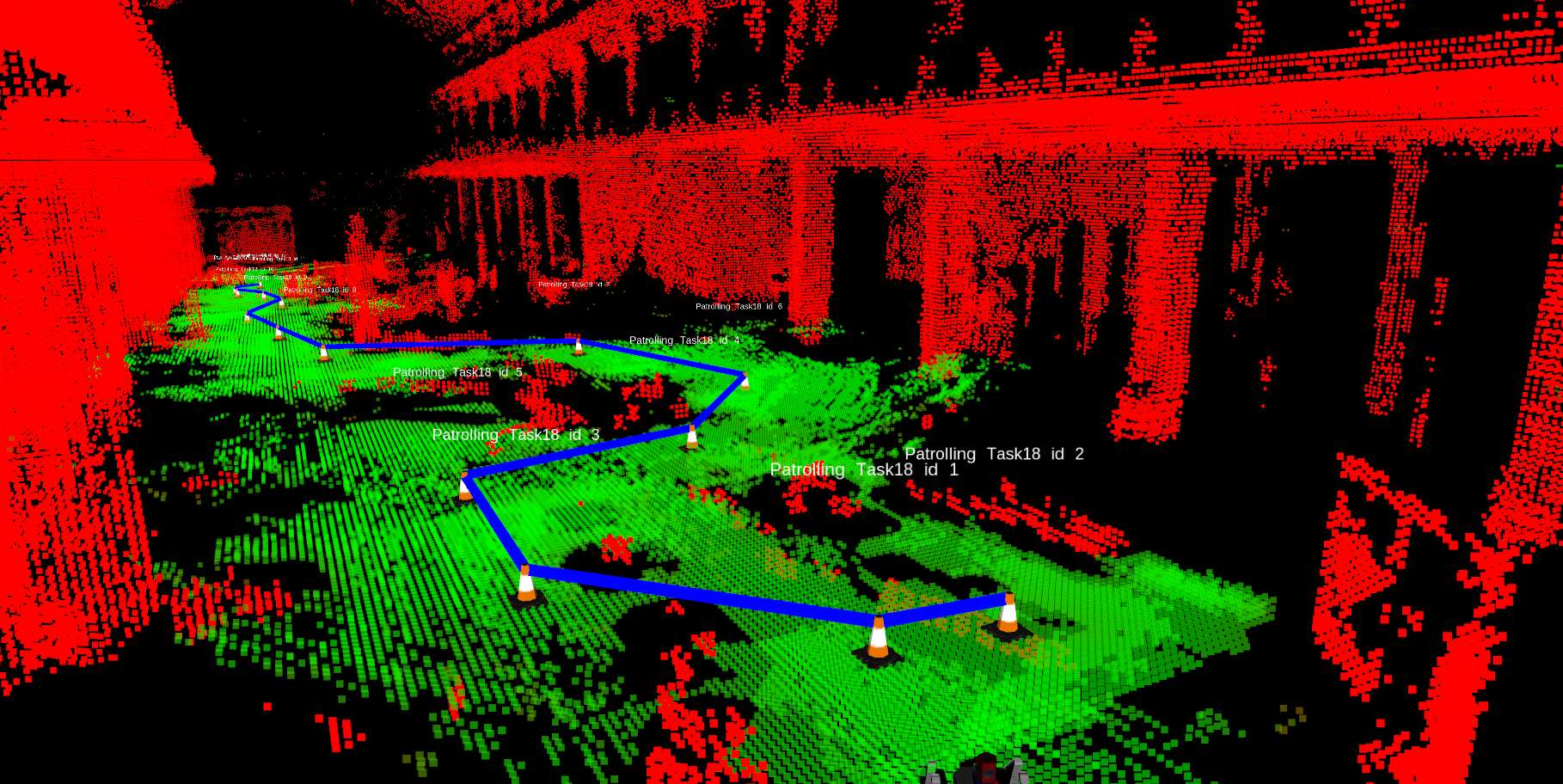}}\\
\subfloat{\includegraphics[width = \gridimagewidth, height=\gridimageheight]{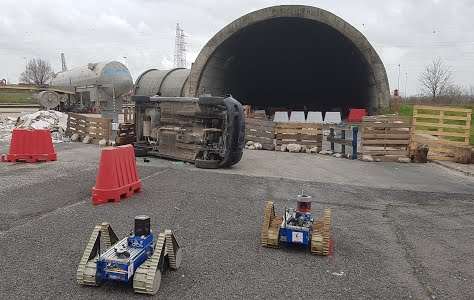}} &
\subfloat{\includegraphics[width = \gridimagewidth, height=\gridimageheight]{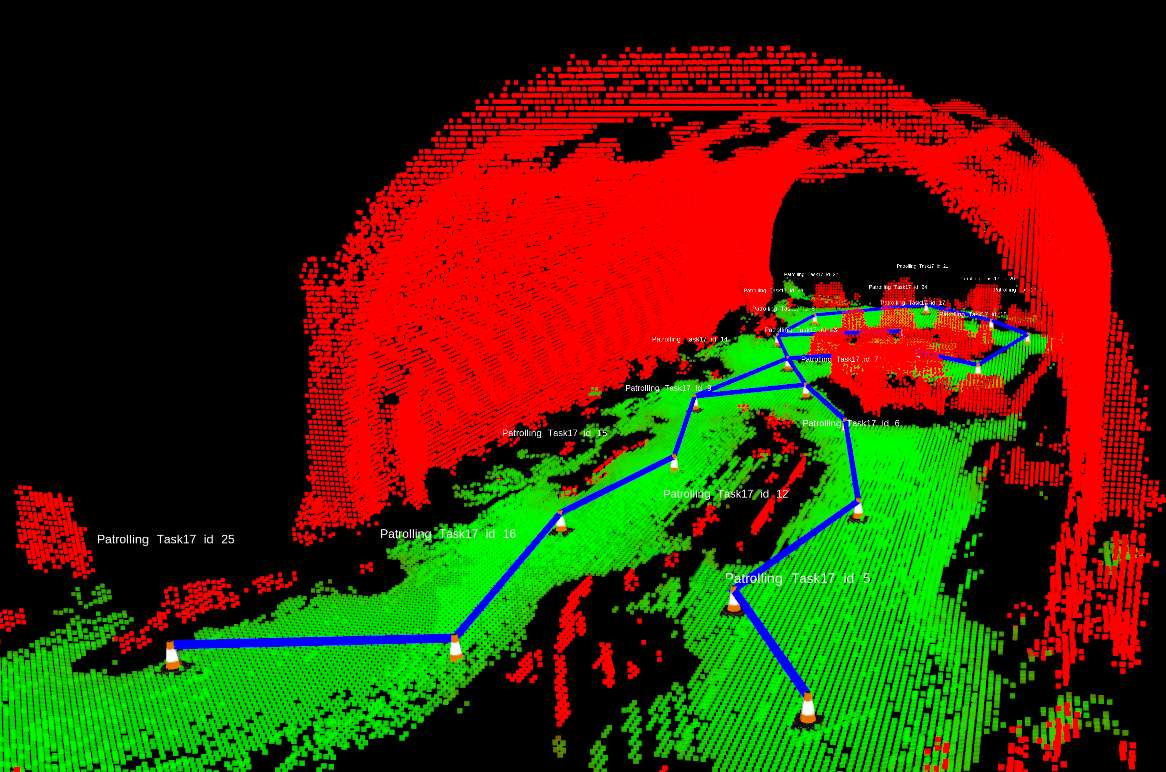}}
\end{tabular}
\caption{Some patrolling scenarios with their 3D maps and patrolling graphs.}
\label{Fig:ExperimentScenarios}
\end{figure}
\fi

Multi-robot patrolling  is   a relevant area of investigation in \ac{AI} and robotics since the early nineties (see \cite{portugal2011} for a survey and \cite{Portugal-2013a} for a study on strategies and algorithms). 
Still, the literature is limited to  abstract agents and robots that hardly can be operated in full 3D environments.

Multi-agents and multi-robot  patrolling methods have been largely treated in the literature for agents and robots operating in laboratory settings and in allegedly 2D flat environments. However, very little has been done so far when patrolling \emph{(i)} concerns real 3D world environments such as emergency or inspection scenarios, and \emph{(ii)} deals with complex robot structures such as Unmanned Ground Vehicles (UGV).
In this regard, the differences are substantial: first, the difficulties to be faced are substantially higher; second, in real scenarios, where professional operators (purportedly trained) act with extreme difficulties,  the problems and tasks that need to be addressed are driven by specific current needs and not by abstract strategies. 

This work addresses the multi-robot patrolling problem for UGVs operating in full  3D environments. 
We propose a strategy that minimizes and explicitly manages the occurrences of conflict and interference. These unwanted events can generate deadlocks and se\-vere\-ly impact a team of robots when patrolling narrow surroundings due to collapsed infrastructures or other wreckages obstructing passages.

Indeed, despite the fact that fully autonomous robots cannot be involved in human rescue so far, they can certainly assist a human team engaged in several difficult tasks.  
For example, robots are expected to lift the operators from the burden of assessing the state of the environment such as reachability of specific areas, footing of collapsed building, dangerous pipes, infrastructures and objects, and safe areas where the rescuers can possibly pass through in order to reach relevant objectives.
There is nowadays a wealth of literature on the tasks and roles a robot team can perform in order to reduce human risks under these circumstances (recent reviews can be found  in \cite{kleiner2016,jung2017}). An analysis of robots' potentials in reducing human risks during disaster response and their associated costs are treated in~\cite{Tardioli-2015}.

Immediate intervention of robot teams to the aftermath of tragic  events (see for example \cite{murphy2004,kruijff2012,nagatani2013,kruijff2014,kruijff2016} for a list of these episodes) 
requires urgent solutions and assessments in terms of communication, mapping and areas to be covered for information acquisition. In this context, response time is often a key factor. 
As a matter of fact, the deployment of several robots in the same disaster area can yield critical success by potentially allowing a faster coverage of larger areas.
Furthermore, different orders of robot autonomy are required and long-term human-robot collaboration is desired to preside a disaster area over several days (see e.g. \cite{kruijff2012}). 

Therefore, a crucial support to the operators is the ability of the UGVs team to collect information by patrolling the hazardous area and reporting to the operators the gathered knowledge. 

To this end, solving  spatial conflicts between several robots is crucially required in order to attain optimal patrolling in full 3D environments with large amounts of obstacles and obstructed paths.

In this work, we delineate methods for handling strategies to safely govern UGVs behaviors in close proximity.
To show our methodology we focus on autonomous multi-robot path planning and frequency-based patrolling, highlighting the role of  robot inference in resolving, sometimes compelling, conflicts.
We present a distributed multi-robot patrolling technique, which uses a two-level coordination strategy that minimizes and explicitly manages the occurrence of conflict and interference, considering both topological and metric strategies  to solve spatial conflicts. 
The topological strategy  deals with the team coordination by allocating nodes to individual UGVs on a patrolling graph. The metric strategy  attains coordination by ensuring safe traversal and collision free multi-robot operation.

We  show that the proposed framework is capable of operating in full 3D environments, allowing robots to successfully patrol in uneven and unstructured terrain. 
The patrolling algorithm is integrated with a 3D pose-graph \ac{SLAM} system, allowing robots to continuously update and extend their traversable area as well as register their data in a common reference frame using an \textit{OctoMap} representation.
We also present a novel multi-robot traversability analysis that is based on the local shape of the map point-cloud, the spatial arrangement of the team and the robots planned paths.

%Results shown in Section~\ref{Sect:Experiments} demonstrate that the proposed system can face and solve an interesting set of spatial conflicts while minimizing interference. 
Results shown in Section~\ref{Sect:Experiments} (some examples in real scenarios are depicted in Fig.~\ref{Fig:PatrollingScenarios}) demonstrate that the proposed system can face and solve an interesting set of spatial conflicts while minimizing interference. 
Due to the difficulties in operating these UGV systems we augment the set of real world experiments, reported in the experiment section, with simulation experiments that reproduce real scenarios.

In summary, the novel contributions of the work are the following: 
\begin{itemize}
[noitemsep,topsep=0pt,parsep=0pt,partopsep=0pt]
\item[(i)] Patrolling on a 2D manifold embedded in the 3D space.
\item[(ii)] A two level coordination strategy for guiding a team of patrolling robots in a distributed fashion. 
\item[(iii)] Multi-robot traversability analysis considering teammates planning decisions.
\item[(iv)] A validation of our approach in real world environments and in realistic simulation scenarios. 
%\item[(v)] An open-source implementation, which will be available upon this paper acceptance\footnote{\href{https://gitlab.com/luigifreda/3dpatrolling}{ https://gitlab.com/luigifreda/3dpatrolling}}.
\item[(v)] An open source implementation is available\footnote{\href{https://gitlab.com/luigifreda/3dpatrolling}{ https://gitlab.com/luigifreda/3dpatrolling}}.
\end{itemize}
The proposed strategy is presented within a comprehensive system for 3D multi-robot patrolling.

The remainder of this paper is organized as follows.
Section~\ref{Sect:Overview} presents an overview of the main challenges that need to be faced by a team of patrolling robots.
In Section~\ref{Sect:RelatedWork}, we survey works on multi-robot patrolling, though none of them faces the real-world conditions we considered in this work. 
Section~\ref{Sect:ProblemSetup} describes the problem setup.
An overview of the proposed multi-robot patrolling system is given in Section~\ref{Sect:TwoLevelCoordinationStrategy}.
In Section~\ref{Sect:DistributedPatrollingStrategies}, we describe the adopted distributed patrolling strategy. Next, Section~\ref{Sect:MultiRobotPathPlanning} describes the used multi-robot path planning approach, followed by details on our 3D SLAM system in Section~\ref{Sect:3DMap}.
Finally in Section~\ref{Sect:Experiments} we present the results in both real world and simulation experiments and provide implementation details.

%==================================================================
%==================================================================

\section{Problem Overview}\label{Sect:Overview}
%==================================================================
%==================================================================

A team of UGVs is called to patrol a 3D complex environment. A set of locations of interest is assigned and must be continuously visited in order to monitor their surroundings. The team objective is to maximize the visit frequency of each assigned location. Such a mission poses many challenges. 

\noindent {\bf 3D uneven and complex terrain}. The UGVs are required to navigate over a 3D uneven and complex terrain. In general, the 3D terrain shape must be efficiently modelled and properly interpreted 
%by taking into account both geometrical and semantic aspects. This is a fundamental prerequisite 
in order to allow UGVs to robustly localize and plan safe and feasible trajectories.  To this aim, a high-level understanding is typically required beyond a basic geometric 3D representation of the scenario. 

\noindent {\bf Spatial conflicts}.
Narrow passages (for example due to collapsed infrastructures or debris) typically generate spatial conflicts amongst teammates. A suitable strategy is required to \emph{(i)} minimize interferences and \emph{(ii)} recognize and resolve possible incoming deadlocks, which can hinder UGVs activities or even provoke major failures.  

\noindent {\bf Dynamic environment}.
The environment may be dynamic and large-scale~\cite{Cadena16tro-SLAMfuture}. In this case, UGVs must continuously update their internal representations of the surrounding scenario in order to best adapt their behaviours and quickly react to changes. This is a crucial requirement for continuous, efficient and safe operations. 

\noindent {\bf Unreliable communication network}.
In order to collaborate, UGVs must continuously exchange coordination messages and share their knowledge over a network infrastructure. Indeed, real world networks might be unreliable and offer only a limited communication bandwidth. Therefore, the patrolling strategy must rely on an efficient coordination protocol and show robustness with respect to possible communication failures.  

\noindent {\bf Long-term operations}.
Patrolling is a long-term task which requires the adoption of suitable persistent models. UGVs are resource-constrained systems which must be able to efficiently select and integrate only relevant information. At the same time, irrelevant sensory data must be filter out and disregarded. These capabilities are crucially required to maintain a compact and usable knowledge representation in the long-term. 

We address the aforementioned challenges in the following.

%==================================================================
%==================================================================

\section{Related Work}\label{Sect:RelatedWork}
%==================================================================
%==================================================================
%\input{related-works}

Multi-robot patrolling 
%with advantages of spatial distribution and fault tolerance 
has found in recent years several applications in real domains where distributed surveillance, inspection, or control are crucial (e.g., computer network administration~\cite{Andrade-2001,Du-2003}, security~\cite{Agmon-2014,Agmon-2008,Hernandez-2014}, \ac{SaR}~\cite{Acevedo-2013,Aksaray-2015,Pippin-2014}, persistent monitoring~\cite{Song-2014}, hotspot policing~\cite{Chen-2017}, military~\cite{Park-2012}). 
Typically, in this contexts, a team of robots is required to repeatedly visit a set of areas of interest to be monitored~\cite{Ahmadi-2006,Chevaleyre-2004,Elmaliach-2007,Machado-2002,Portugal-2013a,Sak-2008}.

%Two main overlapping taxonomies can be identified for the existing approaches. 
%
Existing approaches can be classified either on the basis of the kind of application~\cite{Agmon-2014,Agmon-2008} or with respect to the applied theoretical principles~\cite{Chevaleyre-2004,Elmaliach-2007,Franchi-2009,Hernandez-2014,Machado-2002,Panagou-2016,Portugal-2016,Santana-2004}.    
Considering the type of application, existing approaches can be divided in adversarial patrol~\cite{Yehoshua-2013}, perimeter patrol~\cite{Agmon-2008a}, and area patrol~\cite{Portugal-2013a}. %, to name a few of them.
Regarding the theoretical baseline, they can be distinguished in pioneer methods~\cite{Machado-2002}, graph theory methods~\cite{Chevaleyre-2004,Portugal-2010}, and alternative coordination methods~\cite{Santana-2004}.

On the basis of recent research advancements in this field, alternative subdivisions might be devised.
%
%A proposal could be to further decompose alternative coordination
For instance, alternative coordination methods can be further decomposed in game theory methods~\cite{Hernandez-2014}, methods resorting to statistical approaches~\cite{Santana-2004,Portugal-2016}, methods using principles from control theory~\cite{Panagou-2016}, and logic-based methods~\cite{Aksaray-2015}. 
An alternative up-to-date review of some of the aforementioned works can be found in~\cite{Portugal-2016} and in~\cite{Yan-2016}. %
The presented work  is developed at the intersection of the pioneer methods and the area patrol classes, addressing scalability and computational complexity constraints. 

Pioneer methods are commonly based on simple architectures where heterogeneous robots with limited perception and communication capabilities are guided to locations that have not been visited for a while, aiming to maintain a high frequency of visits~\cite{Portugal-2013a}.
Under this setting, agents can behave either in a reactive (with local information) or in a cognitive (with access to global information) manner~\cite{Elmaliach-2007,Machado-2002}.
Over the years, these methods led to what is today better known as {\em frequency-based patrolling}~\cite{Chevaleyre-2004,Elmaliach-2009}.
In this type of patrolling, the goal of the team of robots is to optimize a given frequency criterion, usually the {\em idleness}~\cite{Portugal-2010}, that is, the time between consecutive visits to a particular point within the patrol region~\cite{Pasqualetti-2012,Portugal-2013c}.
In~\cite{Portugal-2013b}, the authors state that in some cases, simple strategies like the pioneer ones, with reactive agents, even without communication capabilities, can achieve equivalent or improved performance when compared to more complex ones. 
A study of the scalability and performance of some of the patrolling strategies mentioned above has been reported in~\cite{Portugal-2013a}. 

Despite the focus that multi-robot patrolling has received recently, it can be noted that there is a lack of practical real-world implementations of such systems~\cite{Portugal-2013b}.
When dealing with a team of real robots operating in harsh environments, particular attention has to be payed on the communication, the coordination, and the collaboration amongst teammates for safe joint navigation~\cite{Acevedo-2016b,Bereg-2016,Shahriari-2016}. 
Most of the proposed approaches do not account for 3D environments~\cite{Cabrita-2010,Iocchi-2011,Pasqualetti-2012}. 

In this work, we study the patrolling problem from a non-adversarial point of view.
Specifically, we cast the patrolling problem as an online optimization of the point visit frequency (frequency-based patrolling).
Even if optimal or near-optimal solutions can be typically guaranteed by off-line methods~\cite{Chevaleyre-2004}, we select a online framework in order to best face the compelling uncertainty in perceptions, modelling, and action executions.
We present a multi-robot system which is able to patrol a 2D manifold of the 3D space. 

Many previous multi-robot patrolling systems have been demonstrated under strong assumptions, such as perfect localization, perfect communication
%, agent movements constrained to a grid based space
or assuming no major failures at path planning level.
The drawbacks of these assumptions have been already noticed in the community, ``the theoretical strategies need to be adapted to take into account the uncertainties and dynamics of the actual execution'' as stated in~\cite{Farinelli-2016}. 
In this paper, we present a system tested in real-world scenarios aiming at stepping ``towards better validation processes''~\cite{robin2016multi}.
Our system approaches the online multi-robot patrolling task by fully considering the 3D space with a SLAM system running on each robot.
Specifically, the SLAM system allows the team to be aware of, and adapt to, changes in the environment, for instance, by reassigning goals  when a node is no longer reachable for one of the robots due to changes in the traversability map.
Furthermore, the presented implementation uses \emph{nimbro\_network}~\cite{nimbro} to handle the communication bandwidth which can be scarce in any full integrated system. 

%==================================================================
%==================================================================

\section{The Patrolling Model}\label{Sect:ProblemSetup}
%==================================================================
%==================================================================
%\input{problem_setup.tex}

\begin{figure}[t!]
\centering
	\includegraphics[width=\linewidth]{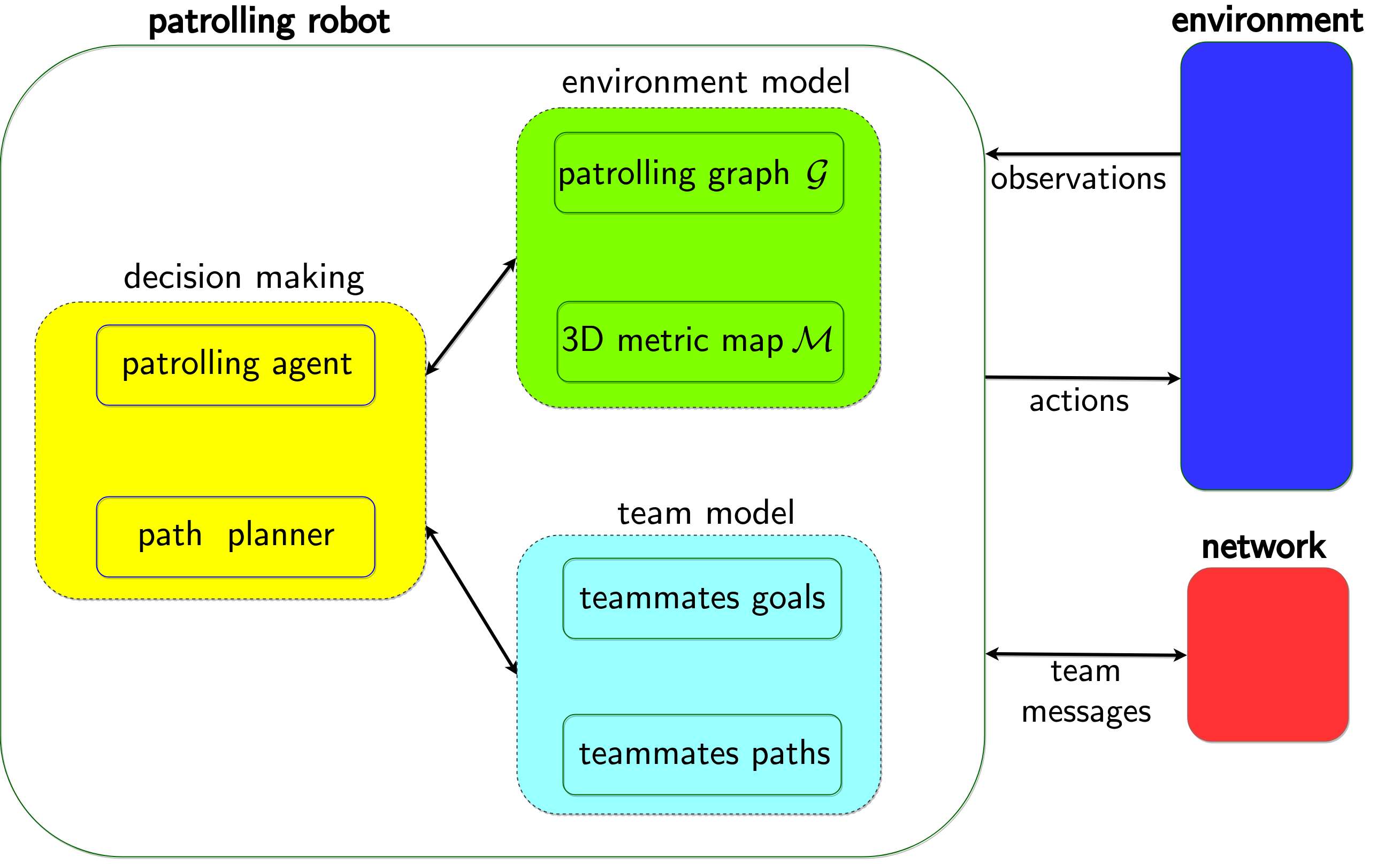}
    \caption{The patrolling robot model.}%
    \label{Fig:PatrollingRobotModel}
\end{figure}

\begin{table}
\begin{center} 
\caption{Table of the main symbols.}
\label{Tab:Symbols}
%\resizebox{0.8\linewidth}{!}{
\begin{tabular}{llll}
Symbol & Description  \\
\cline{1-2}\\[-1.0em]
%\hline
${\cal W}$ & Environment\\
$T$ & Time interval\\
${\cal S}$ & Surface terrain in ${\cal W}$\\
${\cal O}$ & Obstacle region\\
${\cal C}$ & Configuration space of each robot \\
${\cal A}_j(\bm{q})$ & Region occupied by robot $j$ at $\bm{q} \in {\cal C}$ \\
${\cal G}$ & Patrolling graph \\
${\cal M}$ & 3D metric map of the environment  \\
%${\cal U}$ & Robot observation space \\
%${\cal Y}$ & Robot action space \\
\cline{1-2}
%\hline
\end{tabular}
%}
\end{center}
\end{table}

In this section we introduce the model and data structures of our patrolling framework. We focus on a team of robots called to patrol an asperous area for which a terrain condition knowledge is required. 
%The following presentation is based on some standard motion planning concepts~\cite{Lavalle-2006}.

The robot team is composed by $m \geq 2$ ground patrolling robots. The main components of a patrolling robot are represented in Fig.~\ref{Fig:PatrollingRobotModel}.
A patrolling robot interacts with its environment through observations and actions, where an observation consists of a set of sensor measurements and an action corresponds to a robot actuator command. Team messages are exchanged with teammates over a network for sharing knowledge and decisions in order to attain team collaboration.

Decision making is achieved by the patrolling agent and the path planner, basing on the available information stored in the environment model and the team model. In particular, the \emph{environment model} consists of a topological map $\cal G$, aka patrolling graph, and a 3D metric map $\cal M$.
%, which are provided as input to the robot team at the beginning of the mission. 
%The \emph{team model} is an estimate of the current planning state of the team and mainly consists of a list of planned goals and paths.
The \emph{team model} represents the robot belief about the current plans of teammates (goals and planned paths).

%The environment and the robot configuration space are defined in Section~\ref{Sect:3DEnvModel}.The 3D metric map $\cal M$ and the path planner are introduced in Section~\ref{Sect:OverviewMappingTravPathPlanning}.
% The patrolling graph $\cal G$ and the patrolling agent are introduced in Section~\ref{Sect:PatrollingGraph}. 
The main components of the patrolling robots are introduced in the following subsections.
A list of the  main symbols is reported in Table~\ref{Tab:Symbols}.

\subsection{3D Environment, Terrain and Robot Configuration Space}\label{Sect:3DEnvModel}

The 3D \emph{environment} ${\cal W}$ is a compact connected region of $\mathbb{R}^3$. Let $T=[t_0,\infty) \subset \mathbb{R}$ denote a \emph{time interval}, where $t_0 \in \mathbb{R}$ is the starting time. The obstacle region is in general time-varying and denoted by ${\cal O} = {\cal O}(t) \subset \mathbb{R}^3$ for every time $t \in T$. We assume ${\cal O}$ is a collection of low-dynamic objects~\cite{walcott2012dynamic}, whose slow motions do not immediately affect results of robot computations. 

The robots move on a 3D terrain, which is identified as a compact and connected manifold ${\cal S}$ in ${\cal W}$. 

The configuration space $\cal C$ of each robot is the special Euclidean group $SE(3)$ \cite{Lavalle-2006}. In particular, a robot configuration $\bm{q} \in {\cal C}$ consists of a 3D position of the robot representative centre and a 3D orientation. We denote with ${\cal A}_j(\bm{q}) \subset \mathbb{R}^3$ the compact region occupied by robot~$j$ at $\bm{q} \in {\cal C}$.

%The robots are constrained to move over a 3D terrain surface ${\cal S}$, i.e. a compact and connected manifold in ${\cal W}$. 
A robot configuration $\bm{q} \in {\cal C}$ is considered \emph{valid} if the robot at $\bm{q}$ is safely placed over the 3D terrain $\cal S$. This requires $\bm{q}$ to satisfy some validity constraints defined according to \cite{hait2002algorithms}.

A robot path is a continuous function $\bm{\tau}: [0,1] \to {\cal C}$.
A path $\bm{\tau}$ is \emph{safe} for robot $j$ in a time interval $[t_1,t_2] \subset T$ if for each $s \in [0,1]$ and each $t \in [t_1,t_2]$: ${\cal A}_j(\bm{\tau}(s)) \cap {\cal O}(t) = \emptyset$ and $\bm{\tau}(s) \in {\cal C}$ is a valid configuration. %Two paths $\bm{\tau}_i$ and $\bm{\tau}_j$, respectively of robot $i$ and robot $j$, are compatible in $[t_1,t_2] \subset T$ if

We assume each robot in the team is \emph{path controllable}, i.e., each robot can follow any assigned safe path in $\cal C$ with arbitrary accuracy~\cite{Franchi-2009}.

\subsection{Patrolling Graph and Patrolling Agent}\label{Sect:PatrollingGraph}

A \textit{patrolling graph} $\cal G$ is a topological graph-like representation of the environment to be patrolled. 

Namely, ${\cal G} = ({\cal N},{\cal E})$ is an undirected connected graph, with $\cal N$ a set of nodes and ${\cal E} \subseteq {\cal N}^2$ a set of edges. 
%A node represents a region of interest to be continuously monitored (such as a rooms or a specific spot). An edge between two nodes represent a safe path between the corresponding locations. 

A node $n_i \in {\cal N}$ is associated to a 3D region of interest ${\cal B}(n_i) \subset {\cal W}$, and to a \emph{priority weight} $w(n_i) \in \mathbb{R}^+$.  In particular, ${\cal B}(n_i)$ is a ball of pre-fixed radius $R_v \in \mathbb{R}$ centred at the corresponding position $\bm{p}(n_i) \in {\cal S}$. 

An edge $e_{ij} \in  {\cal E}$ between node $n_i$ and $n_j$ denotes the existence of a safe path $\bm{\tau}_{ij}$ connecting the regions ${\cal B}(n_i)$ and ${\cal B}(n_j)$. The length of such a path is used as edge \emph{travel cost} $c(e_{ij}) \in \mathbb{R}^+$. 

A patrolling graph is built before the mission (see Sect.~\ref{Sect:PatrollingGraphBuilding}) and assigned to the team at $t_0$. 

A node $n_j \in {\cal N}$ is \emph{visited} at time  $t \in T$ if a robot centre lies inside the associated region ${\cal B}(n_j)$ at $t$. 

The \emph{instantaneous idleness} $I_j(t) \in \mathbb{R}^+$ of a node $n_j \in {\cal N}$ at time $t \in T$ %is zero if the node is visited at $t$, otherwise is 
is $I_j(t) = w(n_j)(t - t_l)$, where $t_l$ is the most recent time in $[t_0,t]$ the node was visited by a robot. When computing $I_j(t)$, the priority $w(n_j) \in \mathbb{R}^+$ locally ``dilates" or ``contracts" time at node $n_j$. We assume ${I_j(t_0)=0}$ for each node~$n_j$ in ${\cal G}$.

Considering the idleness $I_j(t)$ of a node $n_j$ in a time subinterval $[t_1,t_2] \subset T$, we compute its average idleness $I^a_j[t_1,t_2] = \frac{1}{t_2 - t_1}\int_{t_1}^{t_2} I_j(t) dt$, its standard deviation $I^{\sigma}_j[t_1,t_2]  = \frac{1}{t_2 - t_1}\int_{t_1}^{t_2} \big(I_j(t) - I^a_j[t_1,t_2] \big)^2 dt$ and its maximum value $I^M_j[t_1,t_2] = \underset{t \in [t_1,t_2]}{\text{max}}~I_j(t)$. 

The \emph{average graph idleness} of $\cal G$ is 
\begin{equation}
I_{\cal G}[t_0,t]~=~\frac{1}{N}\sum_{j=1}^{N} I^a_j[t_0,t]
\end{equation}
where $N = \vert {\cal N} \vert$ is the total number of nodes in ${\cal G}$. $N$ is assumed to be constant. 

The \emph{patrolling plan} $\pi$ of a robot is defined as an infinite sequence $\{(n_k,t_k)\}_{k=0}^\infty$, where $n_k \in {\cal N}$ denotes the $k$-th node visited at time $t_k \in T$ by the robot. 
%During the patrolling process, each robot executes its patrolling plan by sequentially visiting the scheduled nodes. 
A \emph{team patrolling strategy} $\Pi = \{\pi_1,...,\pi_m\}$ collects the patrolling plans of all the robots in the team. 

\noindent {\bf Patrolling objective}. In our framework, the goal of the robot team is to cooperatively plan a team patrolling strategy $\Pi$ that minimizes the average graph idleness $I_{\cal G}[t_0,t]$ at all times $t \in T$. 

An instance of the patrolling agent runs on each robot $h$ and is responsible of cooperatively generating the patrolling plan $\pi_h$ according to the above patrolling objective. A pseudo-code description of the patrolling agent is presented in Sect.~\ref{Sect:DistributedPatrollingStrategies}.

\newcommand{\tablespace}{\\[0.1em]}

\begin{table*}[!ht]
\begin{center} 
\caption{Table of broadcast messages.}
\label{Tab:BroadcastMessages}
%\resizebox{0.8\linewidth}{!}{
%\setlength{\tabcolsep}{1em}
\renewcommand{\arraystretch}{1.1}
\begin{tabular}{p{2.8cm}|p{7cm}|p{7cm}}
Broadcast message & Description & Affected data in receiving robot $h$ \\
\cline{1-3}
%\hline
%$\langle ID, node, message type, data $\rangle$ & & \\
$\langle j, t, reached, n \rangle$ & robot $j$ has reached its goal node $n$ &  node $n$ idleness is zeroed in ${\cal I}^{(h)}(t)$; the $j$-th tuple in team model ${\cal T}^{(h)}$ is reset
\tablespace
%\cline{1-3}\\[-1.0em]
$\langle j, t, visited, n \rangle$  & robot $j$ is visiting a non-goal node $n$ along the way to its goal &  node $n$ idleness is zeroed in ${\cal I}^{(h)}(t)$ \tablespace
%\cline{1-3}\\[-1.0em]
$\langle j, t, planned, n \rangle$  & robot $j$ has planned node $n$ as perspective goal &  the $j$-th tuple in team model ${\cal T}^{(h)}$ is filled with $(n,c=\infty,t)$
\tablespace
%\cline{1-3}\\[-1.0em]
%$\langle j, selected, n, \bm{\tau}, c \rangle$  & robot $j$ has selected node $n$ as goal, $\bm{\tau}$ is the planned path to $n$ and $c$ the corresponding path length &  the $j$-th tuple in team model ${\cal T}^{(h)}$ is filled with $(n,\bm{\tau},c)$\\
$\langle j, t, selected, n, c \rangle$  & robot $j$ has actually selected node $n$ as goal and is heading towards it, $c$ is the current path length to the goal &  the $j$-th tuple in team model ${\cal T}^{(h)}$ is filled with $(n,c,t)$
\tablespace
$\langle j, t, path, \bm{\tau}, c \rangle$  & robot $j$ has planned a path $\bm{\tau}$ from its current position to its goal, $c$ is the corresponding path length &  the $j$-th tuple in team model ${\cal T}^{(h)}$ is filled with $(\bm{\tau},c,t)$ and the multi-robot traversability map of robot $h$ is updated (see Sect.~\ref{Sect:TravCost}) \tablespace
%\cline{1-3}\\[-1.0em]
$\langle j, t, aborted, n \rangle$  & robot $j$ aborted its goal node $n$ &  the $j$-th tuple in team model ${\cal T}^{(h)}$ is reset
\tablespace
%\cline{1-3}\\[-1.0em]
$\langle j, t, idleness, {\cal I}^{(j)}(t) \rangle$  & robot $j$ shares its current idleness estimations ${\cal I}^{(j)}(t) = \langle I_1^{(j)}(t),...,I_N^{(j)}(t) \rangle$ &  the current idleness estimations ${\cal I}^{(h)}(t)$ are synchronized with ${\cal I}^{(j)}(t)$ according to Algorithm~\ref{Alg:IdlenessSynchronization}\\ 
\cline{1-3}
%\hline
\end{tabular}
%}
\end{center}
\end{table*}

\subsection{Metric Map and Path-Planning}\label{Sect:OverviewMappingTravPathPlanning}

Each robot of the team is equipped with a rangefinder producing 3D scans\footnote{This can be a rotating laser range-finder or a full 3D scanner.}  and is able to localize in a global map frame, which is shared with its teammates (cfr. Sect.~\ref{Sect:3DMap}).

In our framework, each robot uses a 3D point cloud as a metric representation $\cal M$ of the environment. A map $\cal M$ is built beforehand and assigned to the team at $t_0$. 
A \emph{multi-robot traversability cost} function $trav:\mathbb{R}^3 \to \mathbb{R}$ is defined on ${\cal M}$ (cfr. Sect.~\ref{Sect:TravCost}). %
This function is used to associate a navigation cost $J(\bm{\tau})$ to each safe path $\bm{\tau}$ (cfr. Sect.~\ref{Sect:MixedCostFunction}).

Given the current robot position $\bm{p}_r \in \mathbb{R}^3$ and a goal position $\bm{p}_g \in {\cal S}$, the \emph{path planner} computes a safe path $\bm{\tau}^*$ which minimizes the navigation cost $J(\bm{\tau})$ and connects $\bm{p}_r$ with $\bm{p}_g$ (cfr. Sect.~\ref{Sect:PathPlanningWindowedSearchStrategy}). The path planner reports a failure if a safe path connecting $\bm{p}_r$ with $\bm{p}_g$ is not found.

\subsection{Network Model and Broadcast Messages}\label{Sect:NetworkModel}

Let the \emph{network connectivity graph} $\Phi$ be an undirected graph where a node represents a robot, while an edge represents a communication link between the two connected robot nodes. Specifically, two robots are able to exchange messages if and only if they are connected by an edge in $\Phi$. 

%Two robots are connected in $\Phi$ if they are closer that the communication range $R_c$. 
We assume $\Phi$ is dynamic and stochastic. An edge between any two robots can appear or disappear at any time instant. An independent Bernoulli distribution is associated to each message transmission: any message sent between robots $i$ and $j$ is successfully received with a probability $P^c_{ij} = P^c_{ji}$. We assume the state of $\Phi$ is not observable by the robots.

Each robot can broadcast messages in order to share knowledge, decisions and achievements with teammates.  In particular, a \emph{broadcast message} emitted by robot $j$ at time $t \in T$ is received only by the robots which are connected with robot $j$ on~$\Phi$ at $t$. 

Different types of broadcast messages are used by the robots to convey various information (see Sect.~\ref{Sect:DistributedPatrollingStrategies}). In this process, the identification number (ID) of the emitting robot is included in the heading of any broadcast message. In particular, a broadcast message is emitted by a robot in order to inform teammates when it reaches a goal node (\emph{reached}), 
visits a node (\emph{visited}), 
planned a perspective goal node (\emph{planned}), selected a node as actual goal and it is heading towards it (\emph{selected}), 
and aborts a goal node (\emph{aborted}). 
Additionally, a message \emph{idlenesses} is broadcast in order to enforce the synchronization of idleness estimates amongst teammates (see Sect.~\ref{Sect:SharedKnowledge}). The \emph{path} message will be described in Sect.~\ref{Sect:CoordinatedPathPlanning}. Table~\ref{Tab:BroadcastMessages} summarizes the used broadcast messages along with the conveyed information/data. The vector of estimated idlenesses ${\cal I}^{(h)}(t)$ and the team model ${\cal T}^{(h)}$ are introduced in the next two subsections. The general broadcast message format is $\langle robot\_id, timestap, message\_type, data \rangle$. %Note that, in the \emph{selected} message case, a node $n$ is sent along with the computed safe path $\bm{\tau}$ and its length $c$.

%\removelatexerror
%\rule{\linewidth}{0.1pt}
\begin{algorithm}[t]
 \DontPrintSemicolon
 \nonl\textbf{IdlenessSynchronization}($\langle j, idleness, {\cal I}^{(j)}(t) \rangle$)\;
\nonl~\hspace{-0.5cm}\codecomment{robot $h$ updates ${\cal I}^{(h)}(t)$ by using input \emph{idleness} message}\;
    \For{ $k=1$ {\rm \textbf{to}} $N$} 
    {
    $I^{(h)}_k(t) \leftarrow {\rm min} (I^{(h)}_k(t), I^{(j)}_k(t))$  \;
    }
% 	\textbf{return};
 \caption{IdlenessSynchronization}
 \label{Alg:IdlenessSynchronization}
\end{algorithm}

\subsection{Shared Knowledge Representation}\label{Sect:SharedKnowledge}

%Since our system is totally distributed, 
Each robot of the team stores and updates its individual representation of the world state. 

At $t_0 \in T$, a robot loads as input the 3D map $\cal M$ and the patrolling graph $\cal G$. Then, it internally maintains an instance of these representations. In particular, we denote by ${\cal M}^{(h)}$ and ${\cal G}^{(h)}$ the local instances of $\cal M$ and $\cal G$ in robot $h$, respectively.

Since the environment is dynamic, robot $h$ updates its individual 3D map ${\cal M}^{(h)}$ by using the last acquired 3D scan measurements (see Sect.~\ref{Sect:3DMap}). 
This allows the path planner to safely account for new environment changes. 

At the same time, robot $h$ updates its patrolling graph ${\cal G}^{(h)}$ by using the received broadcast messages and the path planner output. Specifically, the travel cost $c^{(h)}(e_{ij})$ of an edge $e_{ij}$ in ${\cal G}^{(h)}$ is locally updated when a new path is computed between the two corresponding nodes $n_i$ and $n_j$. 

Additionally, robot $h$ locally maintains an idleness estimate $I^{(h)}_j(t)$ for each node $n_j$ in ${\cal G}^{(h)}$. %along with the corresponding graph idleness statistics. 
We denote by ${\cal I}^{(h)}(t) = \langle I^{(h)}_1(t),...,I^{(h)}_N(t) \rangle$ the vector of estimated idlenesses in robot $h$.
Every time a robot visits(reaches) a node $n_j$, a \emph{visited}(\emph{reached}) message is broadcast and each receiving robot $h$ correspondingly updates its local idleness estimate $I^{(h)}_j(t)$. 
Clearly, since broadcast messages may be lost, the idleness estimates $I^{(h)}_j(t)$ may not correspond to the actual idleness values. 
%and may be different on each robot. 
In order to mitigate this problem, each robot continuously broadcasts an \emph{idleness} message at a fixed frequency $1/T_{idln}$. Such messages are used to synchronize the idleness estimates amongst robots according to Algorithm~\ref{Alg:IdlenessSynchronization}.
%and makes it available to its patrolling agent. 

The above information sharing mechanism implements a \emph{shared idleness} representation which allows team cooperation, e.g. minimizing inefficient actions such as re-visiting nodes just inspected by teammates.

\subsection{Team Model}\label{Sect:TeamModel}

In order to cooperate with its team and manage conflicts, robot $h$ maintains an internal belief representation of the current teammate plans (aka \emph{team model}) by using a dedicated table
\begin{equation}
{\cal T}^{(h)} = \langle (n^1_g,\bm{\tau}^1,c^1,t^1),...,(n^m_g,\bm{\tau}^m,c^m,t^m) \rangle
\end{equation} 
which stores for each robot $j$: its selected goal node $n^j_g \in {\cal N}$, the last computed safe path $\bm{\tau}^j$ to $n^j_g$, the corresponding travel cost $c^j \in \mathbb{R}^+$ (i.e. the length of $\bm{\tau}^j$) and the timestamp $t^j \in T$ of the last message used to update $(n^j_g,\bm{\tau}^j,c^j)$. 

The table ${\cal T}^{(h)}$ is updated by using \emph{reached}, \emph{planned}, \emph{selected} and \emph{aborted} messages. In particular, \emph{reached} and \emph{aborted} messages received from robot $j$ are used to reset the tuple $(n^j_g,c^j,\bm{\tau}^j,t^j)$ to zero (i.e. no information available). A \emph{planned} message sets the sub-tuple $(n^j_g,t^j)$, with $c^j = \infty$. A \emph{selected} message sets $(n^j_g,c^j,t^j)$, while a \emph{path} message completes the tuple with $\bm{\tau}^j$ information.

An \emph{expiration time} $T_{exp}$ is used to clean ${\cal T}^{(h)}$ of old invalid information. In fact, part of the information stored in ${\cal T}^{(h)}$  may refer to robots which underwent critical failures or whose connections have been down for a while. In particular, let $t \in T$ be the current time. A tuple $(n^j_g,\bm{\tau}^j,c^j,t^j)$ is reset to zero if $(t - t^j) > T_{exp}$.

\subsection{System Architecture}

The patrolling plan $\pi$ of a robot can be pre-computed \emph{offline}, i.e. before starting the patrolling execution~\cite{Chevaleyre-2004,elmaliach2009multi,Portugal-2010}, or \emph{online}, i.e. by planning and visiting a new node at each patrolling step $k$ \cite{sempe2003adaptive,Portugal-2013a,Portugal-Rocha-2013,Portugal-2016}.

In a \emph{centralized} system, the team patrolling strategy $\{\pi_1,...,\pi_m\}$ is computed by a central control robot (i.e. the \emph{leader}) and communicated to all its teammates. Conversely, in a \emph{decentralized} system, a central leader does not exist. Different levels of decentralization are possible and spans from hierarchical to distributed architectures~\cite{yan2013survey,farinelli2004multirobot,baran1964distributed}. 
In a \emph{distributed} system, each robot independently computes its patrolling plan by possibly taking advantage of exchanged information and coordination messages.

%In this work, the goal of each robot in the team is to cooperatively compute its patrolling plan $\pi$ in order to minimize the average graph idleness $I_{\cal G}[t_0,t]$. 
Our system is online and distributed. In particular, an instance of the \emph{patrolling agent} algorithm (see Sect.~\ref{Sect:DistributedPatrollingStrategies}) runs on each robot and is responsible of online generating its own patrolling plan $\pi$. Namely, at each patrolling step $k$, the patrolling agent plans a new \emph{goal node} $n_k$ in ${\cal G}$. In this process, a patrolling robot exchanges messages with its teammates (see Sect.~\ref{Sect:NetworkModel}) in order to attain \emph{coordination} (avoid conflicts) and \emph{cooperation} (avoid inefficient actions). More details are provided in Sect.~\ref{Sect:TwoLevelCoordinationStrategy}.

%==================================================================
%==================================================================

\section{Two Level Coordination Strategy}\label{Sect:TwoLevelCoordinationStrategy}
%==================================================================
%==================================================================
%\input{two_level_coordination_strategy.tex}

\begin{figure}
\centering
\includegraphics[width=\columnwidth]{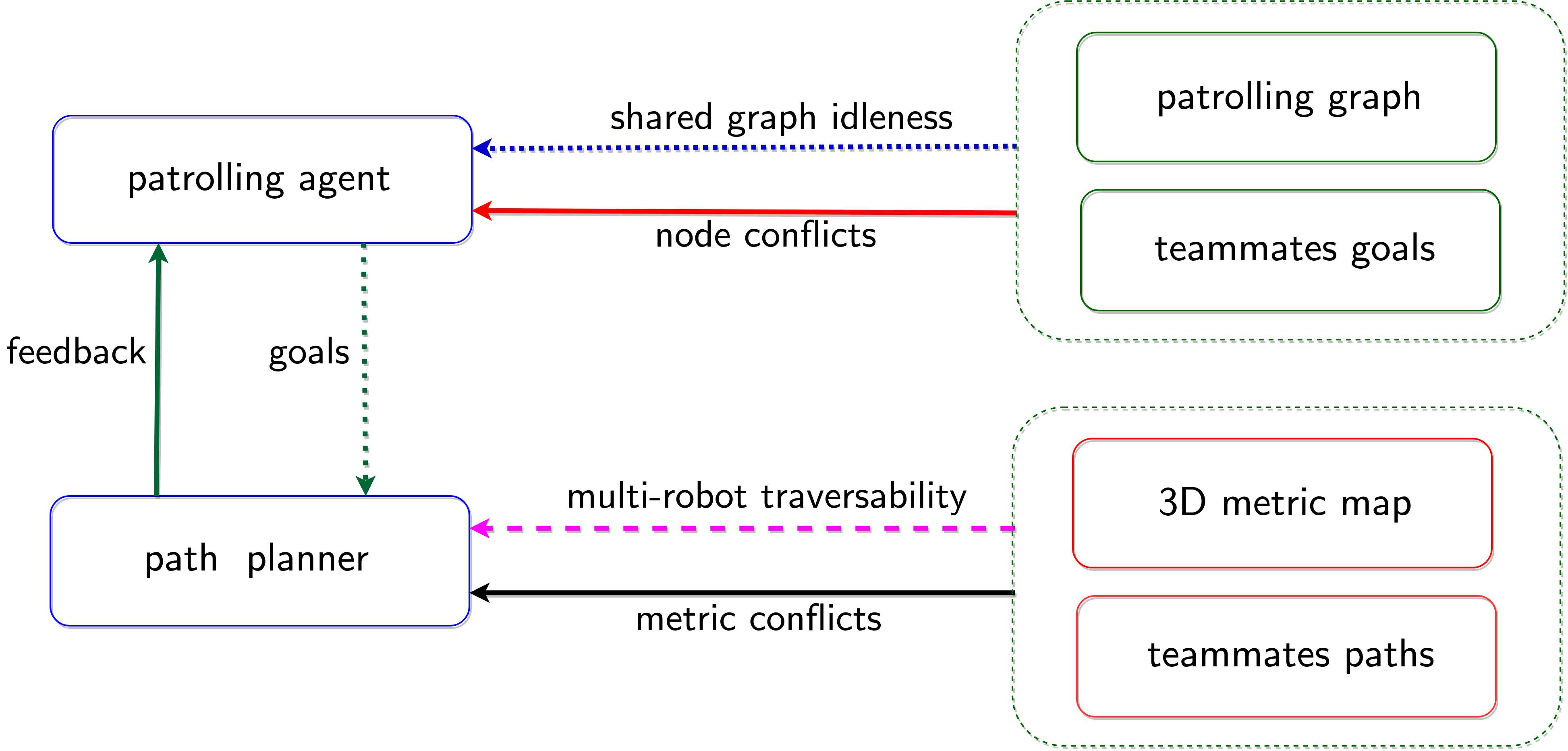}
\caption{The two-level strategy implemented on each robot.}
\label{Fig:TwoLevelStrategy}
\end{figure}

This section first introduces the notions of topological and metric conflicts, and then presents our two-level coordination strategy (see Sect.~\ref{Sect:TwoLevelCoordinationStrategy}).

\subsection{Topological and Metric conflicts}\label{Sect:TopologicalMetricConflicts}

A \emph{topological conflict} between two robots is defined on the patrolling graph $\cal G$. This  
occurs when two patrolling agents select the same node $n_i \in {\cal G}$ as goal (\emph{node conflict}) or plan to simultaneously traverse the same edge $e_{ij} \in {\cal G}$ (\emph{edge conflict}). 

On the other hand, metric conflicts are defined in the 3D Euclidean space where two robots are referred to be in \emph{interference} if their centres are closer than a pre-fixed \emph{safety distance} $D_s$. It must hold $D_s \geq 2 R_b$, where $R_b$ is the bounding radius of each robot, i.e. the radius of its minimal bounding sphere. 
A \emph{metric conflict} occurs between two robots if they are in interference or if their planned paths may bring them in interference\footnote{That is, the distance between the closest pair of points of the two planned paths is smaller than $D_s$. %Clearly, the occurrence of an actual future interference depends on the planned time-scalings. Here, for sake of safety, we consider a worst case scenario.
}.

It is worth noting that topological conflicts may not correspond to metric conflicts. In our framework, an edge may represent a large passage which could be simultaneously traversed by two or more robots without interferences. Similarly, a node may represent a large region which could actually be visited by two or more robots at the same time.

\subsection{Two Level Coordination Strategy}\label{Sect:TwoLevelCoordinationStrategy}

Our patrolling strategy is distributed and supported on both topological and metric levels.

The patrolling agent acts on the \emph{topological} strategy level by selecting the next goal node $n_g$ on ${\cal G}$. In this process, \emph{cooperation} is attained by using the shared idleness representation. This avoids inefficient actions such as selecting nodes just inspected by teammates (see Sect.~\ref{Sect:SharedKnowledge}).

The path planner acts on the \emph{metric} strategy level (see Figure~\ref{Fig:TwoLevelStrategy}) by computing the best safe path from the current robot position to $\bm{p}(n_g)$ by using its internal 3D map ${\cal M}^{(h)}$ (see Sect.~\ref{Sect:PathPlanningWindowedSearchStrategy})).

The patrolling agent guarantees \emph{topological coordination} by continuously monitoring and negotiating possibly incoming node conflicts (see Sect.~\ref{Sect:DistributedPatrollingStrategies}). 
In case multiple robots select the same goal (node conflict), the robot with the smaller travel cost actually goes, while the other robots stop and re-plan towards new nodes. 

The path planner guarantees \emph{metric coordination} by applying a multi-robot traversability function. This induces a prioritized path planning~\cite{Lavalle-2006}, in which robots negotiate metric conflicts by preventing their planned paths from locally intersecting (see Section~\ref{Sect:TravCost}).

The continuous interaction between the patrolling agent and the path planner plays a crucial role. 
When moving towards $\bm{p}(n_g)$, the path planner continuously re-plans the best traversable path till the robot reaches the goal. During this process, 
%the path planner continuously sends a feedback to the patrolling agent. In this process, 
if a safe path is not found, the path planner stops the robot, informs the patrolling agent of a \emph{path planning failure} and the patrolling agent re-plans a new node. On the other hand, every time the path planner computes a new safe path, its length is used as travel cost by the patrolling agent to resolve possible node conflicts.

In our view, the two-way strategy approach allows \emph{(i)} to simplify the topologically based decision making and \emph{(ii)} to reduce interferences and manage possible deadlocks. 
%by relying on the continuous dialogue between patrolling agent and path planner and leveraging topological and metric information. 
In fact, while the patrolling agent focuses on the most important graph aspects (shared idleness minimization and node conflicts resolution), the path planner takes care of possible incoming metric conflicts due to unmanaged topological edge conflicts. %
Moreover, where the path planner strategy may fail alone in arbitrating challenging conflicts, the patrolling agent intervenes and reassigns tasks in order to better redistribute robots over the graph. As a result, these combined strategies minimize interferences by explicitly controlling node conflicts and by planning on multi-robot traversability maps.

%==================================================================
%==================================================================

%==================================================================
%==================================================================
%\input{patrolling_pseudocode.tex}

%===================================================

\begin{figure*}[ht!]

\removelatexerror
 %\rule{\linewidth}{0.1pt}
\begin{algorithm}[H]\label{Alg:PatrolAgent}
 \DontPrintSemicolon
 \nonl \textbf{PatrollingAgent}(\emph{robot\_id}, \emph{patrolling\_graph}, \emph{metric\_map})\;
  %\algolines{$c := y_i$}{ get the class label of the $i$-th sample}
   $is\_goal\_reached\leftarrow$ \emph{true}\;
   $goal\leftarrow \emptyset$\;
  %robot\_status~$\leftarrow$ NORMAL\;  
 \While{true}
 {
% \KwData{this text}
% \KwResult{how to write algorithm with \LaTeX2e }
% update patrolling graph and boolean variables by using Update()\;
  Update() \codecomment{update data structures and boolean variables}\;
 \eIf{ is\_goal\_reached } 
 {	
 broadcast $goal$ is \emph{reached} \;
 $goal~\leftarrow$~PlanNextGoal() \codecomment{plan next goal node}\;
 broadcast $goal$ is \emph{planned} \;
 send $goal$ to path planner \; 
 }
 %else
 {	
% 	\If{ {\rm is\_interference} }
% 	{
%	 	\eIf{ {\rm is\_critical\_interference} {\rm \textbf{and}} {\rm is\_critical\_planning\_failure} }
%	 	{
%	 	Broadcast(ID, node, REALLOCATION)\;
%	 	node~$\leftarrow$~TaskReallocation(ID)\;
%	 	}
%	 	%else
%	 	{
%	 	%SendPathPlanner(node, ABORT)\;
%	 	%Broadcast(ID, node, ABORTED)\;
%	 	\If{ {\rm is\_path\_planning\_failure}}% {\rm \textbf{and}} {\rm \textbf{not}}{\rm(is\_team\_node\_conflict)} }
%	 	{
%	 	SendPathPlanner(node, ABORT)\;
% 		Broadcast(ID, node, ABORTED)\;  
%	 	Wait($T_{wait}$ )\;
%	 	Broadcast(ID, node, PLANNED)\;
%	 	}
%	 	%node~$\leftarrow$~PlanNextNode()\;
%	 	}
% 	}
 	\eIf{ is\_path\_planning\_failure {\rm \textbf{or}} is\_node\_conflict {\rm \textbf{or}}  is\_goal\_visited }
 	{
 	send \emph{abort} to path planner\;
 	broadcast $goal$ is \emph{aborted} \;  		 	 
    $goal~\leftarrow$~PlanNextGoal() \codecomment{replan next goal node}\;
 	broadcast $goal$ is \emph{planned}\;
 	send $goal$ to path planner\; 	
 	} 	
 	%else
 	{
    broadcast $goal$ \emph{selected} \codecomment{broadcast a \emph{selected} message while reaching the goal}\;
 	sleep for $T_{sleep}$\;
 	}
 }
 }% end while
 \caption{PatrollingAgent}
\end{algorithm}
 %\rule{\linewidth}{0.1pt}
 
%\medskip

\bigskip

\removelatexerror
%\rule{\linewidth}{0.1pt}
\begin{algorithm}[H]\label{Alg:Update}
 \DontPrintSemicolon
 \nonl \textbf{Update}()\;
%  \nonl \codecomment{update in robot $h$}\;
    update idlenesses ${\cal I}^{(h)}(t)$ and travel costs in ${\cal G}$ \codecomment{asynchronous update through received messages and path planner feedback}\;
    update metric map ${\cal M}^{(h)}$ and traversability map \codecomment{asynchronous update through sensor callbacks}\;
    update team model ${\cal T}^{(h)}$ \codecomment{asynchronous update through received messages and path planner feedback}\;
    $is\_node\_conflict~\leftarrow$~ check if another robot in ${\cal T}^{(h)}$ has the same goal node \;
    $is\_goal\_reached~\leftarrow$~check if current $goal$ has been reached by this robot \; 
    $is\_goal\_visited~\leftarrow$~check if current \emph{goal} is visited by another robot  \codecomment{check by using received \emph{visited} messages}\;  
    $is\_path\_planning\_failure~\leftarrow$~check if path planner failed to compute a path to $goal$ \codecomment{check continuous replanning}\;
    $is\_critical\_path\_planning\_failure~\leftarrow$~check if path planning failure is lasting more than $T_{pcr}$\;
    $is\_critical\_node\_conflict~\leftarrow$~check if robot is experiencing node conflicts for more than a time interval $T_{ncr}$\;
% 	\textbf{return}\;
$is\_node\_visited~\leftarrow$~check if a non-goal node is visited by this robot while reaching the current $goal$ \;  
\If{ is\_node\_visited }
{
$node~\leftarrow$ get node visited along the way\;
broadcast \emph{node} is \emph{visited} \codecomment{inform teammates about the non-goal node visit}\;
}
broadcast \emph{idleness} message with pre-fixed frequency $1/T_{idln}$
 \caption{Update (in robot $h$)}
\end{algorithm}

\bigskip

\removelatexerror
%\rule{\linewidth}{0.1pt}
\begin{algorithm}[H]\label{Alg:SelectNexNode}
 \DontPrintSemicolon
 \nonl \textbf{PlanNextGoal}()\;
    $goal\leftarrow \emptyset$\;
   	\eIf{ is\_critical\_path\_planning\_failure {\rm \textbf{or}} is\_critical\_node\_conflict}
 	{
	$goal\leftarrow$~ComputeRandomNode() \codecomment{randomized selection of next node} \;
 	}
 	%else
 	{
 	${\cal D} \leftarrow$~BuildSearchSet() \codecomment{build a search set with candidate goal nodes}\;
 	$goal\leftarrow$~ComputeNextBestNode(${\cal D}$) \codecomment{compute next best node in ${\cal D}$}\;
 	}
% 	Broadcast(ID, node, PLANNED)\;
% 	SendPathPlanner(node, GO)\;
 	\textbf{return} $goal$;
 \caption{PlanNextGoal}
 
\end{algorithm}

%\rule{\linewidth}{0.1pt}
\end{figure*}
%===================================================

%==================================================================
%==================================================================

\section{Distributed Patrolling}\label{Sect:DistributedPatrollingStrategies}

%==================================================================
%==================================================================
%\input{patrolling_strategies.tex}

\newcommand{\gridcrimageheight}{6cm}

\begin{figure*}[ht]
\centerfloat
\setlength{\tabcolsep}{0.2em}
\def\arraystretch{0.} % for lines spacing 
\begin{tabular}{ccc}
\subfloat[]{\includegraphics[width = \gridcrimageheight, keepaspectratio]{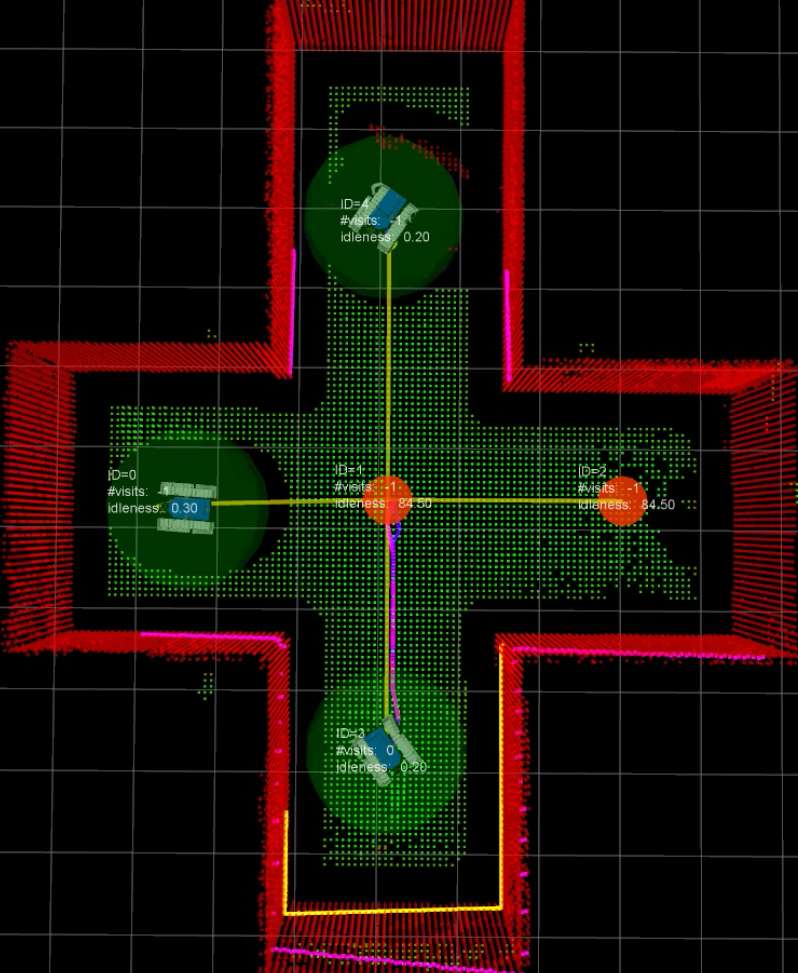}}
&
\subfloat[]{\includegraphics[width = \gridcrimageheight, keepaspectratio]{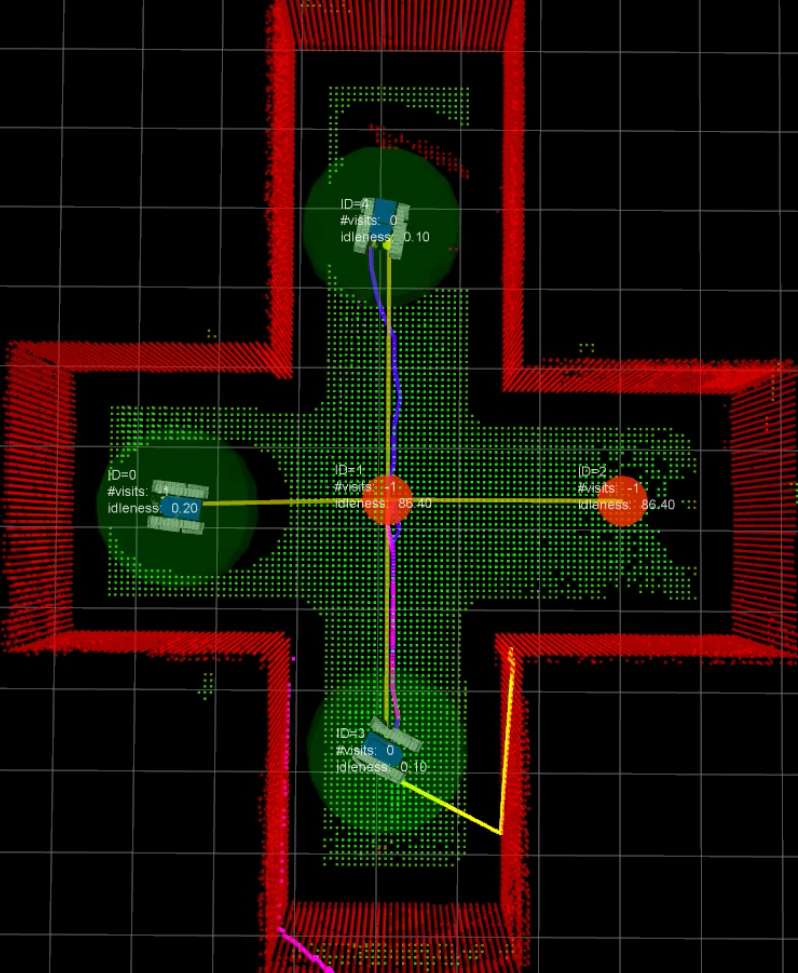}}
&
\subfloat[]{\includegraphics[width = \gridcrimageheight, keepaspectratio]{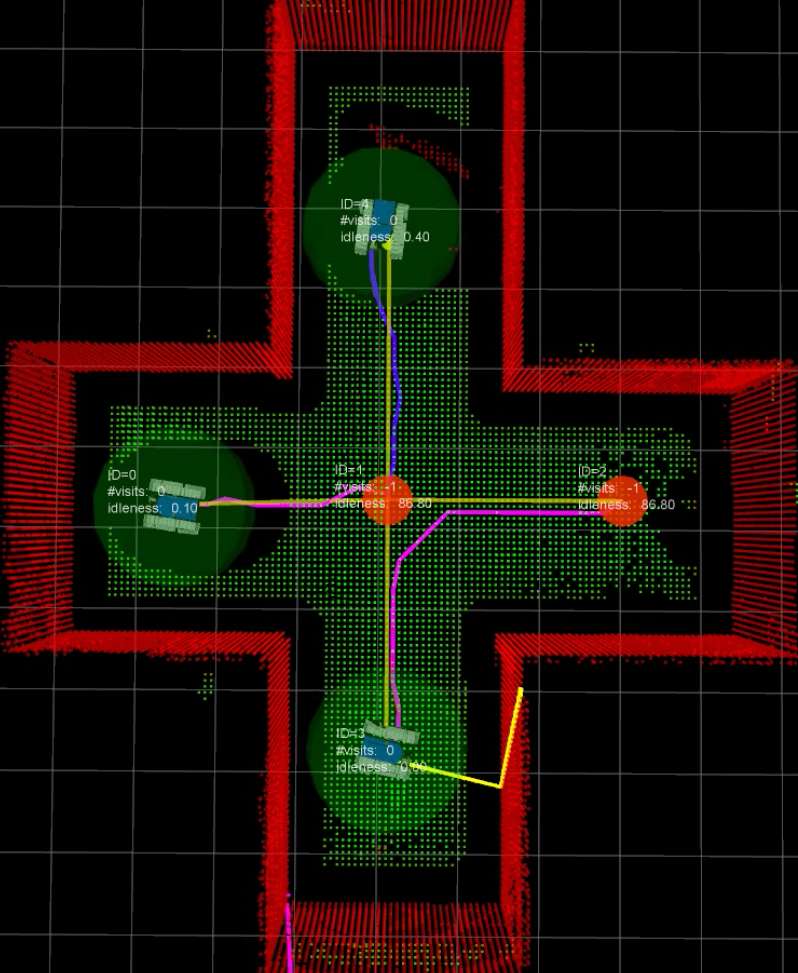}}
\\
\subfloat[]{\includegraphics[width = \gridcrimageheight, keepaspectratio]{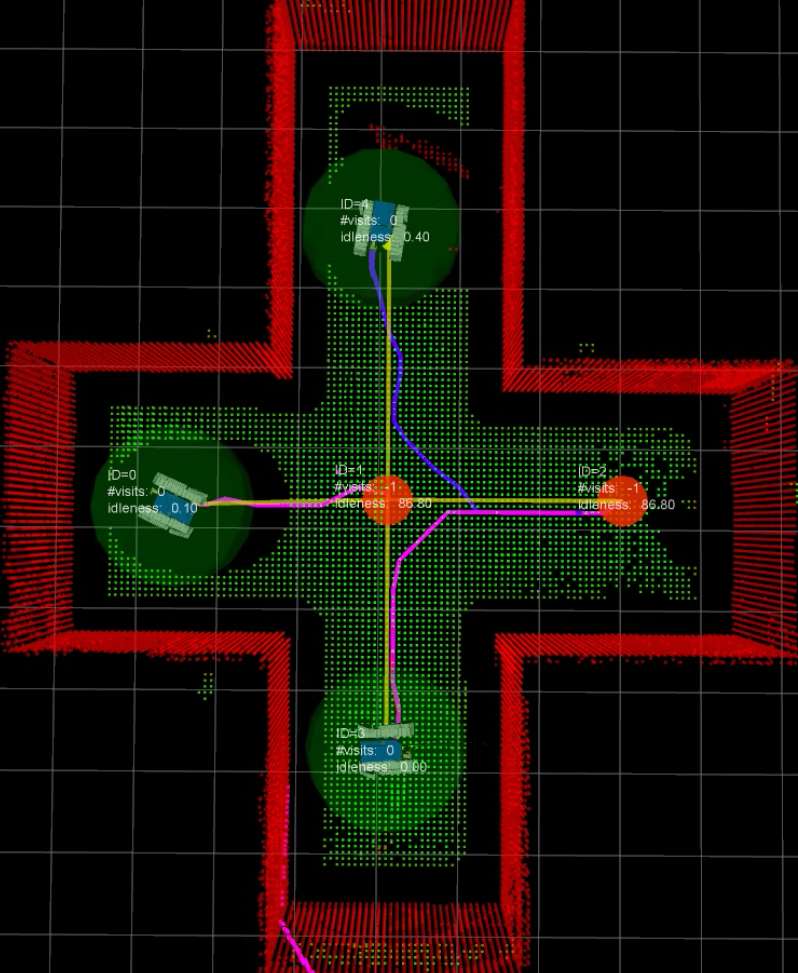}}
&
\subfloat[]{\includegraphics[width = \gridcrimageheight, keepaspectratio]{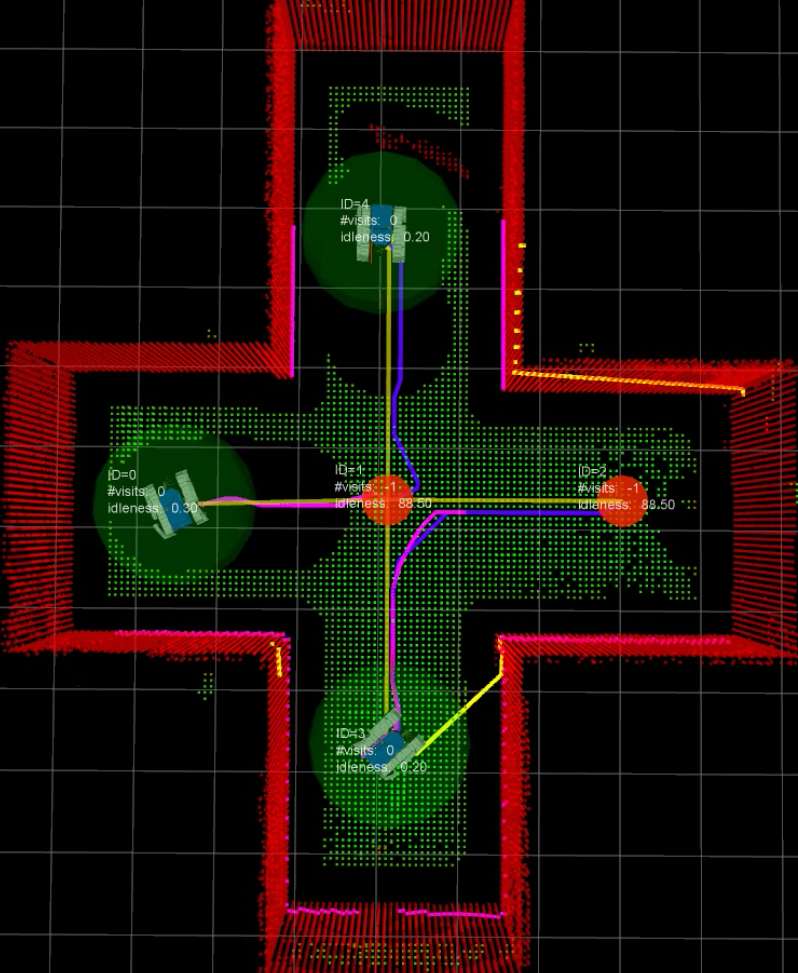}}
&
\subfloat[]{\includegraphics[width = \gridcrimageheight, keepaspectratio]{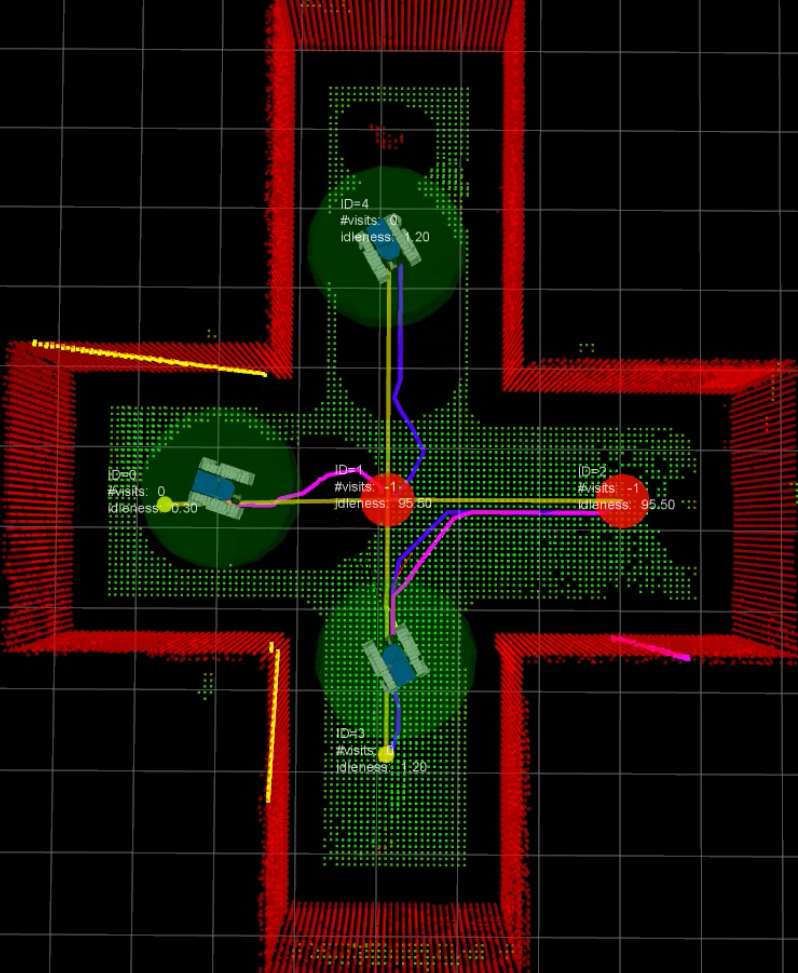}}
\end{tabular}
\caption{A sequence of node negotiations amongst: top robot $t$, left robot $l$ and bottom robot $b$. The  patrolling graph is shown: nodes are depicted as disks; each node has a radius proportional to its idleness. The traversability map of robot $b$ is shown: red points are obstacles; green points are traversable (for robot $b$). Planned paths are emanated from each robot. Both global and local paths are shown (respectively, blue and magenta).  Fig.~(a): robot $b$ plans the central node $n_c$ and then selects $n_c$. Fig.~(b): robot $t$ also plans $n_c$. Fig.~(c): robot $b$ detects a node conflict (with robot $t$) on node $n_c$, aborts $n_c$, plans the right node $n_r$; robot $l$ plans $n_c$. Fig.~(d): robot $l$ selects $n_c$; robot $t$ detects a node conflicts (with robot $l$) on $n_c$, aborts $n_c$ and then plans $n_r$; robot $b$ selects $n_r$. Fig.~(e): robot $t$ detects a node conflict (with robot $b$) on $n_r$, aborts $n_r$ and plan $n_c$. Fig.~(f): both robot $l$ and $b$ are moving towards their goals while robot $t$ is searching for a reachable and non-conflicting node. At this time, robot $t$ observes that each node is either selected by a closer robot, currently visited or unreachable.}
\label{Fig:ConflictResolution}
\end{figure*}

In this section, we present in detail the patrolling agent algorithm. A pseudocode description is reported in Algorithm~\ref{Alg:PatrolAgent}.

A patrolling agent instance runs on each robot. 
It takes as input the robot ID, the patrolling graph and the metric map. A main while loop supports the patrolling algorithm (lines 3--22). 
First, all the relevant data structures and the main boolean variables\footnote{We use an ``\emph{is\_}" prefix to denote boolean variables.} are updated (line 4, see Section~\ref{Sect:Update}). This update takes into account all the information received from teammates and recasts the distributed knowledge. 
If the current goal node has been reached (line 5), a broadcast message informs the team (line 6). Then, a new node is planned, a corresponding broadcast message is emitted and the goal position is sent to the path planner (lines 7--9, see Sect.~\ref{Sect:NextNodeSelection}). 

On the other hand, if the robot is still reaching the current goal node, lines 11--20 are executed. 
If a path planner failure, a node conflict (see Section~\ref{Sect:TeamNodeConflict}), or a node visit (see Section~\ref{Sect:Update}) occurs on the selected goal (line 11), the patrolling agent first sends a goal abort to the path planner, next broadcasts its decision and then triggers a new node selection (lines 12--16). Otherwise (lines 18--19), a \emph{selected} message is broadcast and a sleep for a pre-fixed time interval $T_{sleep}$ allows the robot to continue its travel towards the selected goal (line 18). 

It is worth noting that the condition at line 11 of Algorithm~\ref{Alg:PatrolAgent} allows each robot to modify its plan at need while reaching the goal. Moreover, a \emph{selected} message broadcast is repeated at each step\footnote{Or at a pre-fixed frequency, after a first \emph{selected} is broadcast along the way to the current goal.} in order to add robustness with respect to network failures.

\subsection{Data Update}\label{Sect:Update}

The Update() function is summarized in Algorithm~\ref{Alg:Update}. This is in charge of refreshing the robot data structures presented in Sect.~\ref{Sect:ProblemSetup}. Indeed, these structures are asynchronously updated by callbacks which are independently triggered by received broadcast messages or path planner feedback messages. 

Lines 1--3 of Algorithm~\ref{Alg:Update} represent the asynchronous updates of the local instances of the patrolling graph ${\cal G}$, the point cloud map ${\cal M}$ and the team model ${\cal T}$. The remaining lines describe how the reported boolean variables are updated depending on the information stored in the team model and received through path planner feedback.

\subsection{Node Conflict Management}\label{Sect:TeamNodeConflict}

The concept of topological conflict was defined in Section~\ref{Sect:TopologicalMetricConflicts}. During the patrolling process, a topological node conflict occurs when two or more patrolling agents select the same goal node, which we refer to as \emph{contended node}. Our strategy resolves a topological conflict by assigning the contended node to the robot which can reach it with the smallest travel cost.  

A robot checks for node conflicts by using the
information stored in its individual team model (cfr. Sect. \ref{Sect:TeamModel}). In this process, it compares its plan with those of teammates. In particular, robot $j$ detects a \emph{node conflict} with robot $i$ at node $n_g \in {\cal N}$ if the following conditions are verified:
\begin{enumerate}
[noitemsep,topsep=0pt,parsep=0pt,partopsep=0pt]
%\item $n$ is a contended node between robot $i$ and $j$.
\item robots $j$ finds in its team model ${\cal T}^{(j)}$ that robot $i$ has the same goal, i.e., $n^j_g = n^i_g$ in ${\cal T}^{(j)}$.
%\item and t can currently reach node $n$ with a finite navigation cost. 
%This means 
%\emph{a)} the last messages both robots received from their path planners is a SUCCESS message (see Section~\ref{Sect:PathPlanningMessages}) \emph{b)} 
%both robots broadcast a SELECT message  with $n$ (see Section~\ref{Sect:Update}).  
\item the travel cost $c_j$ is  higher than $c_i$ in ${\cal T}^{(j)}$, or $j>i$ in the unlikely case the travel costs $c_j$ and $c_i$ are equal (robot priority by ID as a fall-back). 
%Each robot is able to independently verify this condition by using the received SELECTED messages.  
\end{enumerate}

When the two above conditions are verified, robot $j$ sets the boolean variable \emph{is\_node\_conflict} to true (line 4 of Alg.~\ref{Alg:Update}), aborts its current goal $n_g^j$ and re-plans a new node (lines 12--16 of Alg.~\ref{Alg:PatrolAgent}). 

%A node conflict becomes a \emph{critical node conflict} when it lasts more that a pre-fixed time $T_{ncr}$. 
If a robot experiences node conflicts for more than a pre-fixed time interval $T_{ncr}$, it enters in a \emph{critical node conflict} state. 
In this case, a boolean variable \emph{is\_critical\_node\_conflict} is set true (line 9 of Alg.~\ref{Alg:Update}).

As an example, we report in Fig.~\ref{Fig:ConflictResolution} a sequence of node negotiations amongst three robots.

\subsection{Next Node Planning and Selection}\label{Sect:NextNodeSelection}

The strategy adopted for planning the next node is described in Algorithm~\ref{Alg:SelectNexNode}. First, the algorithm verifies if a \emph{critical condition} is occurring (line 2), i.e., if either a critical path planning failure (see Section~\ref{Sect:CoordinatedPathPlanning}) or a critical node conflict is occurring (see Section~\ref{Sect:TeamNodeConflict}). 
If a critical condition is not occurring (line 3), a \emph{search set} $\cal D$ (i.e., a set of candidate goal nodes) is built (line 5), then the next best node is computed in $\cal D$ (line 6). Here, the functions BuildSearchSet($\cdot$) and Compute\-Next\-Best\-Node($\cdot$) can encode any user-defined strategy with the proviso that $\cal D$ must not contain the possible contended node in case \emph{is\_node\_conflict} is true. 

On the other hand, if a critical condition occurs (line 2), a randomized node selection is performed on the graph (line 3). 
Such a \emph{randomized} selection is used to crucially discharge the planner from any search space restriction (line 5) and selection strategy (line 6). In fact, these may trap the algorithm in a ``\emph{local minimum}", where the planner continuously selects a temporary unreachable node as goal. 

For instance, a search space restriction (line 5) at graph depth $d=1$ (aka reactive strategy) makes the robot stuck idle when reachable nodes are available only at depth $d > 1$.

On the other hand, ``\emph{local minima traps}" can be envisioned on the top of any \emph{deterministic} selection strategy (line 6) by introducing a virtual objective function which combines together the explicit user-defined ``utility" function\footnote{In our case, this depends on the idlenesses of the nodes.} and the navigation cost-to-go. 
%For instance, a local minima trap occurs when a door is suddenly closed in front of a robot and blocks its way towards the node with the highest ``utility".
Indeed, a local minima trap occurs when an obstruction blocks the robot way towards the node $\bm{n}^*$ with the highest ``utility". For instance, the obstruction ``disconnecting" $\bm{n}^*$ can be a door suddenly closed or a group of teammates persisting in front of the robot. In such cases, a randomized selection technique results in an effective method to escape local minima in terms of computational efficiency, generality and reliability~\cite{barraquand1992numerical}.

Algorithm~\ref{Alg:SelectNexNode} can be used as a base to support any online strategy. 
In this work, as an example, we use a reactive strategy for the implementations of the functions Build\-Search\-Set($\cdot$) and Com\-pute\-Next\-Best\-No\-de($\cdot$). Such a strategy effectively provides readiness in resolving incoming spatial conflicts and in making decisions on rapidly changing patrolling graphs. Specifically, we build $\cal D$ as the current node neighbourhood (line~4, Algorithm~\ref{Alg:SelectNexNode}) and select as best node the one in $\cal D$ with the highest idleness estimate (line~5, Algorithm~\ref{Alg:SelectNexNode}). This implementation can be considered as an improved version of the Conscentious Reactive algorithm~\cite{Portugal-2013a}. In fact, here we explicitely manage interferences and spatial conflicts in order to prevent deadlocks. 

For efficiency reasons, in the function \linebreak Com\-pute\-Random\-Node($\cdot$) (line~3, Algorithm~\ref{Alg:SelectNexNode}), the randomized strategy first selects a node at a graph depth one, then it linearly increases the depth of the search with time if the current critical condition is not readily escaped. In order to preserve probabilistic completeness, the randomized selection is performed on the full patrolling graph after a number of consecutive failures. 

Two important observations are in order. First, local minima (critical conditions) are detected thanks to the continuous interaction between the patrolling agent and the path planner.
Second, the presented Algorithm~\ref{Alg:PatrolAgent} puts into effect a cooperative strategy if the adopted Com\-pute\-Next\-Best\-No\-de($\cdot$) function selects the next node on the basis of the shared idleness representation (cfr. Sect. \ref{Sect:SharedKnowledge}). The latter allows to avoid inefficient actions, such as selecting a goal node recently visited by a teammate.

%==================================================================
%==================================================================

\section{Multi-robot Traversability and Path Planning}\label{Sect:MultiRobotPathPlanning}
%==================================================================
%==================================================================
%\input{multi_robot_pp.tex}

\begin{figure}
\centering
\includegraphics[width=\columnwidth]{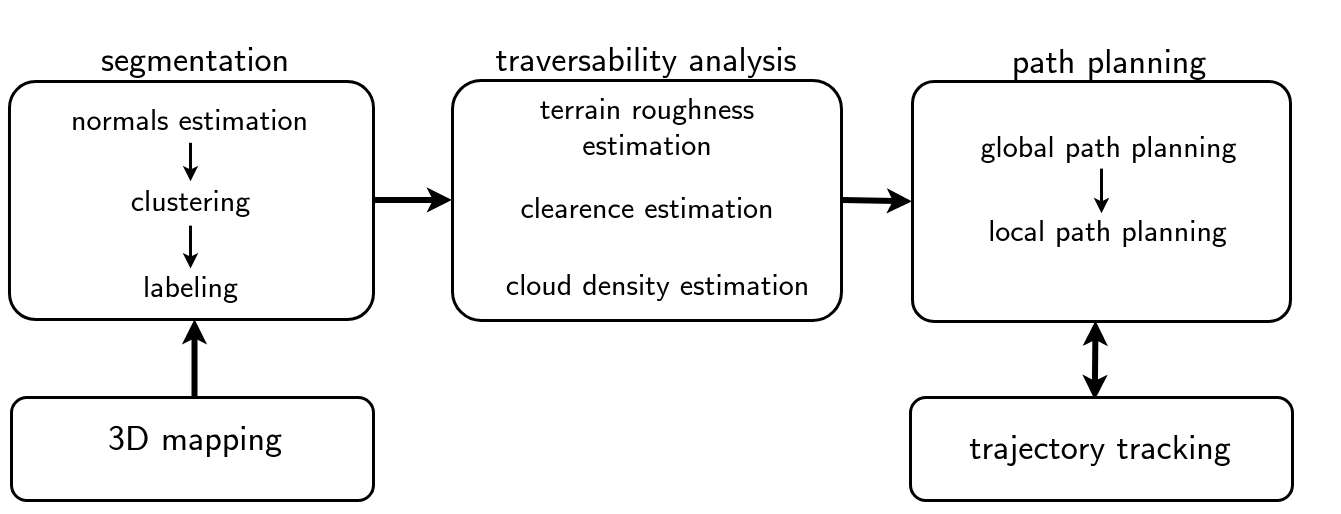}
\caption{The metric level and its main modules.} 
\label{Fig:MetricLevel}
\end{figure}

Basing on the metric strategy, the path planner attains local coordination by applying a multi-robot traversability function. This allows to compute a traversable path towards the designated goal node and to locally negotiate metric conflicts. Figure~\ref{Fig:MetricLevel} presents the metric level and its main modules, which are described in the following subsections.

\subsection{Point Cloud Segmentation}\label{Sect:MapSegmentation}

At each new scan, the robot updates its individual 3D map (see Section.~\ref{Sect:3DMap}). 
%and a structure interpretation of the map is updated. 
Map points are then segmented in order to estimate a traversability of the terrain. 
%In a first step, the point cloud map $\cal M$ is actually built and filtered by using an Octomap representation~\cite{Hornung-2013}.
First, geometric features such as surface normals and principal curvatures are computed and organized in histogram distributions. Clustering is applied on 3D coordinates of points, mean surface curvatures and normal directions~\cite{Menna-2014,Ferri-2014}. As a result, a classification (\emph{labeling}) of the 3D map in regions such as \textit{walls}, \textit{terrain}, \textit{surmountable obstacles} and \textit{stairs/ramps} is obtained. All regions which are not labeled as walls are referred to as \emph{non-walls}.

\subsection{Multi-robot Traversability}\label{Sect:TravCost}

%\subsection{Local Terrain Assessments and Robot Body Representation}
%\label{Sect:TerrainAssessments}

The path planner computes a traversable path $\bm{\tau}$ directly on the segmented non-walls regions of the individual robot 3D map. %The connectivity of the configuration space ${\cal C}$.

Denote with $\mathbb{S}$ a metric space on $\mathbb{R}^3$. Let $\bm{p} \in \mathbb{S}$ and $\varepsilon \in \mathbb{R}^+$ be the center and the radius of a ball ${\cal B}(\bm{p},\varepsilon) \subset \mathbb{S}$, in which we consider a suitably connected neighbourhood of $\bm{p}$. Each non-wall point $\bm{p}$ is evaluated along with its local neighbourhood ${\cal B}(\bm{p},\varepsilon)$ and ``back-projected" onto a robot pose $\bm{q}$ by using the local surface normal at $\bm{p}$~\cite{Krusi-2016}.

For efficiency reasons\footnote{The metric level modules must run on the robot main board and share computational resources with other demanding processing nodes~\cite{Kruijff-2015}.}, each robot body is represented by its bounding sphere when computing its clearance from obstacles and teammates. 
This allows faster computations for both the traversability analysis and the path planner (see Section.~\ref{Sect:CoordinatedPathPlanning}). In this context, the path planner can restrict the path search in a ``projection" of $\cal C$ on a 3D Euclidean space\footnote{At this stage, we found this approach to perform very well in practice without significantly limiting the robot manoeuvres in the tested scenarios.}.

\begin{figure}
\centering
\includegraphics[width=\columnwidth]{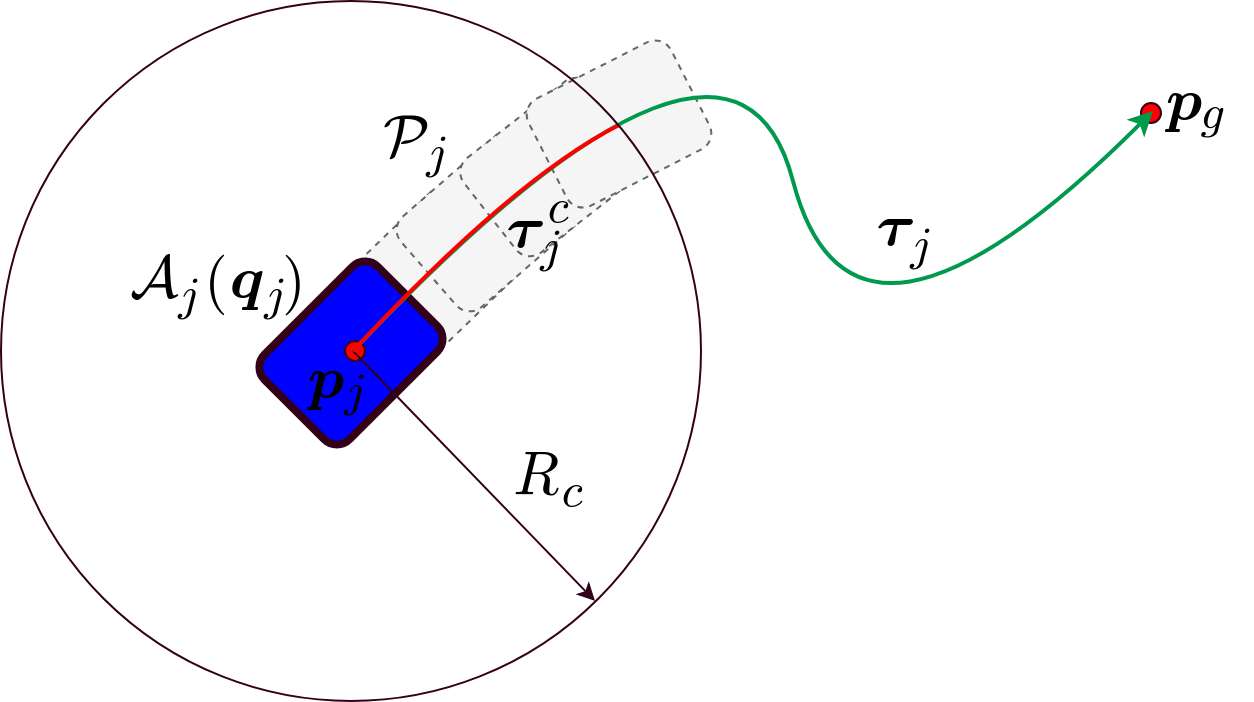}
\caption{A 2D sketch of a robot future trail. The blue(dark) rectangle represents the footprint of robot $j$ at its current pose $\bm{q}_j$. The current robot position $\bm{p}_j$ is the centre of the blue rectangle. ${\cal A}_j(\bm{q}_j)$ corresponds to the area of the blue rectangle. The 2D projection of the current planned path $\bm{\tau}_j$ joins $\bm{p}_j$ with the goal position $\bm{p}_g$.  $\bm{\tau}^c_j$ is the portion of $\bm{\tau}_j$ that keeps the robot centre within ${\cal B}(\bm{p}_j,R_c)$. The 2D projection of $\bm{\tau}^c_j$ is represented in red. Some future robot footprints along $\bm{\tau}^c_j$ are sketched in light grey. %The future trail ${\cal P}_j$ contains the portions of future robot footprints lying in ${\cal B}(\bm{p}_j,R_c)$. Indeed, 
The future trail ${\cal P}_j$ is the union of all the footprints whose centres lie in  ${\cal B}(\bm{p}_j,R_c)$.}
\label{Fig:FutureTrail}
\end{figure}

\begin{figure}
\centering
\includegraphics[width=\columnwidth]{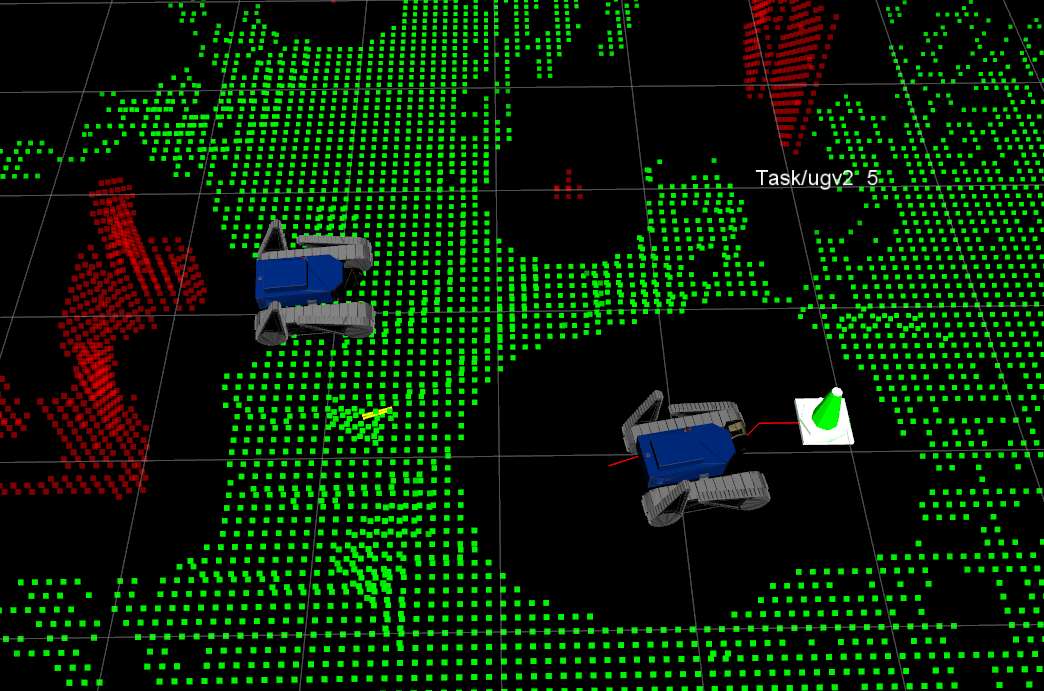}
\caption{The multi-robot traversability map of the left robot $l$. Green points can be traversed by robot $l$. Segmented obstacle points are shown in red. The planned path of the right robot $r$ is reported in red on the ground. The future trail of robot $r$ generates a local ``repelling region" on the green carpet around robot $r$ itself. }
\label{Fig:MultiRobotTraversability}
\end{figure}

Traversability for each robot is computed as a cost function on its 3D map.  
%by taking into account the described point cloud classification and the computed local geometric features. 
To this end, each neighbourhood ${\cal B}(\bm{p},\varepsilon)$ of a map point $\bm{p} \in \mathbb{R}^3$ is evaluated along with its local geometric features and segmented aspects (see Sections~\ref{Sect:MapSegmentation}).

In particular, the traversability cost function $trav:\mathbb{R}^3\to\mathbb{R}$
is computed as
%\begin{equation}\label{Eq:multi_traversability}
%trav(\bm{p}) = w_L(\bm{p})(1+w_{Cl}(\bm{p}))(1 + w_{Dn}(\bm{p}))(1 + w_{Rg}(\bm{p})))
%\end{equation}
\begin{equation}
\label{Eq:multi_traversability}
\Scale[0.91]{
trav(\bm{p}) = w_L(\bm{p})(1+w_{Cl}(\bm{p}))(1 + w_{Dn}(\bm{p}))(1 + w_{Rg}(\bm{p}))
}
\end{equation}
Here  the weight $w_L: \mathbb{S} \to \mathbb{R}^+$ depends on the point classification, $w_{Cl}: \mathbb{S} \to \mathbb{R}^+$ is the multi-robot clearance (defined below), $w_{Dn}: \mathbb{S} \to \mathbb{R}^+$ depends on the local point cloud density and $w_{Rg}: \mathbb{S} \to \mathbb{R}^+$ measures the local terrain roughness (average distance of outlier neighbour points from a local fitting plane).

In order to attain a look-ahead path planning with local coordination and obstacle avoidance behaviours, the traversability analysis of a robot is ``informed" with the current positions and planned paths of its teammates. 

In particular, let $\bm{q}_j \in {\cal C}$ and $\bm{p}_j \in \mathbb{R}^3$ respectively denote the current pose and position of robot~$j$.
${\cal A}_j(\bm{q}) \subset \mathbb{R}^3$ is the compact region occupied by robot~$j$ at $\bm{q} \in {\cal C}$. Denote with  $\bm{\tau}_j: [0,1] \to {\cal C}$ the current planned path, which leads robot~$j$ to its assigned goal configuration. Moreover, let $\bm{\tau}^c_j: [0,1] \to {\cal C}$ be the portion of $\bm{\tau}_j$ which keeps the robot centre within ${\cal B}(\bm{p}_j,R_c)$, a closed ball of radius $R_c$ centred at $\bm{p}_j$ (see Fig.~\ref{Fig:FutureTrail}). Here $R_c$ is a pre-fixed cropping radius. 

The \emph{future trail} of robot $j$ is defined as the compact region: 
\begin{equation}
{\cal P}_j \triangleq \underset{s \in [0,1]}{\bigcup} {\cal A}_j(\bm{\tau}^c_j(s)).
\end{equation}
 In other words, the future trail of robot $j$ is the 3D region the robot would cover along $\bm{\tau}_j$ 
%. This future trail along $\bm{\tau}_j$ spans 
up to a maximum distance $R_c$ from $\bm{p}_j$ (see Fig.~\ref{Fig:FutureTrail}). If no goal is assigned, one has ${\cal P}_j \equiv {\cal A}_j(\bm{q}_j)$. 

Robot~$i$ computes the \emph{multi-robot clearance} $w_{Cl}(\bm{x})$ as its clearance at $\bm{x} \in \mathbb{R}^3$ with respect to \emph{a)} obstacles sensed at its current position $\bm{p}_i$ \footnote{Here we include the segmented obstacles in the map and the most recent nearby obstacle points which have been detected by the rangefinder and are not segmented yet in the map
.} 
\emph{b)} 
each teammate future trail ${\cal P}_j$, with $j\neq i$, such that ${\cal P}_j~ \cap ~{\cal B}(\bm{p}_i,R_t) \neq \emptyset$. Here, $R_t$ is a pre-fixed radius greater than $R_c$. Specifically, when computing $w_{Cl}(\bm{x})$, any teammate future trail ${\cal P}_j$ that is distant more than $R_t$ from the current robot position $\bm{p}_i$ is discarded. 

The \emph{multi-robot traversable map} ${\cal M}_t$ is obtained from the current map by suitably thresholding the function $w_{Cl}(\cdot)$ and collecting the resulting points along with their traversability cost (see Fig.~\ref{Fig:MultiRobotTraversability}).

It is worth noting that the multi-robot traversability allows the implementation of a prioritized path planning which takes into account prospective robot interactions~\cite{Lavalle-2006}. Planning priorities are implicitly assigned to teammates according to the time order in which their planned paths are received and integrated in the robot traversability map ${\cal M}_t$. In this process, the balls ${\cal B}(\bm{p}_j,R_c)$ and ${\cal B}(\bm{p}_i,R_t)$ are used in order to locally bound the coordination on the traversable map. 

It should be emphasized that, in case of strong communication delays, the sole knowledge of teammates' positions cannot be used to attain a safe robot navigation. In such a case, the multi-robot traversability (with its integrated knowledge of the teammates prospective paths) allows to attain metric coordination by \emph{(i)} minimizing interferences and \emph{(ii)} safely steering each robot ahead of time towards its goal. Moreover, given the fact that robots ``reserve" their motion space (by concurrently laying down prospective paths over the multi-robot traversability), node conflicts are often prevented.

\subsection{Path Planning and Windowed Search Strategy}\label{Sect:PathPlanningWindowedSearchStrategy}

%Path planning is performed both on global and local scales. 
For implementation and efficiency reasons we make use of a global and a local path planners. 
Given a set of 3D waypoints as input, the \textit{global} path planner is in charge of \emph{a)} checking the existence of a traversable path joining them and \emph{b)} minimizing a \emph{mixed cost function} along the computed path (see Sect.~\ref{Sect:MixedCostFunction}). This mixed cost function combines together the multi-robot traversability cost (see Sect.~\ref{Sect:TravCost}) along with an optional task dependent cost function. 

Once a global path solution $\bm{\tau}_g$ is found, the \textit{local} path planner \emph{continuously} replans a traversable path $\bm{\tau}_l$ that safely drives the robot from its current configuration $\bm{q}$ to the first configuration of $\bm{\tau}_g$ that intersects a sphere of radius $R_l$ centred at $\bm{q}$. This allows the path planner to more readily react to possible dynamic changes in the environment.

Both the global and the local path planners capture the connectivity of the configuration space ${\cal C}$ by using a sampling-based approach. The path search is restricted to a ``projection" of $\cal C$ on a 3D Euclidean space (Sect.~\ref{Sect:TravCost}). In fact, the path planner computes trajectories directly on the traversability map. %and restricts the path search to a projection of ${\cal C}$ onto a 3D Euclidean space 
%(see Section~\ref{Sect:TravCost}). 

A tree $\cal K$ is expanded on the traversability map ${\cal M}_t$ by using a randomized A* approach~\cite{Ferri-2014,Diankov-2007}. The start node $\bm{n}_s \in {\cal M}_t$ and the goal node $\bm{n}_g \in {\cal M}_t$ are computed as the projections of the start and goal robot positions on ${\cal M}_t$. $\bm{n}_s$ is used as root in order to initialize $\cal K$. The tree expansion at the current node $\bm{n} \in {\cal M}_t$ proceeds as follows 
\begin{enumerate}
[noitemsep,topsep=0pt,parsep=0pt,partopsep=0pt]
\item The clearance $w_{Cl}$ is computed at the position corresponding to  $\bm{n}$ (see Sect.~\ref{Sect:TravCost})
\item A \emph{safety radius} $\delta_{n}$ at $\bm{n}$ is computed as the minimum between $w_{Cl}$ and a pre-fixed maximum robot step; 
\item A set $\cal V$ of neighbours is created by collecting all the points of the traversable map that fall in a ball of radius $\delta_{n}$ centred at the position of $\bm{n}$;
\item A subset of neighbours in $\cal V$ are randomly selected as new children of $\bm{n}$ by using a probability inversely proportional to the corresponding traversability cost (this biases the expansion towards more traversable regions);
\item The A* cost-to-go of each new child is computed by taking into account the mixed cost function presented in  Sect.~\ref{Sect:MixedCostFunction},  eq.~(\ref{eq:costfunction});
\item The computed A* cost-to-go is used for inserting with priority the new child in a search queue;
\item The element of the search queue with the minimum cost-to-go is selected as next node to expand.
\end{enumerate}
In this process, a kd-tree is used for fast nearest neighbour search. 
The algorithm ends when a child node is found close enough to the desired goal position.

In order to further improve the efficiency and the response time of both the local and global path planners, a \emph{windowed search strategy} has been implemented around the basic path planner. Let $\bm{p}_s\bm{p}_g$  be the Euclidean line segment joining the assigned start position $\bm{p}_s$ and the goal position $\bm{p}_g$. Each time the global/local path planner is called to compute a new path: 
\begin{enumerate}
[noitemsep,topsep=0pt,parsep=0pt,partopsep=0pt]
\item First, the path search is restricted in the subset of points of the traversable map ${\cal M}_t \cap {\cal R}_1$, where ${\cal R}_1 \subset \mathbb{R}^3$ is a box with medial axis containing $\bm{p}_s\bm{p}_g$. Roughly speaking, this region is shaped as a narrow corridor with a longitudinal axis aligned to $\bm{p}_s\bm{p}_g$.
\item If a path cannot be found within ${\cal R}_1$, then it is searched within a new region ${\cal R}_2$ which is built by suitably growing ${\cal R}_1$ along its axes of symmetry.
\item If the path search fails then this process is repeated by incrementally growing the search region until a pre-fixed number of attempts is reached.
\end{enumerate}
In order to preserve the probabilistic completeness of the basic path planning algorithm, the last attempt uses the full traversable map as search region. For sake of safety, the most updated traversability map is considered as input at each planning attempt. 

In this process, the different attempts allow the robot to process different world ``snapshots" over time, with the benefit of possibly finding a solution after an initial failed attempt (due to new occurring favourable conditions).

\subsection{Mixed Cost Function}\label{Sect:MixedCostFunction}

The randomized A* algorithm computes a sub-optimal\footnote{The sub-optimality of the solution is due to the used incremental sampling-based approach~\cite{Karaman-2010,Diankov-2007}.} path $\bm{\tau} = \{\bm{n}_t\}_{t=0}^N$ in the configuration space\footnote{As explained in Sect.~\ref{Sect:TravCost}, each point of ${\cal M}_t$ can be associated to a robot pose.} 
$\cal C$ by minimizing the total cost:
\begin{equation}
J(\bm{\tau}) = \sum_{t=1}^N c(\bm{n}_{t-1},\bm{n}_t)
\end{equation} 
where $\bm{n}_0$ and $\bm{n}_N$ are the start and the goal respectively, and  ${ \bm{n}_t \in \cal C}$. 
The cost-to-go function $c:{\cal C}\times{\cal C}\to \mathbb{R}$ combines together the traversability cost and an optional task dependent function\footnote{This can be used for instance to steer the robot toward regions where an estimated WIFI radio signal strength map returns higher values~\cite{caccamo2017rcamp}.}. In particular
\begin{equation}\label{eq:costfunction}
\begin{multlined}
c(\bm{n}_t,\bm{n}_{t+1}) = \Big(d(\bm{n}_t,\bm{n}_{t+1})  +  h(\bm{n}_{t+1}, \bm{n}_N) +  \\ 
+ \lambda_z\mid \bm{n}_{t+1}^z - \bm{n}_{t}^z \mid\Big)\omega_1(\bm{n}_{t+1}) \omega_2(\bm{n}_{t+1})
\end{multlined} 
\end{equation}
\begin{equation}
\omega_1(\bm{n}) = \lambda_t \frac{trav(\bm{n})-trav_{min}}{trav_{max}-trav_{min}+\epsilon} +1 
\end{equation} 
where $d:{\cal C}\times{\cal C}\to \mathbb{R}^+$  is a distance metric, $h:{\cal C}\times{\cal C}\to \mathbb{R}^+$ is a goal heuristic, $\bm{n}_t^z \in \mathbb{R}$ is the z-coordinate of the node $\bm{n}_t \in {\cal C}$, $\lambda_z  \in \mathbb{R}^+$ and $\lambda_t \in \mathbb{R}^+$ are positive scalar weights, $\omega_1 : {\cal C} \to \mathbb{R}^+$ is the normalized traversability function, $\epsilon \in \mathbb{R}^+$ is a small quantity which prevents division by zero and $\omega_2  : {\cal C} \to \mathbb{R}^+$ is a normalized task-dependent cost function. 
The first factor in eq.~(\ref{eq:costfunction}) sums together the distance metric, the A* heuristic function (usually the distance to the goal) and a weighted difference of the z-coordinates of the nodes. The other two factors $\omega_1$,  and $\omega_2$ represent a normalized traversability cost and a normalized task-dependent cost respectively, whose strengths can be trade-off by using the weight $\lambda_t$. Note that $\omega_i \geq 1$ for $i=1,2$. 
The normalized task dependent function $\omega_2$ is typically built with a structure very similar to $\omega_1$~\cite{caccamo2017rcamp}. 
%(by possibly including an additional scalar positive trade-off weight in $\omega_2$)).  

\begin{algorithm}[t]
 \DontPrintSemicolon
 \nonl\textbf{PathPlanning}(\emph{goal}, \emph{traversability\_map}, \emph{team\_model})\;
\nonl~\hspace{-0.5cm}\codecomment{find initial solution}\;
  $path \leftarrow \emptyset$, \, $is\_goal\_aborted\leftarrow false$\; 
 %\While{ path $== \emptyset$ {\rm \textbf{and} $l \leq l_{max}$} }
  \For{ $l=1$ {\rm \textbf{to}} $l_{max}$}  
 {
 	Update() \codecomment{asynchronous}\;
 	$path \leftarrow$ ComputePath(\emph{goal}, \emph{traversability\_map})\;
 	\eIf{ path $ \neq \emptyset$ }
 	{
 		{\rm \textbf{break}}\;
 	}
 	{
 		sleep for $T_{wait}$\;
 	}
 }% end while
% \nonl~\hspace{-0.5cm}\codecomment{broadcast status and possibly start moving}\;
%\eIf{ path $ \neq \emptyset$ }
%{
% 	broadcast \emph{path} and \emph{success}\;
% 	TrajectoryTracking(\emph{path})\;
%}
%{
%	broadcast \emph{failure}\;
%	wait for a new \emph{goal} and then go to line $1$\;
%} 
\nonl~\hspace{-0.5cm}\codecomment{move along the path and broadcast status}\;
\While{\hspace{-0.05cm}{\rm (\textbf{not}}~is\_goal\_reached{\rm )}~{\rm\textbf{and}}~{\rm(\textbf{not}} is\_goal\_aborted{\rm )}}
{
\eIf{ path $ \neq \emptyset$ }
{
 	broadcast \emph{path} and \emph{success}\;
 	TrajectoryTracking(\emph{path}) \codecomment{asynchronous}\;
}
{
	broadcast \emph{failure}\;
	{\rm\textbf{return}};
}
Update() \codecomment{asynchronous}\;
$path \leftarrow$ ComputePath(\emph{goal}, \emph{traversability\_map})\;
}
% 	\textbf{return};
 \caption{PathPlanning}
 \label{Alg:PathPlanning}
\end{algorithm}

\subsection{Coordinated Path Planning and Message Protocol}\label{Sect:CoordinatedPathPlanning}

The path planner continuously replans a path on the multi-robot traversability map in order to react to possible dynamic changes in the environment. In this process, it uses the most updated map, the knowledge of prospective teammates paths and the current sensory information. A pseudocode description is reported in Algorithm~\ref{Alg:PathPlanning}. The function PathPlanning is invoked by the path planner every time a new goal is received from the patrolling agent.

Specifically, when a new goal position is designated, the path planner first tries to compute an \emph{initial solution} (lines 2--9), up to a maximum number of attempts $l_{max}$ (set to 5 in our experiments). At each failed attempt, it waits for a pre-fixed time interval $T_{wait}$ (line 8), then it retries by using the most updated information (line 3, see Sect.~\ref{Sect:Update}). If after $l_{max}$ attempts an initial solution is not found, the path planner communicates its failure to the patrolling agent (line 16) and then waits for a new goal; otherwise, a solution is found and a success message is sent to the patrolling agent (line 13).

Once an initial solution is found, the robot starts moving toward its goal (line 14) along the computed path. In this process, the path planner continuously replans a new path by using the most updated information (lines 11--20). 
Since the environment is assumed to be dynamic and populated by moving robots, a path planning failure can be verified by the local path planner during its continuous replanning, even after an initial solution is found by the global path planner. In case of failure, the path planner communicates it to the patrolling agent and then a new goal is received (line 12--16, Algorithm~\ref{Alg:PatrolAgent}). 

The path planner is managed at the topological level by the patrolling agent, whose decisions \emph{(i)} support cooperation and coordination with teammates, and \emph{(ii)} allow to detect and manage deadlocks. In fact, the patrolling agent continuously checks the path planner status and, in case of critical conditions (see Section.~\ref{Sect:NextNodeSelection}), pre-empts its current task and reassigns it a new goal (lines 12--16, Algorithm~\ref{Alg:PatrolAgent}). In particular, if the path planning keeps on failing for more than a pre-fixed time interval $T_{pcr}$, we say that a \emph{critical path planning failure} is occurring. This can be provoked by a local minima trap, as discussed in Sect.~\ref{Sect:NextNodeSelection}. In this case or when a goal is aborted by the patrolling agent (line 12, Algorithm~\ref{Alg:PatrolAgent}), the variable $is\_goal\_aborted$ is set to true and the continuous re-planning loop (lines 11--21, Algorithm~\ref{Alg:PathPlanning}) is stopped. 

It is worth noting that, in the initial solution search, the basic wait-retry process allows the robot to process different world ``snapshots" over time.
In some situations, this works as a virtual traffic-light and it allows teammates to move, reach their goals and free the way. 
In general, this basic wait-retry process alone is not sufficient to avoid deadlocks. For instance, it is not able to resolve the conflict experienced by two robots moving in opposite directions (e.g. along a narrow
corridor) and reciprocally blocking their ways. Indeed, such a case defines a local minima trap for both robots (continuous path planning failures would be generated on both sides). 
In our approach, many ingredients are used to prevent such deadlocks: the structure of our patrolling agent, the topological and metric coordination (Sect.~\ref{Sect:TwoLevelCoordinationStrategy}), the continuous interaction between the patrolling agent and the path
planner. In particular, the ability to detect critical conditions (Sect.~\ref{Sect:NextNodeSelection}), node conflict resolutions (topological coordination) and the randomized selection strategy allow to escape from local minima traps (e.g. the
situation described above).

The path planner continuously publishes the following messages after each plan or re-plan step, as a feedback.
\begin{compactitem} 
\item The \emph{path planner status}: this message is sent to the patrolling agent in order to inform it if a solution path was found (\emph{success}) or not (\emph{failure}), or if the assigned goal has been reached (\emph{reached}). A \emph{success} message also includes the navigation cost of the computed path.
%\item The \emph{path planning coordination message}:
\item The \emph{path message}: this is broadcast to teammates and contains the current estimated robot position and the current planned path (see Table~\ref{Tab:BroadcastMessages}). These data are essential for computing the multi-robot traversability. 
\end{compactitem}
On the other hand, the path planner can receive command messages from the patrolling agent. In particular, a \emph{command message} contains the current goal node position along with the desired action: \emph{go} or \emph{abort}. 

%==================================================================
%==================================================================

\section{3D Mapping and Localization}\label{Sect:3DMap}
%==================================================================
%==================================================================
%\input{3D_slam.tex}

\begin{figure}
\centering
\includegraphics[width=0.95\columnwidth]{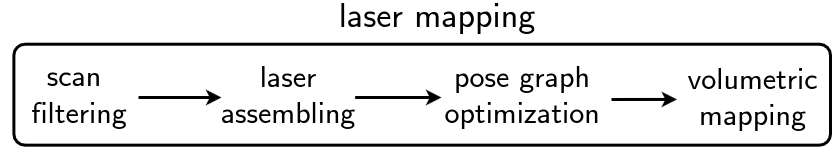}
\caption{The 3D SLAM pipeline.}
\label{Fig:SLAMpipeline}
\end{figure}

\begin{figure}
\centering
\includegraphics[width=1.0\columnwidth]{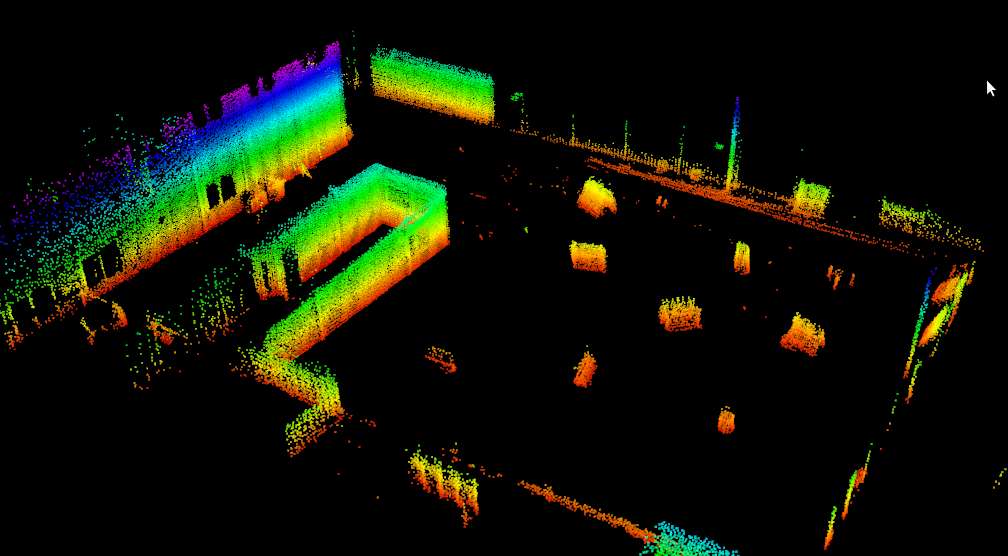}
\caption{A 3D map generated prior to the patrolling experiment which took place at the Deltalinqs training site, Rotterdam. The point cloud is colored by height and the ground plane has been removed for facilitating localization.}
\label{Fig:deltalinqs_map}
\end{figure}

In order to apply the distributed patrolling technique introduced in Section~\ref{Sect:DistributedPatrollingStrategies}, the robots need to localize in a common global reference when moving in the environment.
This multi-robot localization is performed against a 3D map which is built prior to the patrolling mission.
This map is also used for generating the initial patrolling graph presented in Section ~\ref{Sect:ProblemSetup}.
In the present system, the prior map and the individual maps of each robot, are built using the pose-graph \ac{SLAM} pipeline depicted in Figure~\ref{Fig:SLAMpipeline}.
For the experiments presented in Section~\ref{Sect:Experiments}, the maps are generated using the observations from a rotating 2D LiDAR sensor.
However, our system is flexible and accepts LiDAR sensors which directly provide 3D information.

Once the prior map has been generated, it is uploaded to each robot participating in the patrolling mission.
The multiple robots globally localize themselves using a place recognition strategy based on 3D segment extraction matching~\cite{dube2017segmatch}.
During the mission, each robot is responsible of (1) communicating to the other robots its location with respect to the prior map and (2) updating its local 3D volumetric representation of the environment to reflect dynamic changes.
The multi-robot localization solution detailed in the present section is inspired from earlier work (\cite{dube2017segmatch} and \cite{dube2017multirobot}) and has been adapted and integrated for fulfilling the needs of our patrolling framework.

In the remaining of this section we describe in more detail the \ac{SLAM} approach used, the chosen map representation, and the multi-robot localization on the prior map. 

\begin{figure}[!t]
\begin{center}
\subfloat[Positional error\label{subfig-1:lidar}]
{\includegraphics[width=1\linewidth]{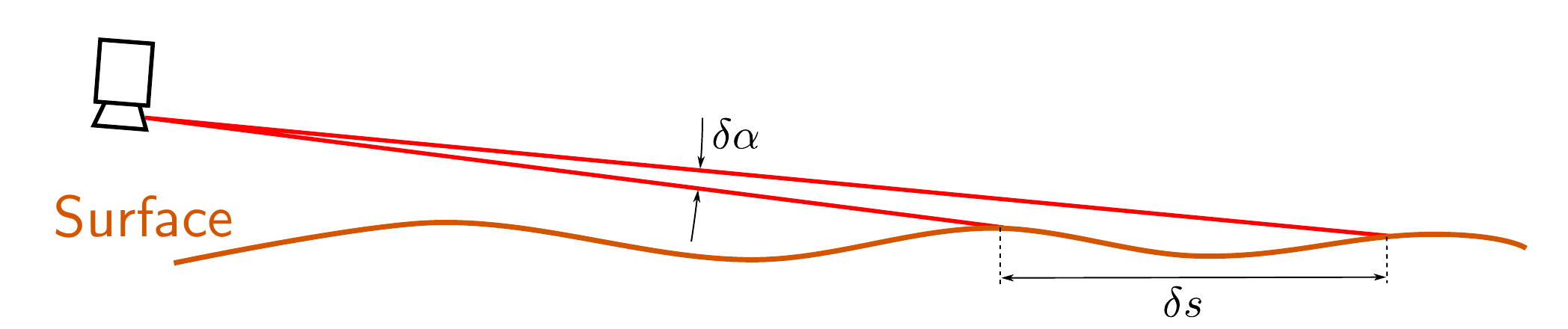}}\,
\subfloat[\textit{OctoMap} error\label{subfig-2:octo}]{\includegraphics[width=1\linewidth]{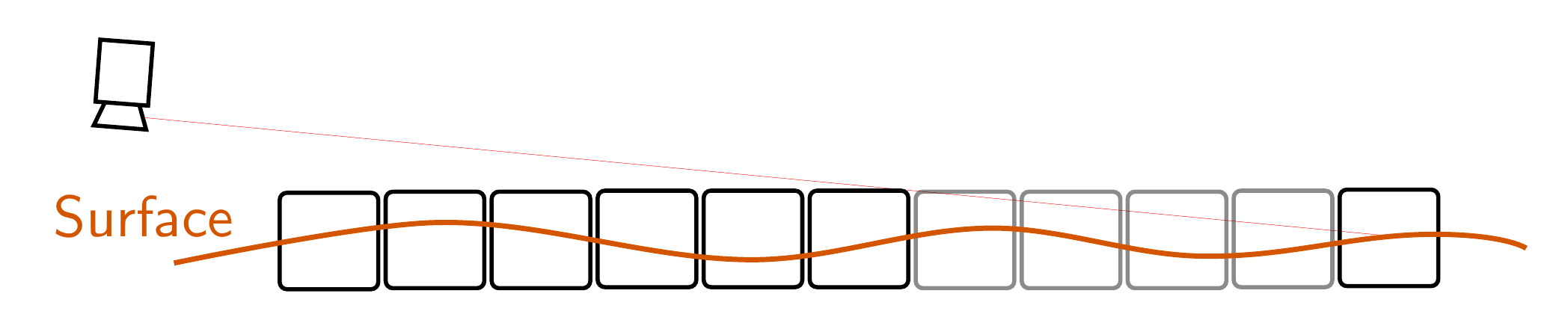}}
\caption{Challenges of small incidence angles using lidar: (a)   Small variations in the angle ($\delta \alpha$) inflict large positional uncertainty ($\delta s$). (b) Low incident angles inflict voxels falsely set as free (grey) in the \textit{Octomap}.}
\label{fig:small_angle_problems}
\end{center}
\end{figure}

\subsection{3D Pose-Graph SLAM}\label{Sect:PoseGraph}
In order to generate the prior map and to perform persistent \ac{SLAM} on each robot, the \ac{SLAM} system relies on a pose-graph optimization back-end~\cite{grisetti2010tutorial}. 
The states of our framework are robot poses $\boldsymbol{c}(t_i) {\in} SE(3)$ collected at times $\{t_i\}^N_{i=0}$.
%\footnote{The Special Euclidean group $SE(3)$ forms the mathematical basis of rigid transformations considered in this work.}.
%
These are estimated by optimizing a negative log-posterior $E$, an error function that  sums over a series of constraints $\boldsymbol{\Theta}(\boldsymbol{c}_{i,j}) {=} \boldsymbol{e}_{i,j}^T \boldsymbol{\Omega}_{i,j} \boldsymbol{e}_{i,j}$.
Here, $\boldsymbol{e}_{i,j}$ defines the error between the predicted state $\boldsymbol{z}_{i,j}$ and the observed state $\boldsymbol{\tilde z}_{i,j}$ of the system, i.e., $\boldsymbol{e}_{i,j}=\boldsymbol{z}_{i,j} - \boldsymbol{\tilde z}_{i,j}$, and $\boldsymbol{\Omega}_{i,j}$ the information matrix.
The \ac{SLAM} framework implements three different types of constraint that are summed up in $E$:
\begin{itemize}[noitemsep,topsep=0pt,parsep=0pt,partopsep=0pt]
\item prior constraints $\boldsymbol{\Theta}_P(\boldsymbol{c}_{i})$,
\item odometry constraints $\boldsymbol{\Theta}_O(\boldsymbol{c}_{i,j})$, and
\item scan-matching constraints $\boldsymbol{\Theta}_S(\boldsymbol{c}_{i,j})$.
\end{itemize}

Prior constraints can be created by using global localization information as described in Section~\ref{Sect:Localization}.% or interactively by manually setting the reference frame on the first robot pose $\boldsymbol{c}_{0}$.
Secondly, odometry constraints define pose displacements of consecutive robot locations by fusing IMU and wheel odometry measurements using an Extended Kalman Filter as described in~\cite{kubelka2015robust}.
Scan-matching constraints are finally obtained using \ac{ICP} to match the current scan against all previous scans within a sliding time window $[t-w,t] {\subset} \mathbb{R}$ where $t$ is the current time and $w$ is the chosen fixed time window.
The output of the \ac{ICP} algorithm is a set of rigid transformations which can directly be translated into pose-graph constraints.

Let $\boldsymbol{c}(t_1{:}t_2)$ be the sequence of robot poses acquired in the time interval $\left[t_1,t_2\right] {\subset} \mathbb{R}$. Denote by $\mathcal{C}_O$ and $\mathcal{C}_S$ respectively the set of pairs of timestamps for which odometry and scan-matching constraints exist over the same time interval $\left[t_1,t_2\right]$. 
The error function is then defined as
%
%\begin{equation}
%E(\boldsymbol{c}\left(t-w{:}t\right))=
%\boldsymbol{\Theta}_P(\boldsymbol{c}_{0})+ \sum_{<t_i,t_j> {\in} \mathcal{C}_O}{\boldsymbol{\Theta}_O(\boldsymbol{c}_{i,j})}+  \sum_{<t_i,t_j> {\in} \mathcal{C}_S}{\boldsymbol{\Theta}_S(\boldsymbol{c}_{i,j})}
%\end{equation}
\begin{equation}
\begin{multlined}
E(\boldsymbol{c}\left(t-w{:}t\right)) =
\boldsymbol{\Theta}_P(\boldsymbol{c}_{0})+ \sum_{\langle t_i,t_j \rangle {\in} \mathcal{C}_O}{\boldsymbol{\Theta}_O(\boldsymbol{c}_{i,j})}+ \\ + ~ \sum_{\langle t_i,t_j \rangle {\in} \mathcal{C}_S}{\boldsymbol{\Theta}_S(\boldsymbol{c}_{i,j})}
\end{multlined}
\end{equation}
on the sliding time window. 
This error function is finally minimized using the Gauss Newton algorithm and the robot trajectory is updated with the optimization result.

The pose-graph model therefore serves as an implicit estimation of the robot trajectory and map.
The latter can explicitly be generated, in the form of an \textit{OctoMap}, by projecting individual scans from the optimized robot poses into the global frame of reference.
An example of this 3D representation is illustrated in Figure~\ref{Fig:deltalinqs_map}.

%\begin{figure}
%\centering
%\includegraphics[width=1.0\columnwidth]{deltalinqs_map_cropped}
%\caption{A 3D map generated prior to the patrolling experiment which took place at the Deltalinqs training site, Rotterdam. The point cloud is colored by height and the ground plane has been removed for facilitating localization.}
%\label{Fig:deltalinqs_map}
%\end{figure}

\subsection{\textit{OctoMap} representation}\label{Sect:OctoMap}

We select the \textit{OctoMap}~\cite{Hornung-2013} representation for modelling occupied and free space explicitly.
The \textit{OctoMap} representation exhibits several advantageous properties for multi-robot applications.
This representation first allows to register mapping data from different sources in a common frame of reference, enabling the distributed patrolling strategy introduced in Section~\ref{Sect:DistributedPatrollingStrategies}.
Moreover, this probabilistic framework accounts for dynamic objects which can be filtered over multiple observations due to the explicit modelling of free space using \textit{ray-casting}.
The \textit{OctoMap} can be obtained by either loading an existing map and applying potential online extension, or building it online using our LiDAR-based \ac{SLAM} approach.

In order to use this representation for navigation and patrolling, a `clamping policy' is adopted by setting a lower and upper bound on the log-likelihood of the occupancy estimate in the \textit{OctoMap}.
The final decision about occupancy is made by thresholding this bounded estimate which ensures that the 3D map representation can quickly adapt to changes in the environment\footnote{The dynamic update of the OctoMap and its reactive behaviour is demonstrated in a video \href{https://youtu.be/caECYcYdrgo}{https://youtu.be/caECYcYdrgo}}.

For the \ac{UGV}s used in our experiments, the LiDARs are mounted at low heights which requires an adaptation over the classic \textit{OctoMap} approach.
As displayed in Fig.~\ref{fig:small_angle_problems}, the motivation behind this adaptation is that a low angle of incidence relative to the ground may cause voxels to be falsely marked as free space which is in turn critical for the traversability analysis introduced in Section~\ref{Sect:MapSegmentation}--\ref{Sect:TravCost}.
We therefore limit the angle of incidence at which ray-casting can lower the occupancy probability of voxels to a lower bound $\alpha_{min}$.

The center-points of occupied \textit{OctoMap} cells are thus used for traversability analysis as shown in Section~\ref{Sect:TravCost}.

\subsection{Multi-robot localization}\label{Sect:Localization}
\begin{figure}
\centering
\includegraphics[width=1.0\columnwidth]{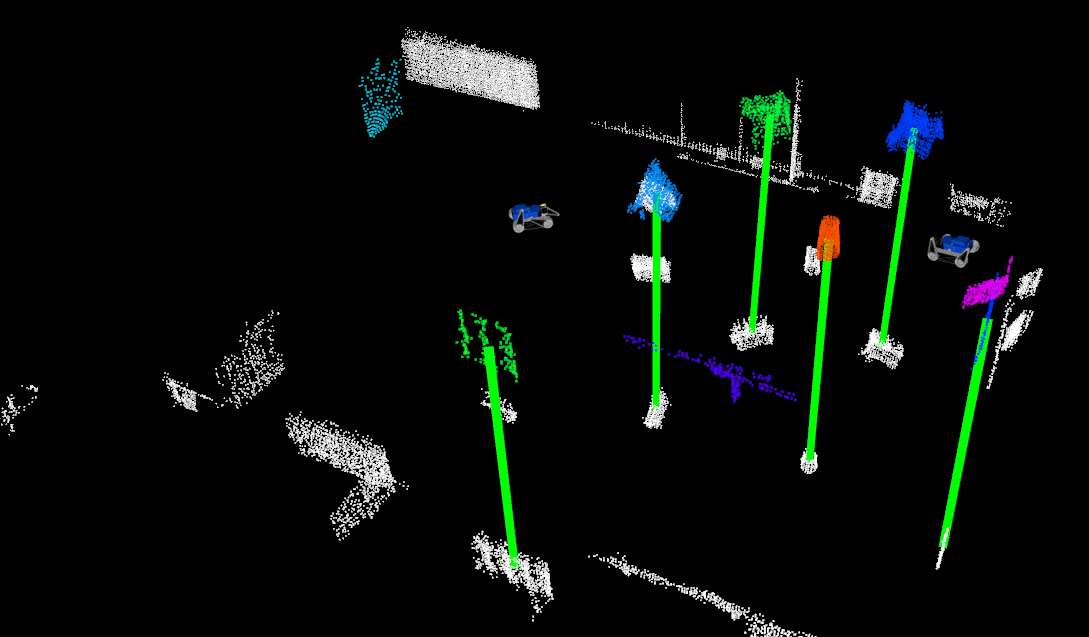}
\caption{A localization example in the map illustrated in Fig.~\ref{Fig:deltalinqs_map}. 
Segments extracted from the target map are shown in white below whereas colors are used to depict segments extracted from the local representation of the robot located at the right. 
Matching segments resulting in a localization are illustrated with vertical green lines.}
\label{Fig:deltalinqs_localization}
\end{figure}

At the beginning of a patrolling mission, the global location of each robot is estimated using the \textit{SegMatch} algorithm~\cite{dube2017segmatch}.
Specifically, 3D point cloud segments are extracted from the prior map and all local maps by applying ground-plane removal, followed by Euclidean clustering with a growing distance $d$ ~\cite{douillard2011segmentation}.
Eigen-value based features are then extracted in order to uniquely describe each segment~\cite{weinmann2014semantic}.
Candidate segment matches are identified between each local map and the target map by considering the $k$ nearest neighbours in feature space.
An $SE(3)$ transformation is finally obtained for each robot by selecting the largest group of consistent candidates using RANSAC with a resolution $r$.
Figure~\ref{Fig:deltalinqs_localization} illustrates a localization example with the consistent group of matches depicted with green vertical lines.
The paremeters used in this algorithm throughout the experiments are presented in Table~\ref{Tab:SlamParameters}.

Each robot uses this localization information for initializing its own SLAM algorithm, as presented in Section~\ref{Sect:PoseGraph}.
Given that an unique prior map is shared amongst all robots, scan-matching factors $\boldsymbol{\Theta}_S$ are generated by performing \ac{ICP} against this shared map.
Thus, ensuring that the multiple robots are globally localized in real-time and in a common reference frame, enabling the multi-robot patrolling technique presented in this work.
This localization paradigm is able to account for changes in the environment, if a sufficient amount of structure is similar, enabling \ac{ICP} to converge to correct solutions.

%==================================================================
%==================================================================

\section{Patrolling Graph Building}\label{Sect:PatrollingGraphBuilding}
%==================================================================
%==================================================================
%\input{patrolling_graph_building.tex}

\begin{figure}[t]
\begin{center}
     \subfloat[Radius search\label{subfig-1:rs}]{%
       \includegraphics[height=0.3\columnwidth]{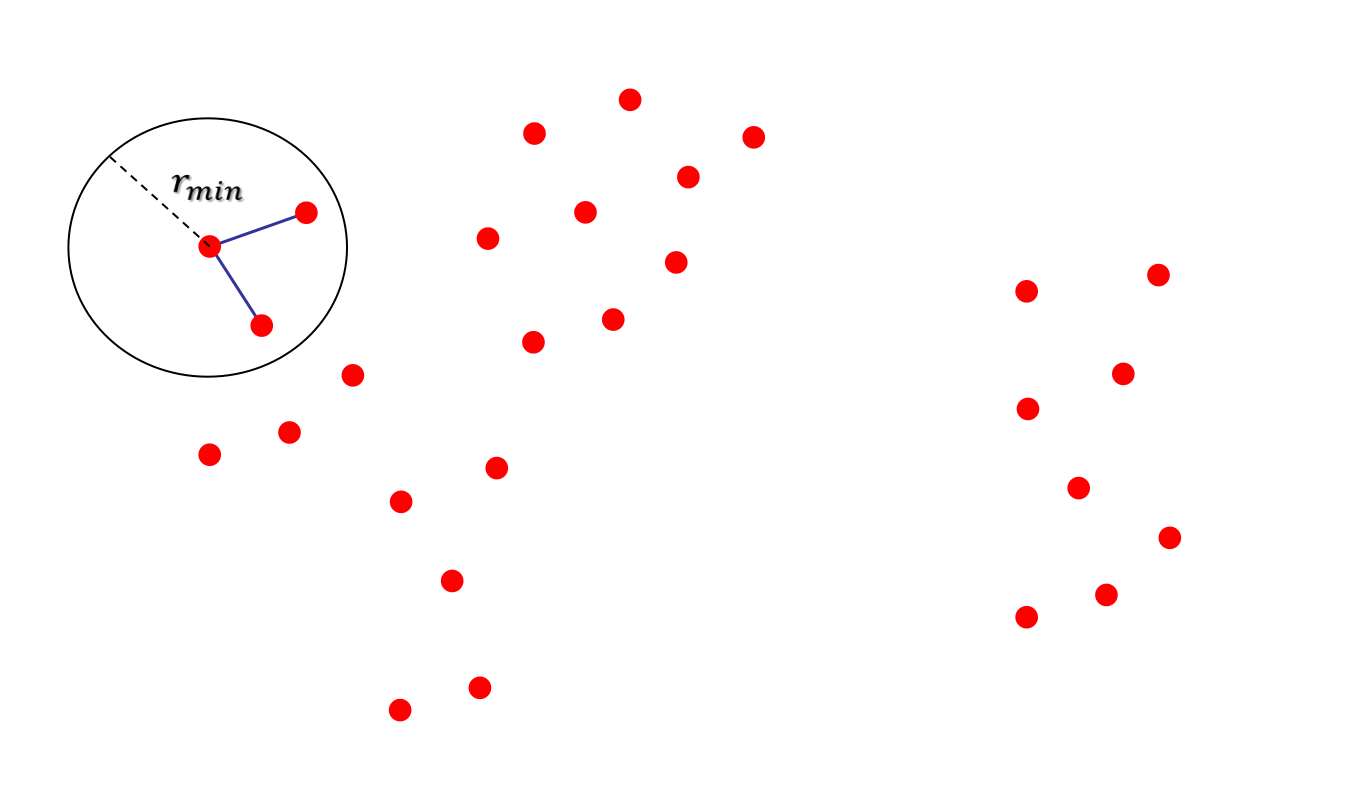}
     }
%     \,
     \subfloat[Connected Components\label{subfig-2:cc}]{%
       \includegraphics[height=0.3\columnwidth]{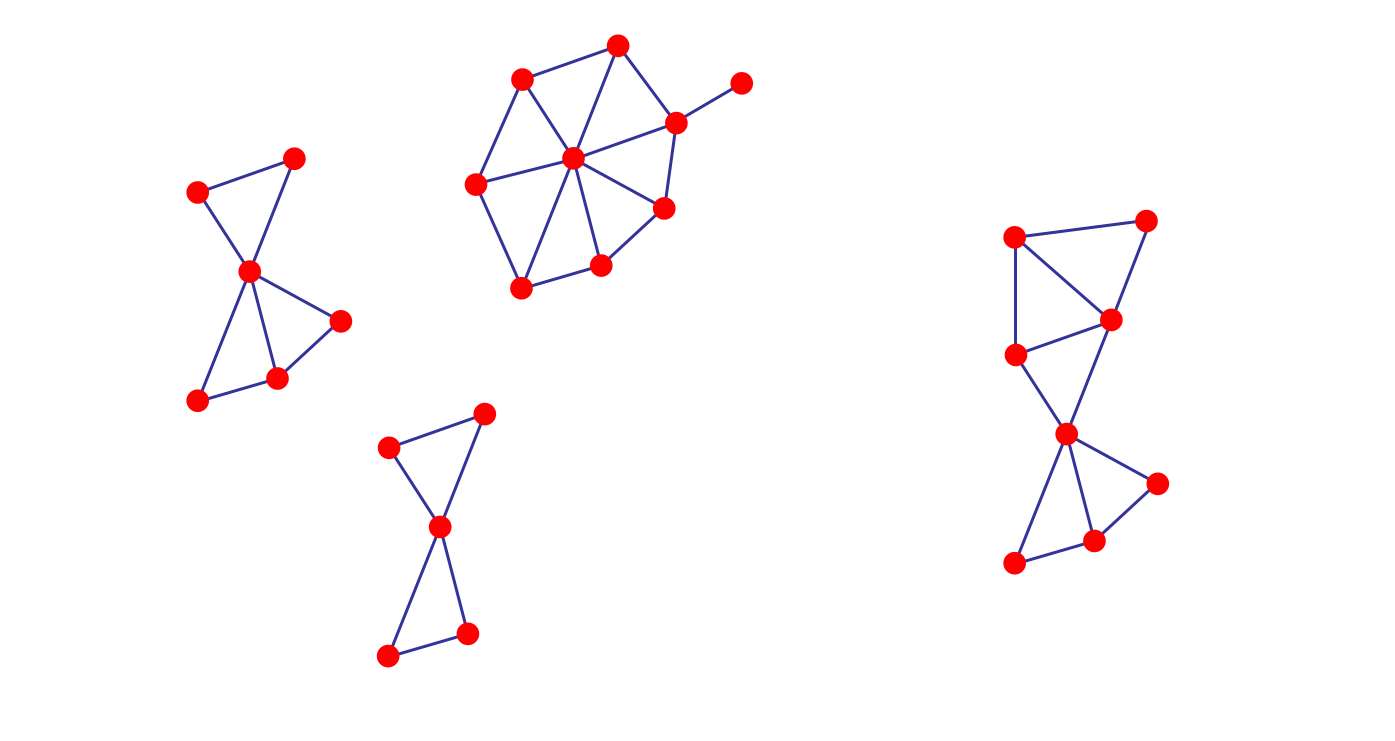}
     }
     \\
     \subfloat[Iterative search with adaptive radius\label{subfig-3:irs}]{%
       \includegraphics[height=0.3\columnwidth]{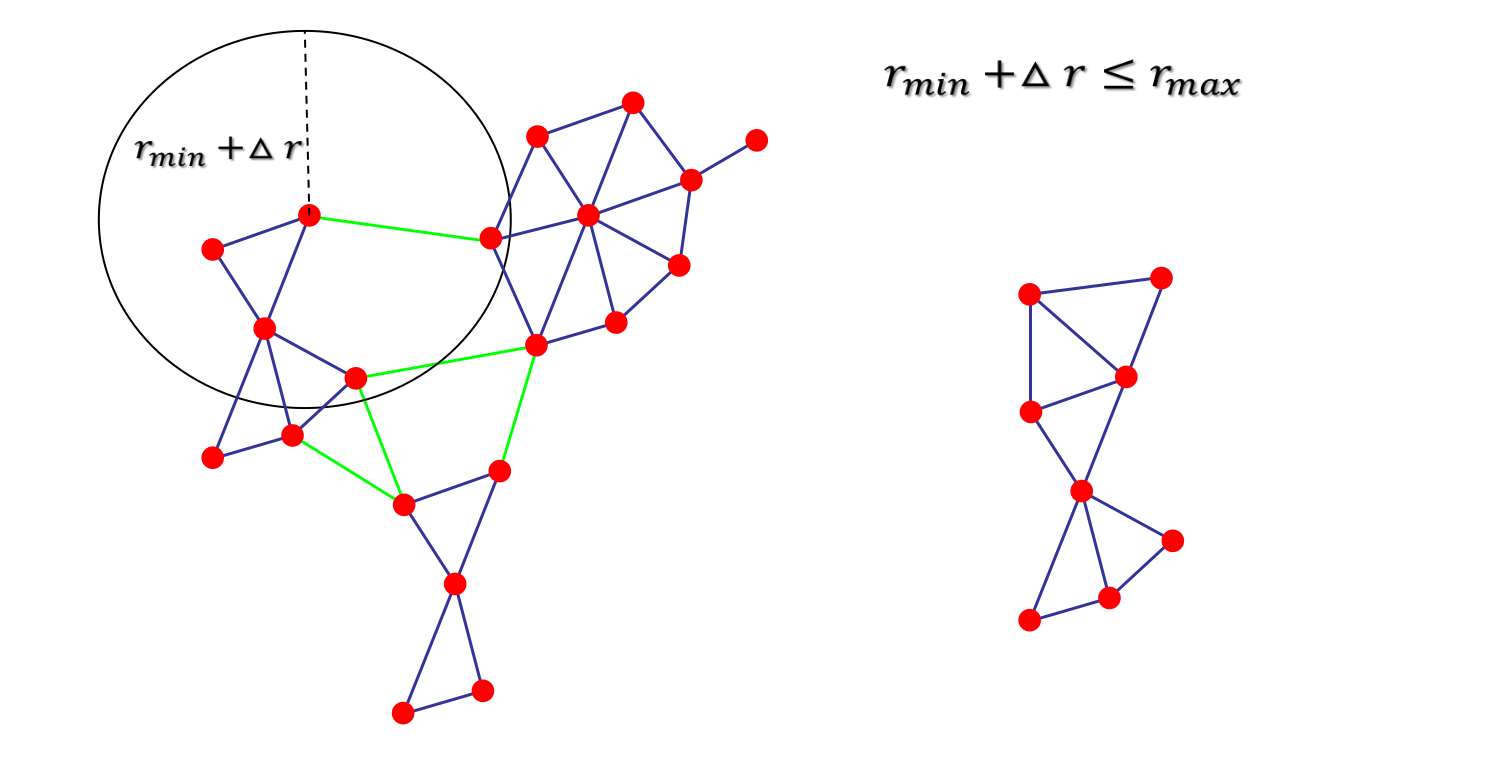}
     }
     \caption{Main steps of the automatic procedure for building a graph: this is used for processing an history of robot trajectories.}
     \label{fig:dummy}
\end{center}
\end{figure}
 
This section briefly presents two procedures for building a patrolling graph: the first (interactive) takes as input a set of \ac{POIs} selected by the user on the 3D interface; the second (automatic) automatically computes the patrolling graph from an  history of robot trajectories.

\subsection{Patrolling Graph from a User-assigned Set of Waypoints}
%The 3D GUI allows the user to load a saved map and to interactively build a patrolling graph on the corresponding traversable map. 
In the interactive procedure, a set of \ac{POIs} (or waypoints) are selected by the user on the map. These are potentially considered as patrolling graph nodes.
%First, the user selects a sequence of points of interest on the traversable map and sends this as input to the patrolling graph builder (see Fig.~\ref{Fig:Architecture}). 
Then, an algorithm automatically adds an edge between each pair of nodes $(n_i,n_j)$ that satisfy the following conditions:
\noindent
{
\begin{enumerate}[noitemsep,topsep=0pt,parsep=0pt,partopsep=0pt]
\itemsep0em 
\item the Euclidean distance between the corresponding points $\bm{p}_i, \bm{p}_j \in \mathbb{R}^3$ is smaller than a maximum distance $d_{max} \in \mathbb{R}$ (set to $5m$ in our experiments);
\item the line segment connecting $\bm{p}_i$ and $\bm{p}_j$ does not intersect the map;
\item the line segment between the positions $\bm{p}_i$ and $\bm{p}_j$ has an elevation angle smaller than a maximum angle $\alpha_{max} \in \mathbb{R}$ (we set this to $30^\circ$);
\item a traversable path between the node positions $\bm{p}_i, \bm{p}_j \in \mathbb{R}^3$ exists. 
\end{enumerate}
}
The first condition is added for containing the branch factor of each node and avoid too long travels between nodes. The second condition checks if the line segment $\bm{p}_i\bm{p}_j$ intersects the ground or an obstacle.
The second and third conditions together avoid connecting nodes which belong to different floor levels or which can be joined by a too steep passage. 
%in practice, these conditions also force the construction of a planar graph. 

If some of the points are not connected, they are not considered as nodes, the user can move or delete them, and then repeat the procedure. In this process, kd-trees are efficiently used in order to perform collision checking.

\subsection{Patrolling Graph from a Saved History of Robot Trajectories}

The automatic graph building procedure is based on the approach presented in \cite{Menna-2014}. 
First, each input robot trajectory is initially discretized via uniform sampling, in order to obtain a sparse sequence of poses.
Then, each resulting sequence is accumulated in a suitable space-partitioning data structure, where the robot orientation is disregarded.
Next, a voxel grid filter is applied to this data structure to reduce the number of points stored therein.

For each resulting point in the filtered data structure a node is generated. 
Connections among nodes are established as follows.
A preliminary procedure is applied to the filtered data structure to find a set of distinct connected components (see Figure~\ref{subfig-2:cc}). 
This procedure searches for all the nearest neighbours of a query point in a given radius (see Figure~\ref{subfig-1:rs}).
Finally connected components are linked together through an iterative radius search procedure, where at each iteration, the value of the radius is incremented in order to ensure connectivity (see Figure~\ref{subfig-3:irs}).  

%==================================================================
%==================================================================

\section{Results}\label{Sect:Results}
%==================================================================
%==================================================================
%\input{results.tex}

\begin{figure}[t!]
\centering
	\includegraphics[width=\linewidth]{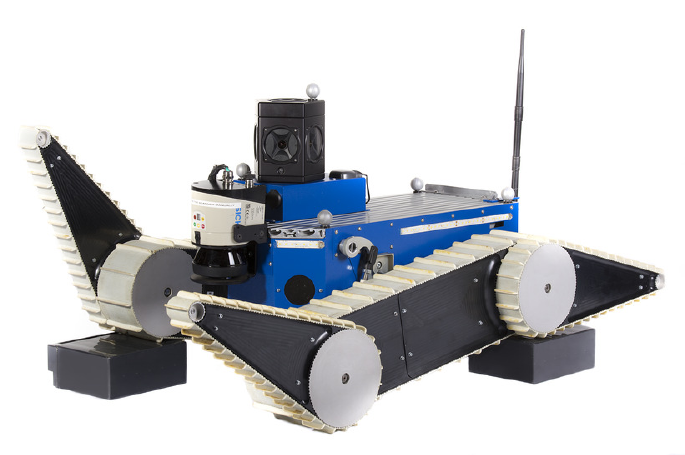}
    \caption{TRADR UGV equipped with multiple encoders, an IMU and a rotating laser-scanner.}%
    \label{Fig:Robot}
\end{figure}   

\begin{figure*}[ht]
\centering
\includegraphics[width=0.8\linewidth]{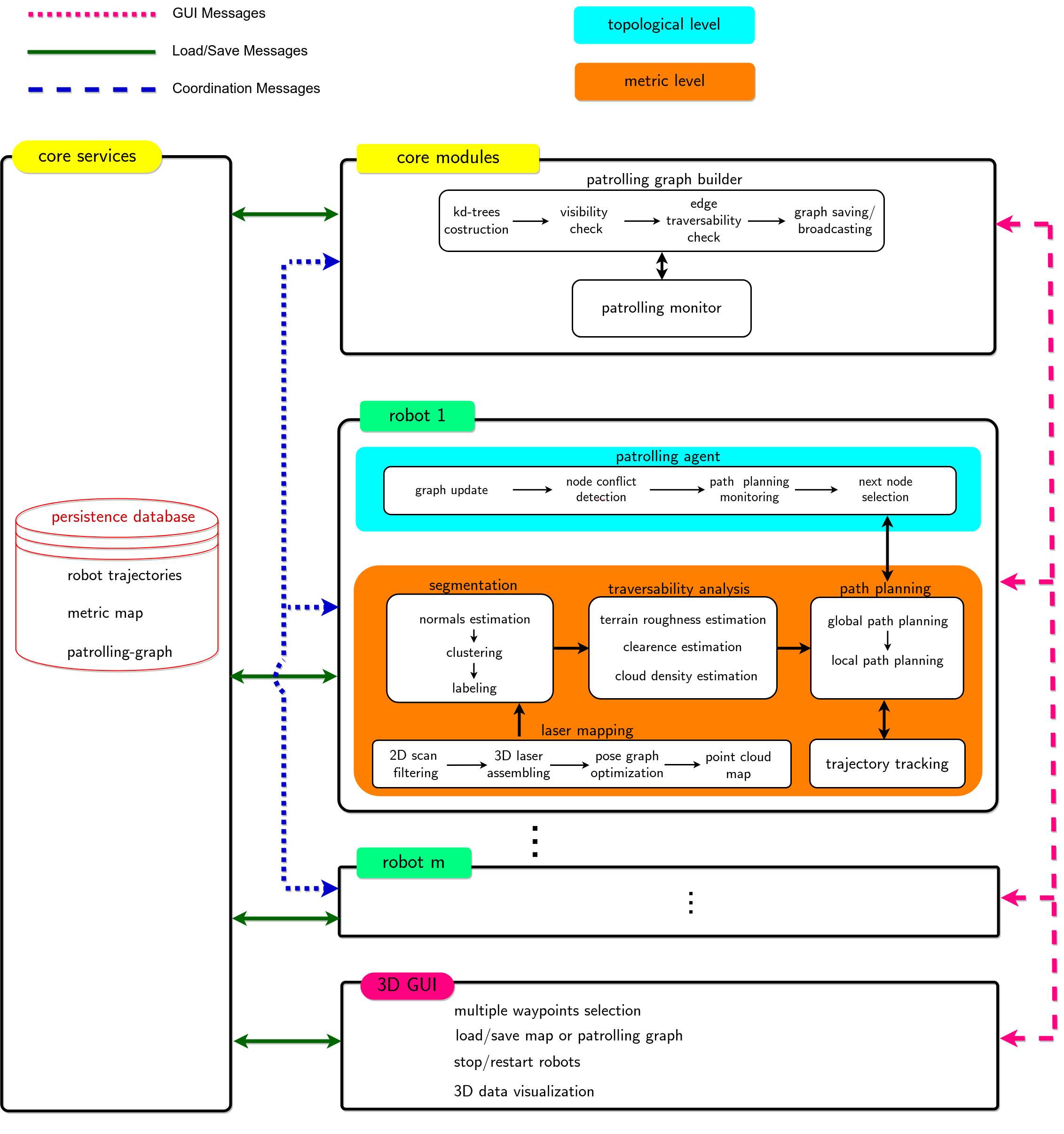}
\caption{A functional diagram of the implemented multi-robot system. Robots share the same internal software architecture. In particular, each robot hosts an instance of the patrolling agent and of the path-planner. The legend on the top left represents the different kind of exchanged messages. The architecture is detailed in Sect.~\ref{Sect:FunctionalArchitecture}.}
\label{Fig:Architecture}
\end{figure*}

This section presents the results we obtained with an implementation in 3D. %Afterwards, we present the simulation results obtained in V-REP, which still live in 3D but are analysed by focusing more on 2D decision-making and patrolling aspects treated by other works in the literature. 
We validated the proposed strategy on the TRADR UGV robots~\cite{Kruijff-2015} (cfr. Fig.~\ref{Fig:Robot}), both in simulations and real-wold experiments. 
These vehicles are skid-steered and satisfy the path controllability assumption (see Sect.~\ref{Sect:3DEnvModel}). Amongst other sensors, the robots are equipped with a $360^\circ$ spherical camera and a rotating laser scanner. 

We considered 3D scenarios which are typical for our TRADR UGVs (see Sect.~\ref{Sect:Intro}). Here, interferences are very likely and the UGVs need to navigate by \emph{(i)} avoiding conflicts in narrow passages, \emph{(ii)} performing reliable traversability analysis and coordinated path-planning, \emph{(iii)} reliably localizing in 3D while simultaneously updating and extending the input 3D metric map. In these scenarios, there is typically an high ratio between team size and patrolling graph size.

%\subsection{List of Used Parameters}\label{Sect:UsedParameters}
%==================================================================
%==================================================================
%\input{used_parameters.tex}

For convenience, we report in Tables\ref{Tab:PlanningParameters}--\ref{Tab:SlamParameters} the list of the main parameter values we used both in simulations and experiments. 

\begin{table*}
\begin{center} 
\caption{Path planning and patrolling agent main parameters used in the evaluation.}
\label{Tab:PlanningParameters}
%\resizebox{0.8\linewidth}{!}{
\renewcommand{\arraystretch}{1.1}
\begin{tabular}{llll}
Description & Symbol & Value \\
\cline{1-3}
Robot max linear velocity speed & $v_{max}$ & $0.2m/s$ & \rdelim\}{6}{3mm}[Path Planning]\\
Robot bounding radius & $R_b$& $0.47m$\\
Robot safety distance & $D_s$ & $1.2m$ \\
Future trail crop radius & $R_c$ & $1.5m$ \\
Radius for considering future trails & $R_t$ & $1.5m$ \\
Path planning waiting time & $T_{wait}$ & $0.5s$ \\
Critical path planning failure time & $T_{pcr}$ & $5s$  & \rdelim\}{6}{3mm}[Patrolling] \\
Critical node conflict time & $T_{ncr}$ & $5s$\\
Patrolling sleep time & $T_{sleep}$& $0.1s$ \\
Patrolling main loop rate & $f_{patrol}$& $30Hz$ \\
Idleness message broadcast period & $T_{idln}$ & $5s$ \\
Team model expiration time & $T_{exp}$ & $10s$ \\
\cline{1-3}
\end{tabular}
%}
\end{center}
\end{table*}

\begin{table*}
\begin{center} 
\caption{Laser mapping parameters used in the evaluation.}
\label{Tab:SlamParameters}
\renewcommand{\arraystretch}{1.1}
%\resizebox{0.8\linewidth}{!}{
\begin{tabular}{llll}
Description & Symbol & Value \\
\cline{1-3}
Maximum laser range & $r_{max}$ & $20m$ & \rdelim\}{8}{3mm}[3D SLAM]\\
Scan maximum density & $\rho_{max}$& $50000 \frac{1}{m^3}$\\
Scans in Sliding window estimation & $n_{scans,SWE}$ & $3$ \\
knn surface normal computation & $n_{knn}$ & 20 \\
\ac{ICP} error metric & point-to-plane &\\
Prior noise model & $\boldsymbol{\Omega}_P$ & $\boldsymbol{0}_{6x6}$ \\
Odometry noise model & $\boldsymbol{\Omega}_O$ & $(500, 500, 500, 500, 0.015, 500)^T \boldsymbol{I}_{6x6}$ \\
Scan matching noise model & $\boldsymbol{\Omega}_S$ & $(0.05, 0.05, 0.05, 0.015, 0.015, 0.015)^T \boldsymbol{I}_{6x6}$ \\
\textit{OctoMap} resolution & $\phi$& $0.075m$ & \rdelim\}{4}{3mm}[\textit{OctoMap}] \\
\textit{OctoMap} occupancy thresholds & $o_{min}, o_{max}$ & $0.12, 0.97$\\
\textit{OctoMap} hit / miss probabilities & $P_{hit}, P_{miss}$& $0.75, 0.2$\\
\textit{OctoMap} min angle ground removal & $\alpha_{min}$ & $4$ degrees\\
Region growing distance & $d$ & $0.2m$ & \rdelim\}{3}{3mm}[\textit{SegMatch}] \\
Number of nearest neighbours & $k$ & $5$ \\
RANSAC resolution & $r$ & $0.3m$ \\
\cline{1-3}
\end{tabular}
%}
\end{center}
\end{table*}

%==================================================================
%==================================================================

All the algorithms are implemented in C++ (cfr. Sect.~\ref{Sect:CodeImplementation}). ROS is used as middleware. The code has been designed to seamlessly interface with both simulated and real robots. This allows to use the same code both in simulations and experiments.
An open source implementation is available\footnote{\href{https://gitlab.com/luigifreda/3dpatrolling}{ https://gitlab.com/luigifreda/3dpatrolling}.}.

A functional diagram of the presented multi-robot system is reported in Fig.~\ref{Fig:Architecture}. This is detailed in Sect.~\ref{Sect:FunctionalArchitecture}.

\subsection{Simulation Experiments}\label{Sect:Simulations}
%==================================================================
%==================================================================
%\input{simulations.tex}

%\begin{figure}[t!]
%\centering
%	\includegraphics[width=\linewidth]{nifti.eps}
%    \caption{TRADR UGV equipped with multiple encoders, an IMU and a rotating laser-scanner.}%
%    \label{Fig:Robot}
%\end{figure}   

This section presents simulation results obtained with the V-REP simulation framework~\cite{vrep}.  
V-REP allows to simulate laser range finder and odometry noise. Grousers have been added to the simulated robot tracks in order to obtain realistic robot interactions with the terrain. 
%In our view, such a framework allows to perform very realistic simulations.  

For convenience, we have adopted a single-core ROS architecture during our simulation runs. A different and more efficient network architecture is used for the real-world experiments (see Section~\ref{Sect:Experiments}). 
%As already mentioned (see Section~\ref{Sect:CodeImplementation}), the same code is used for both simulations and experiments.
In simulation, we introduced a fixed delay of $0.2s$ in the publishing of each broadcast message. 

In this work, since the focus is on patrolling  aspects, we do not consider the articulated tracks during motion planning\footnote{This aspect can be managed for instance as proposed in~\cite{Zimmermann-2014} or ~\cite{Colas-2013}}. 

We perform simulations with teams up to four \linebreak TRADR robots.
While this is a typical team size in the considered \ac{SaR} applications, it is mainly a limitation from the V-REP simulations which is computationally very demanding.
To face this limitation, our setup distributes the V-REP simulations, and the ROS nodes performing SLAM, segmentation, traversability analysis, path planning and patrolling on distinct computers.
However, in our setup, V-REP is not able to stably simulate more than four robots under realistic conditions.
On the other hand, the presented multi-robot patrolling strategy is fully distributed and the implementation of its coordination protocol does not require special hardware.

\newcommand{\gridsimimageheightfirst}{5.22cm}

\begin{figure*}[!t]
\begin{center}
\subfloat[Three-ways\label{SubFig:scenario:threeways:vrep}]
{
\includegraphics[height=\gridsimimageheightfirst,keepaspectratio]{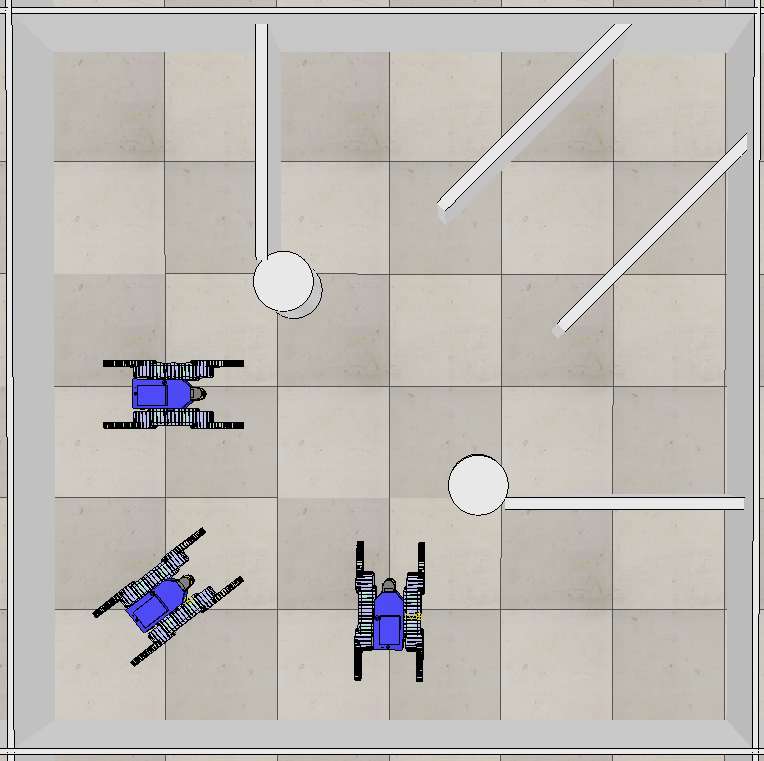}
}
\,
\subfloat[Cyclic paths\label{SubFig:scenario:threeways:paths}]
{
\includegraphics[height=\gridsimimageheightfirst,keepaspectratio]{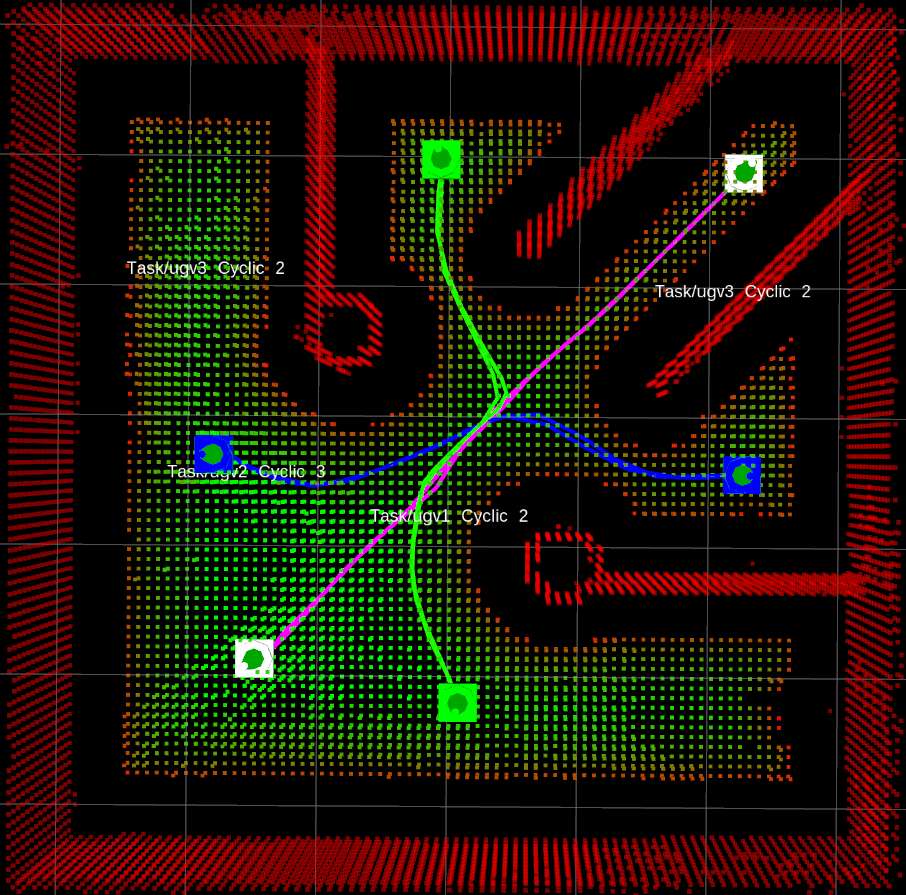}
}
\,
\subfloat[Maps\label{SubFig:scenario:threeways:maps}]
{
\includegraphics[height=\gridsimimageheightfirst,keepaspectratio]{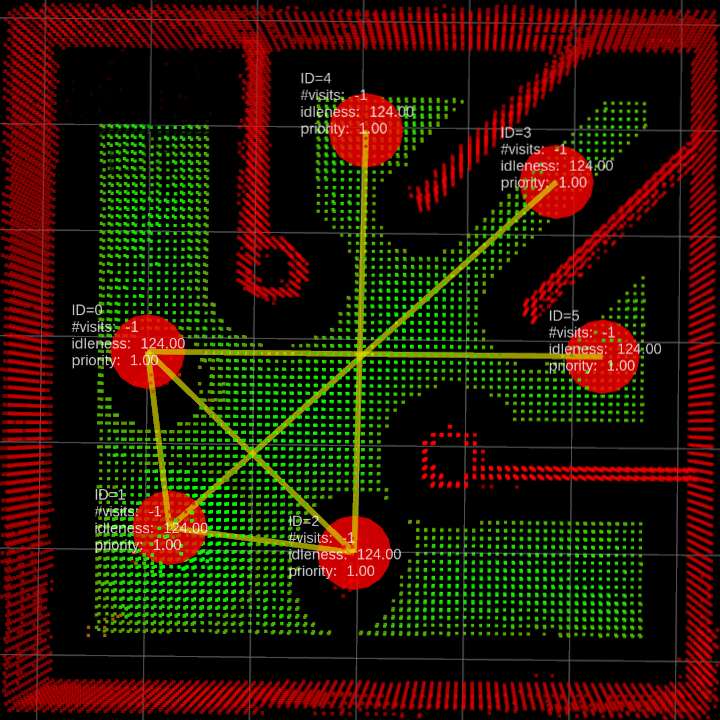}
}
\caption{\emph{Left}: the three-ways scenario in V-REP. \emph{Center}: the three cyclic paths assigned to the robots (in different colours). Each robot is required to move back and forth between its two assigned waypoints (mainly along the horizontal, diagonal or vertical direction). \emph{Right}: the environment maps, i.e. patrolling graph (red circular vertex and yellow edges), traversable regions (green point cloud), obstacle regions (red point cloud).}
\label{Fig:PlanarScenarioThreeWays}
\end{center}
\end{figure*}

\begin{figure*}[!t]
\begin{center}
\subfloat[Multi-floor ramp \label{SubFig:scenario:3d_ramp}]
{
\includegraphics[height=7.28cm, keepaspectratio]{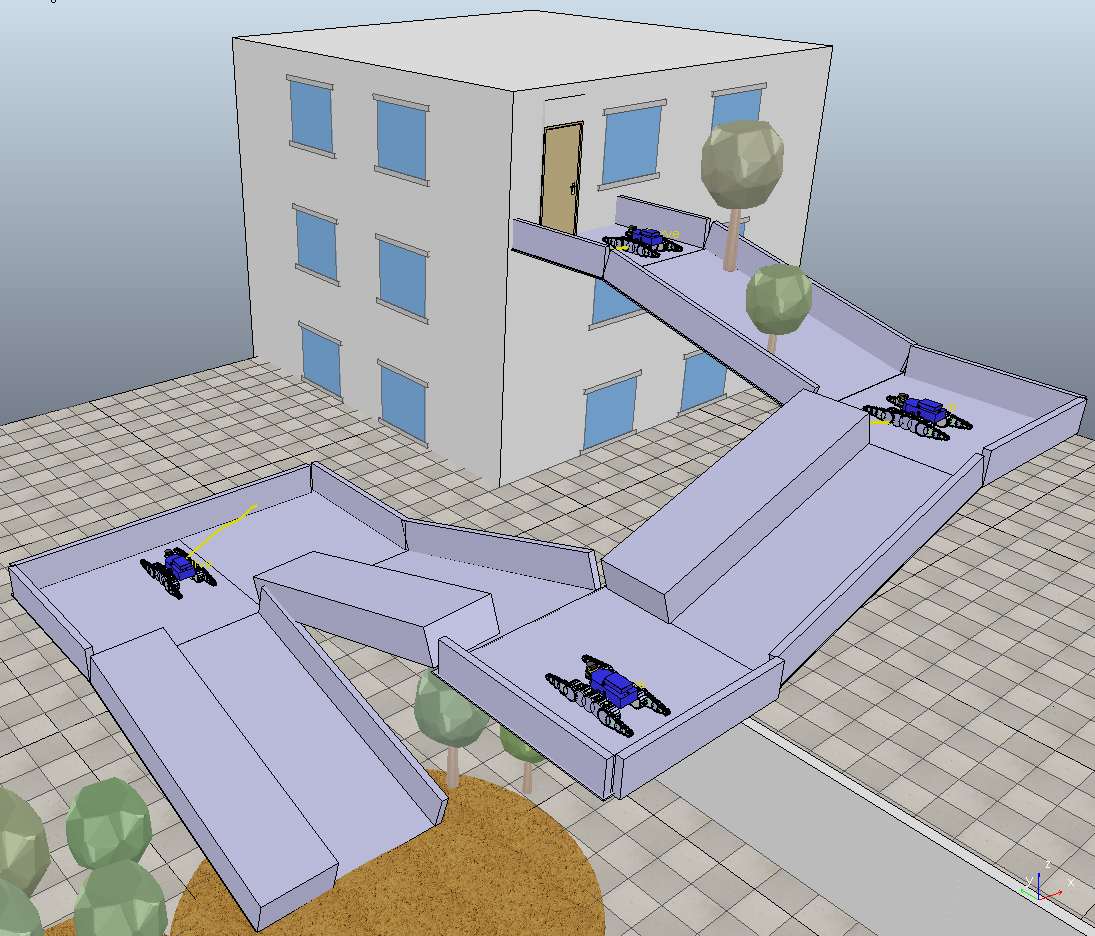}
\,
\includegraphics[height=7.28cm, keepaspectratio]{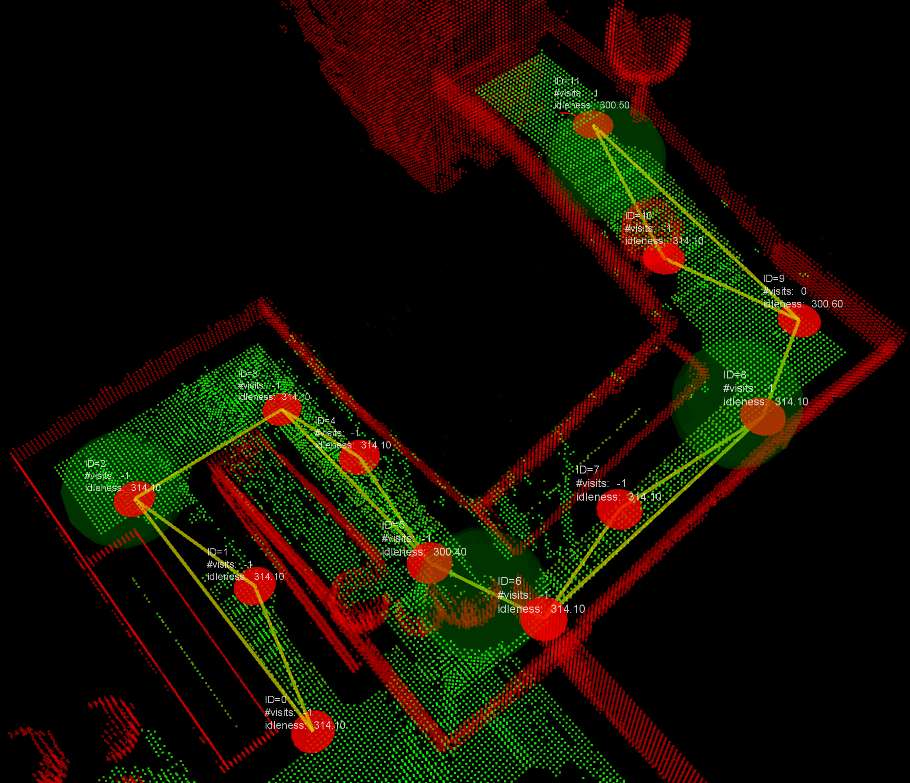}
}
\,
\subfloat[Two-floor ring \label{SubFig:scenario:3d_ring}]
{
\includegraphics[height=4.55cm,keepaspectratio]{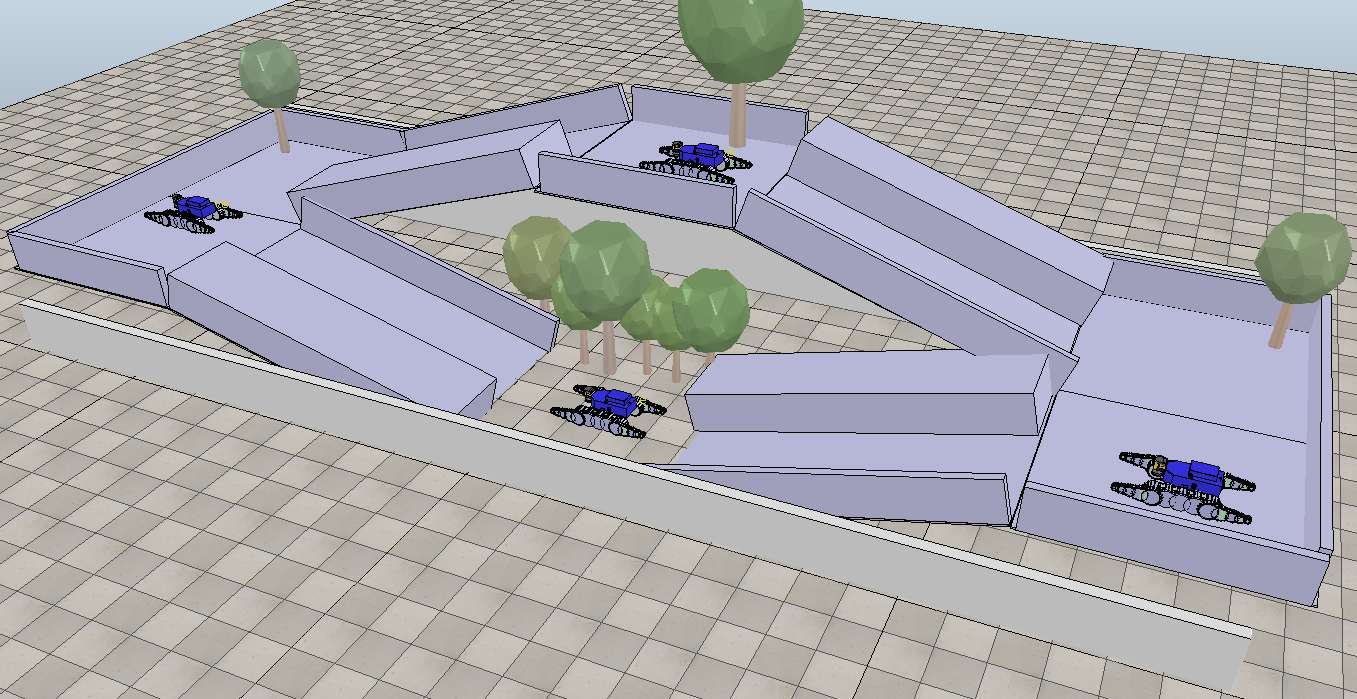}
\,
\includegraphics[height=4.55cm,keepaspectratio]{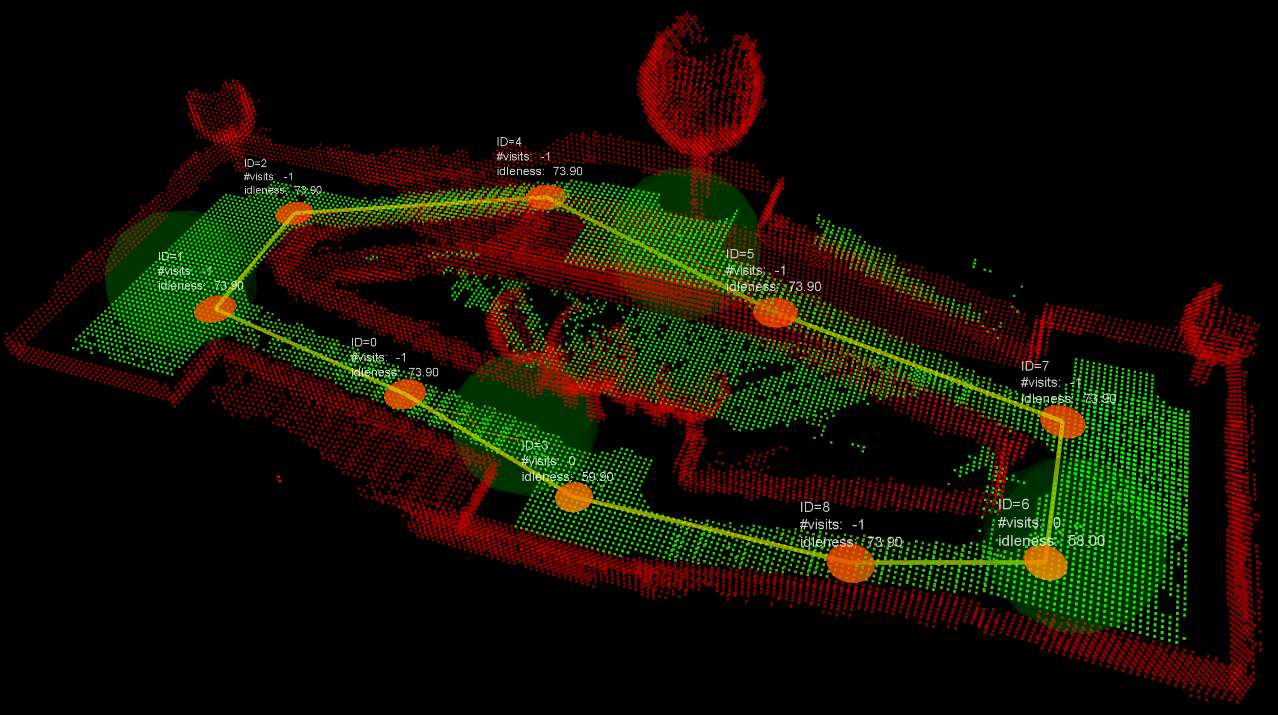}
}
\caption{Multi-floor scenarios in V-REP (\emph{left}) and their maps (\emph{right}): patrolling graph (red circular vertex and yellow edges), traversable regions (green point cloud), obstacle regions (red point cloud).}
\label{Fig:ThreedScenarios}
\end{center}
\end{figure*}

\newcommand{\gridsimimageheight}{4.6cm}

\begin{figure*}[!t]
\begin{center}
\subfloat[Corridor \label{SubFig:scenario:corridor}]
{
\includegraphics[height=1.47cm,keepaspectratio]{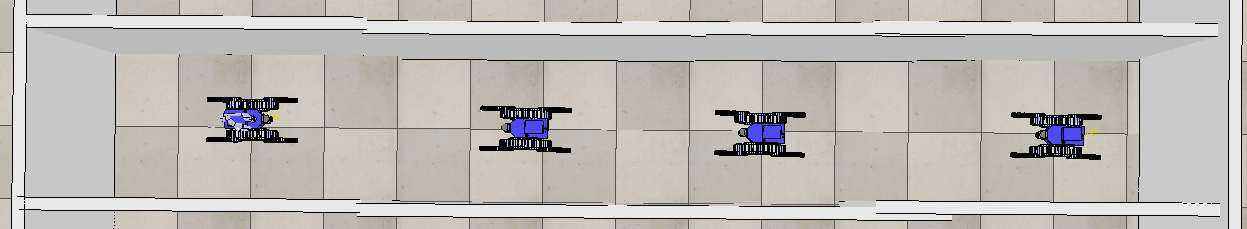}
\,
\includegraphics[height=1.47cm,keepaspectratio]{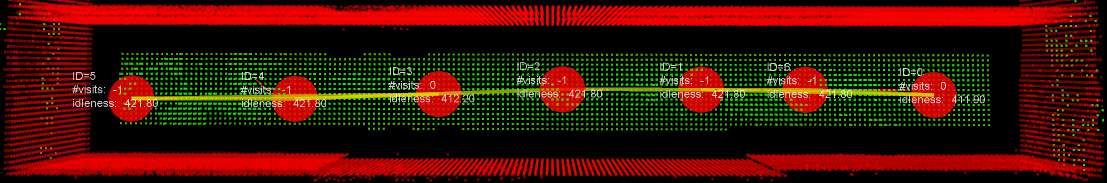}
}
\,
\subfloat[Crossroad \label{SubFig:scenario:crossroad}]
{
\includegraphics[height=\gridsimimageheight,keepaspectratio]{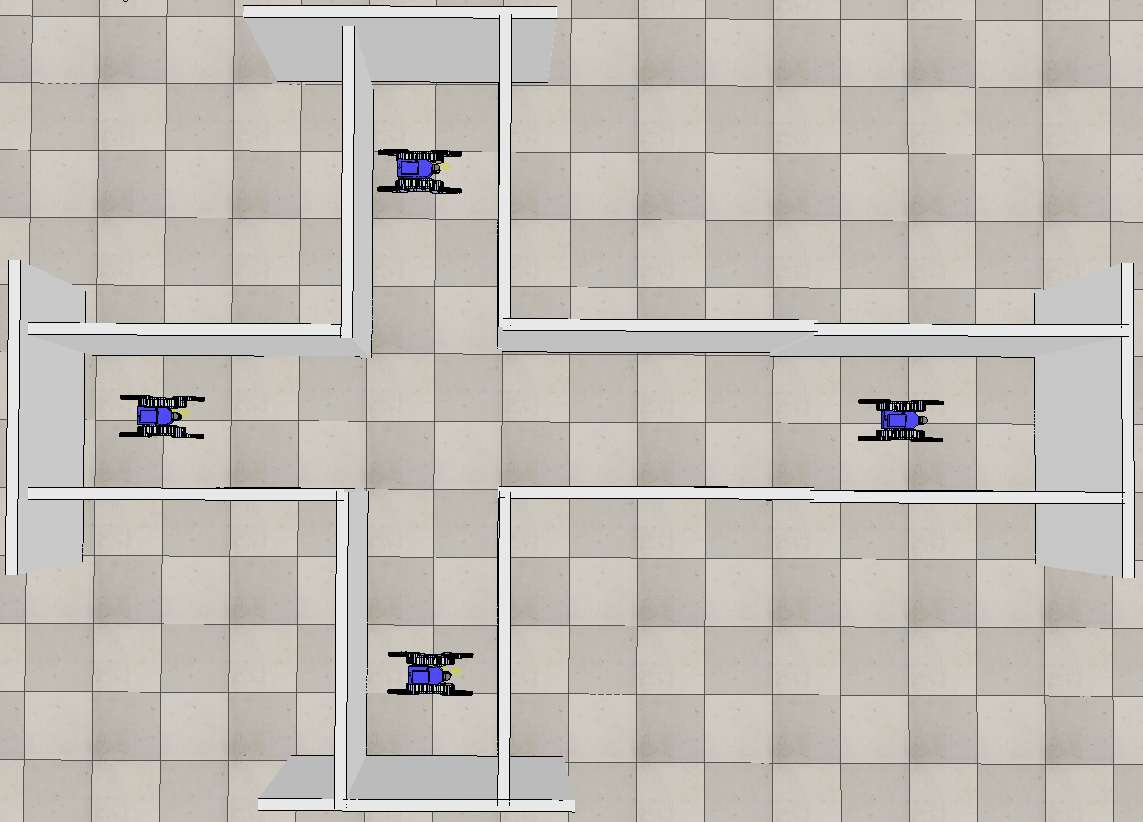}
\,
\includegraphics[height=\gridsimimageheight,keepaspectratio]{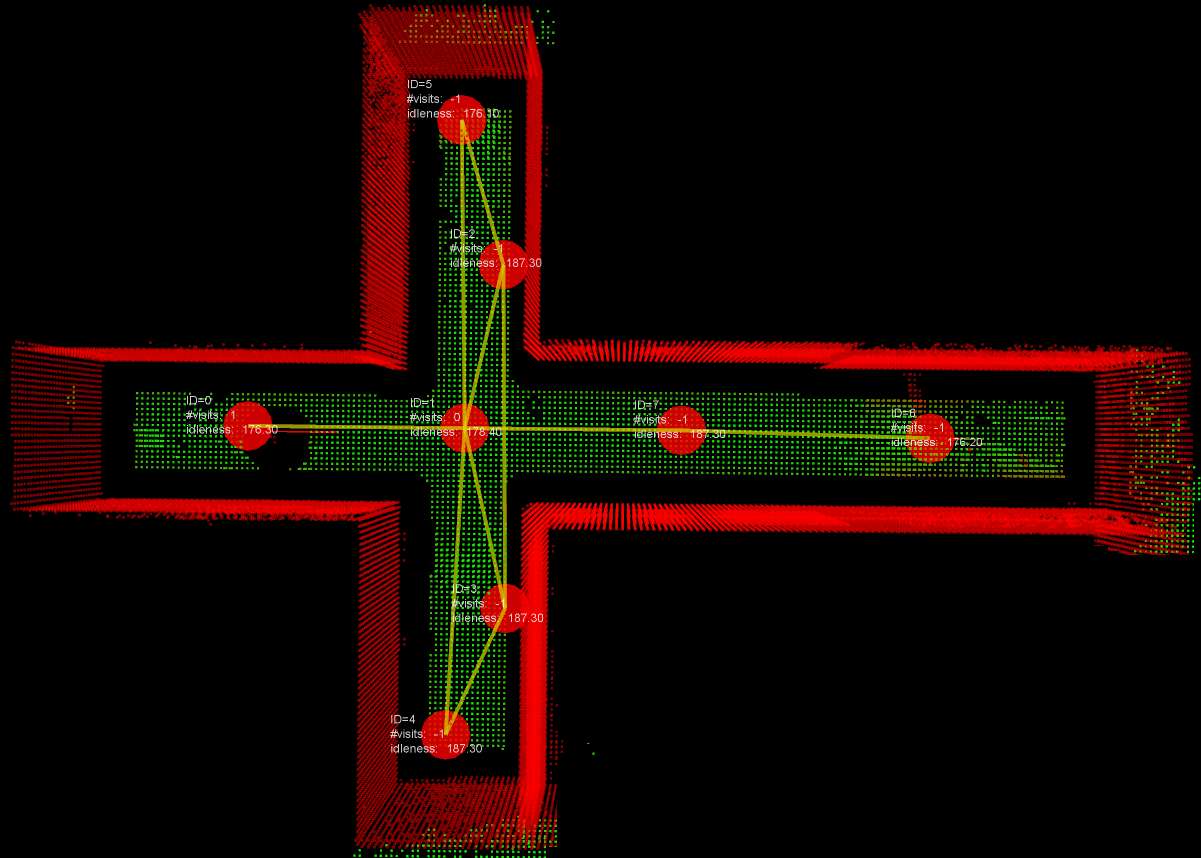}
}
\,
\subfloat[Small crossroad \label{SubFig:scenario:small_crossroad}]
{
\includegraphics[height=\gridsimimageheight,keepaspectratio]{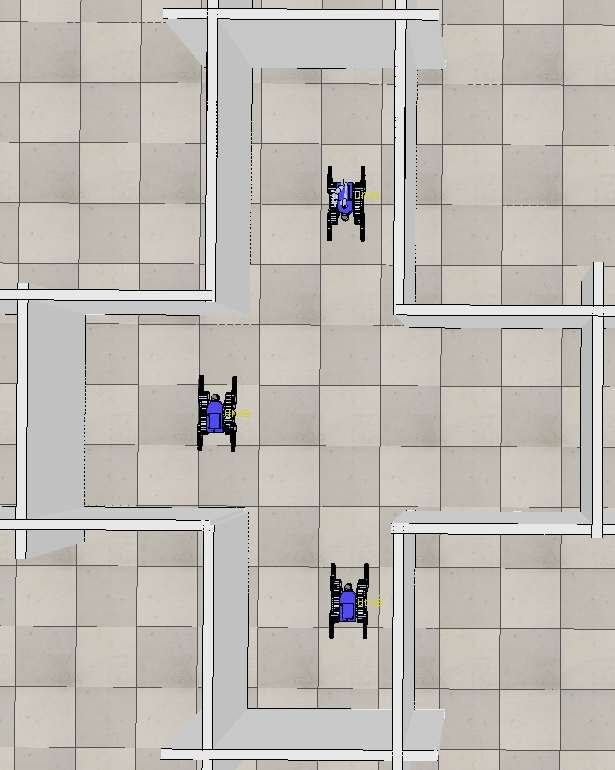}
\,
\includegraphics[height=\gridsimimageheight,keepaspectratio]{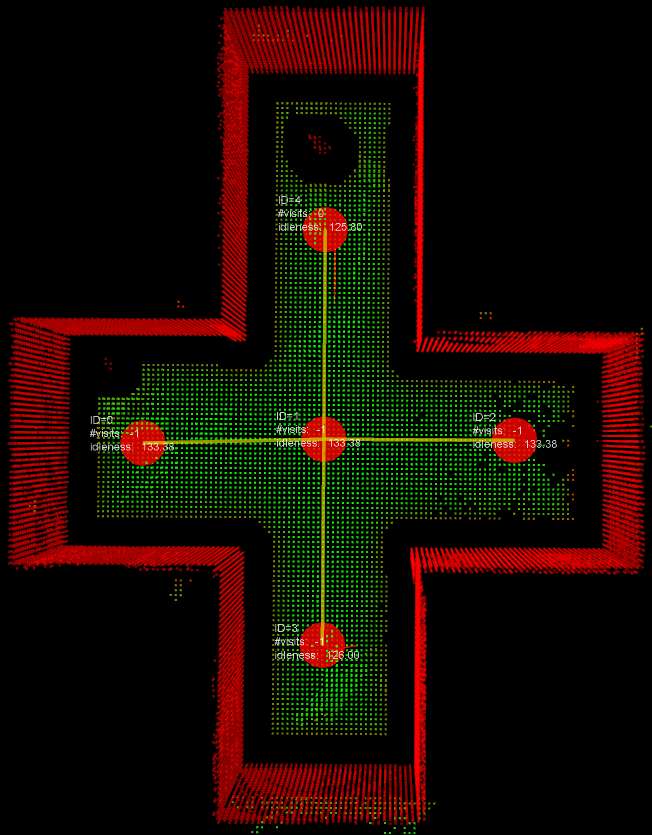}
}
\,
\hspace{0.8cm}
\subfloat[Fork \label{SubFig:scenario:fork}]
{
\includegraphics[height=\gridsimimageheight,keepaspectratio]{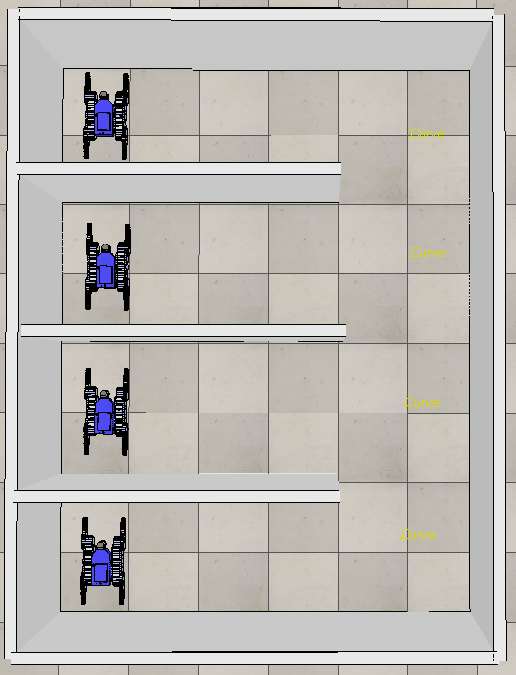}
\,
\includegraphics[height=\gridsimimageheight,keepaspectratio]{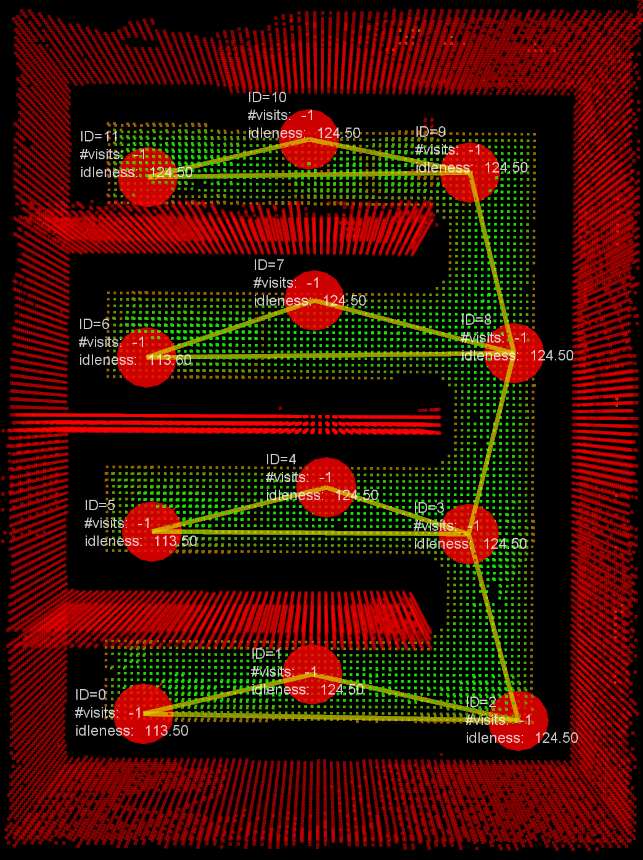}
}
\,
\subfloat[Ring \label{SubFig:scenario:2d_ring}]
{
\includegraphics[height=\gridsimimageheight,keepaspectratio]{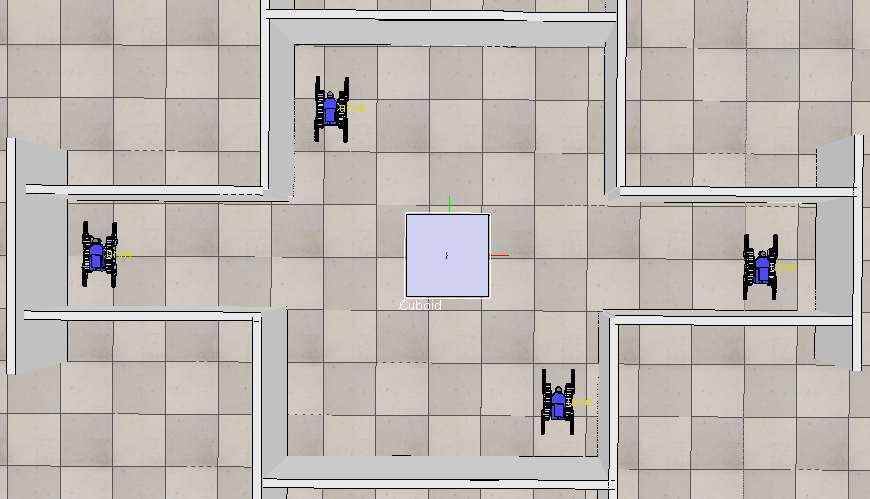}
\,
\includegraphics[height=\gridsimimageheight,keepaspectratio]{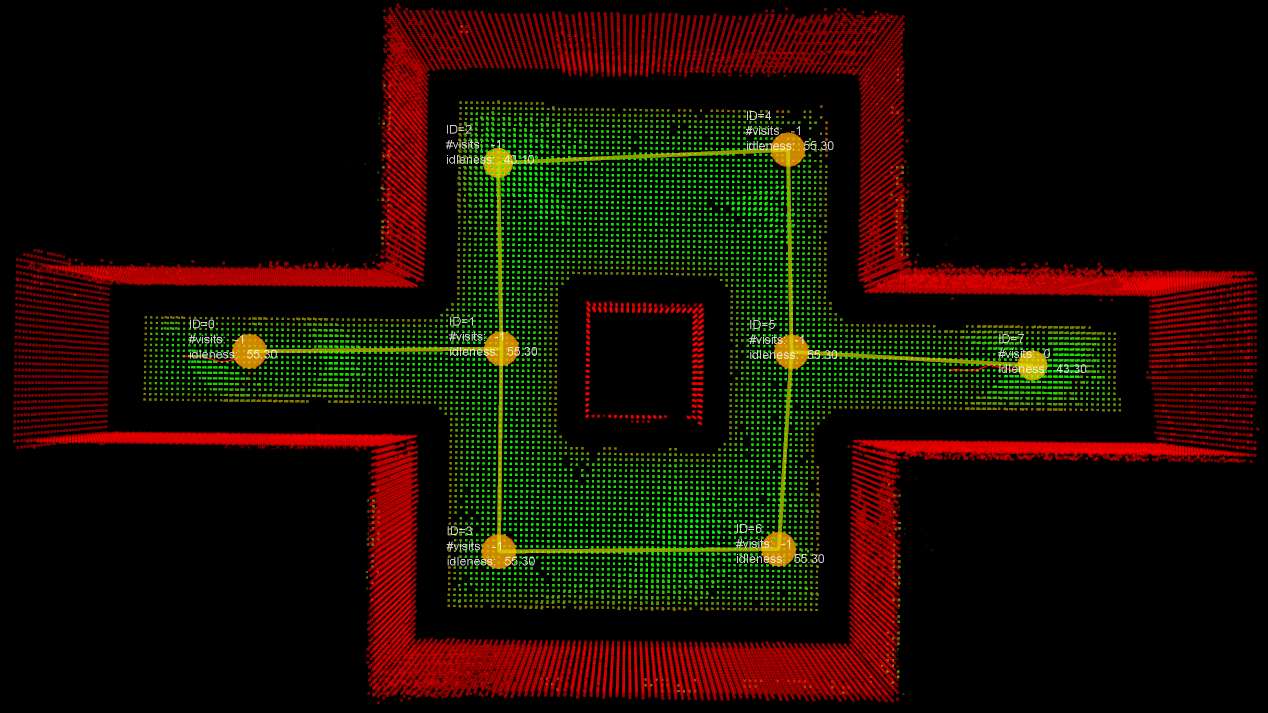}
}
\caption{Single-floor scenarios in V-REP (\emph{left}) and their maps (\emph{right}): patrolling graph (red circular vertex and yellow edges), traversable regions (green point cloud), obstacle regions (red point cloud).}
\label{Fig:PlanarScenarios}
\end{center}
\end{figure*}

The simulated scenarios are depicted in Figures~\ref{Fig:PlanarScenarioThreeWays}, \ref{Fig:ThreedScenarios} and~\ref{Fig:PlanarScenarios}. In particular,   Fig.~\ref{Fig:ThreedScenarios} collects the used multi-floor scenarios, while \ref{Fig:PlanarScenarioThreeWays} and~\ref{Fig:PlanarScenarios} show the single-floor scenarios. In our view, the scenarios of Fig.~\ref{Fig:PlanarScenarios} can be considered as representative topological types of environment junctions, which may be found in common single-floor scenarios. In particular, we simulated the challenging scenario ``small crossroad" with teams of three robots and four robots. 
Some videos of the simulations and further details are publicly available\footnote{ \scriptsize
\href{https://sites.google.com/a/dis.uniroma1.it/3d-cc-patrolling/}{ https://sites.google.com/a/dis.uniroma1.it/3d-cc-patrolling/}}.

In a first stage, we separately evaluated the multi-robot traversability in order to show how it improves the behaviour of the path-planner. To this aim, we used the challenging scenario reported in Fig.~\ref{SubFig:scenario:threeways:vrep} and assigned to the each robot one of the distinct cyclic paths shown in Fig.~\ref{SubFig:scenario:threeways:paths}. Here, each robot was required to move back and forth between its two assigned waypoints by using only the path planner (no patrolling graph and no patrolling agent). We compared the behaviour of the path planner with and without the multi-robot traversability. In the scenario of Fig.~\ref{SubFig:scenario:threeways:paths}, we run 10 simulations, each one lasting 10 minutes. We observed that a team of three robots, which used the basic path planners, always got stuck in a deadlock (around the intersection of the three cyclic paths). On the other hand, path planners and multi-robot traversability succeed in nicely coordinating the robots without congestions or deadlocks\footnote{Two simulation videos are available on our website and show these behaviour.}. 

In similar environments, characterized by narrow crossroads, we obtained comparable results. In general, when considering only the path planner, we observed that the multi-robot traversability improves the navigation ability of a robot team. This becomes particularly evident in scenarios where significant congestions and deadlocks may occur. Clearly, there are complex cases which cannot be managed by the multi-robot traversability, given the high complexity of the general multi-robot path planning problem~\cite{Lavalle-2006}. Nonetheless, we empirically show that our two level coordination strategy (multi-robot traversability plus patrolling agent) can resolve conflicts and prevent deadlocks in complex patrolling scenarios. 

In a second stage, we evaluated the (full) two level coordination strategy. To this aim, we used as performance metrics the idleness statistics introduced in Section~\ref{Sect:PatrollingGraph} and the total number of occurred interference events. In particular, we continuously measured in a moving-window $[t-\Delta,t] \subset \mathbb{R}$ the average graph idleness $I^a_G[t-\Delta,t]$, its standard deviation $I^{\sigma}_G[t-\Delta,t]$ and its maximum value $I^M_G[t-\Delta,t]$, where $t$ denotes the current time and we selected $\Delta=600s$. In particular, we considered a moving-window in order to better observe transient dynamics. We found that a time width of 600$s$ was a good compromise to significantly capture both transients and regime behaviours.

Moreover, we counted the total number of interference events that are broadcast by the robots when their centres get closer than the safety distance $D_s$ (see Section~\ref{Sect:TopologicalMetricConflicts}). These checks are executed at $2 Hz$ and recorded at a pre-fixed frequency of $0.2 Hz$. Indeed, such an interference measure overall represents how long the robot team experienced interference and conflicts\footnote{Since V-REP simulations are computationally demanding in our setup, the simulated robots were not able to move in real time and their motions were very slow (this can be observed in our simulation videos on our website). As a result, when robots got in interference, they persisted in such conditions for longer times with respect to a normal real time simulation.}.  

We compared the patrolling strategy presented in this paper (see Algorithms~\ref{Alg:PatrolAgent}--\ref{Alg:SelectNexNode}) with two simplified versions of it. The first simplified strategy is obtained by only disabling the multi-robot traversability (metric coordination). The second one is obtained by disabling node conflict management (topological coordination) and shared idleness estimation (cooperation), but it preserves metric coordination. 
In the remainder of this paper, we refer to the full patrolling strategy as \emph{CC strategy} (Cooperation plus Coordination), to the first simplified strategy as \emph{CwMC strategy} (Cooperation without Metric Coordination) and to the second simplified strategy as \emph{No-CC strategy}. 
As explained in 
Sect.~\ref{Sect:NextNodeSelection},
in this work, we selected a reactive strategy for the implementations of the functions BuildSearchSet($\cdot$) and Compute\-Next\-Best\-No\-de($\cdot$) of Algorithm~\ref{Alg:SelectNexNode}.

\begin{figure*}[!t]
\begin{center}
\subfloat[Three-ways \label{SubFig:performances:3ways:cwc}]
{
\includegraphics[height=\PerformancesHeight,keepaspectratio]{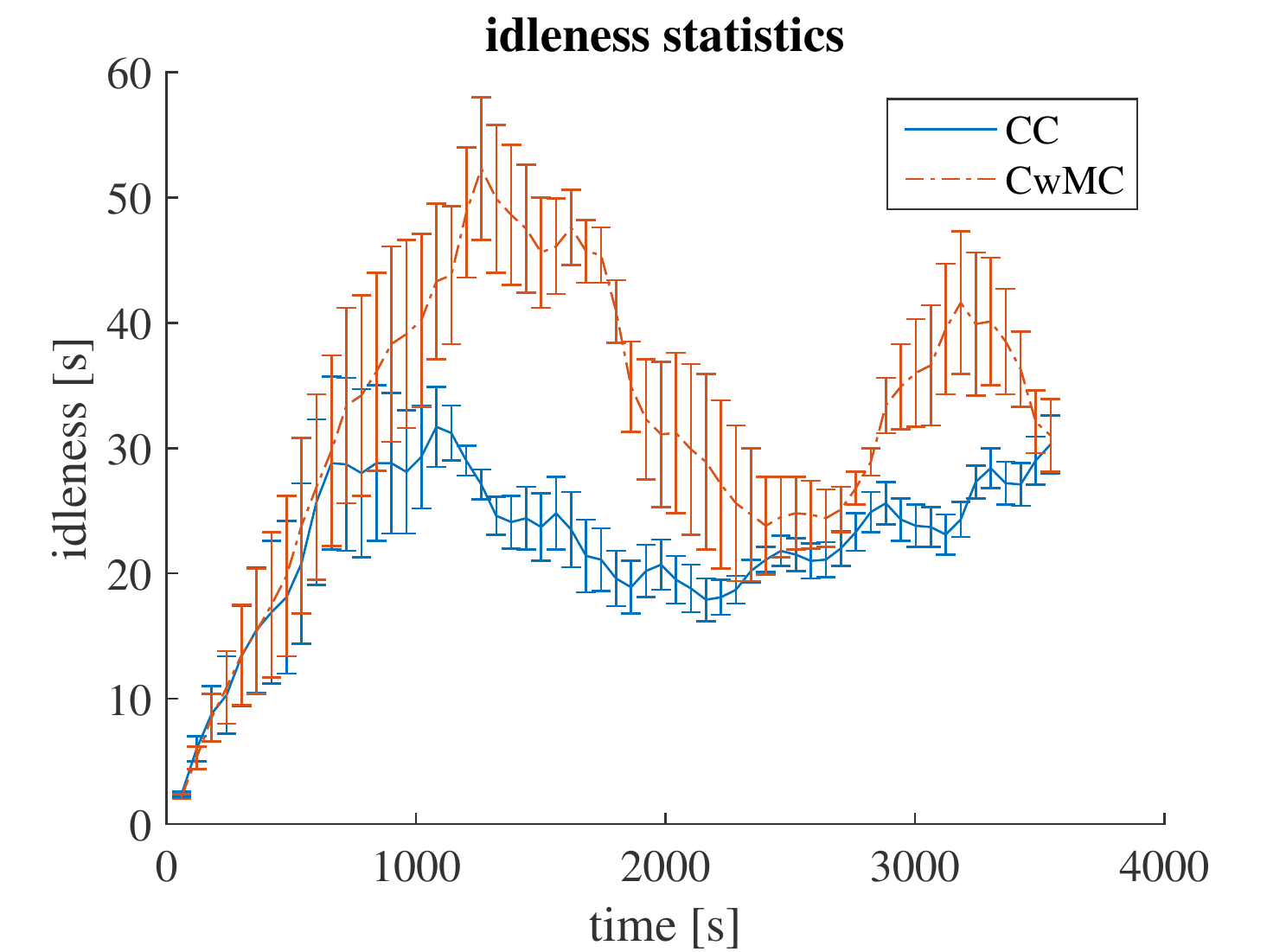}
\includegraphics[height=\PerformancesHeight,keepaspectratio]{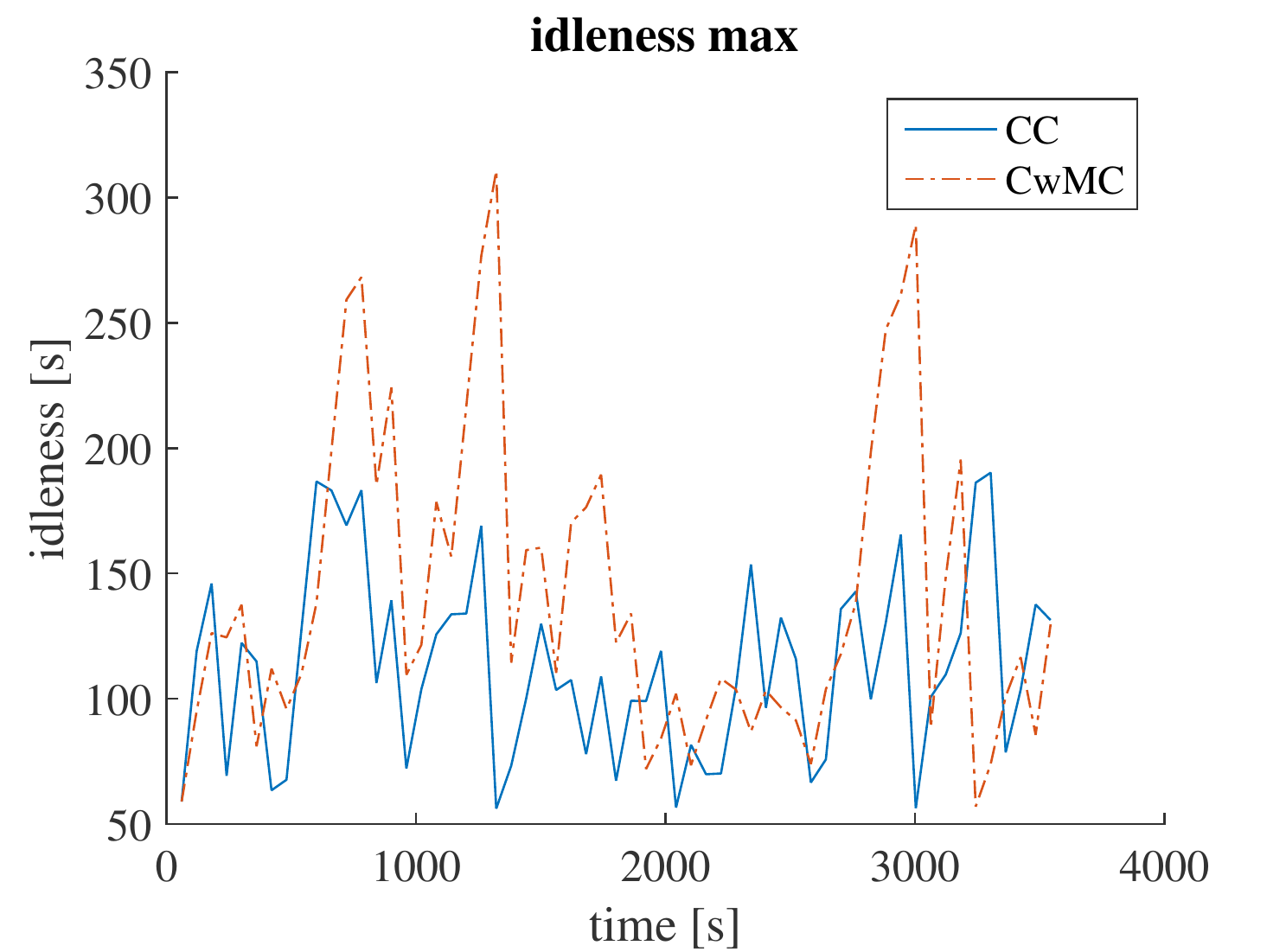}
\includegraphics[height=\PerformancesHeight,keepaspectratio]{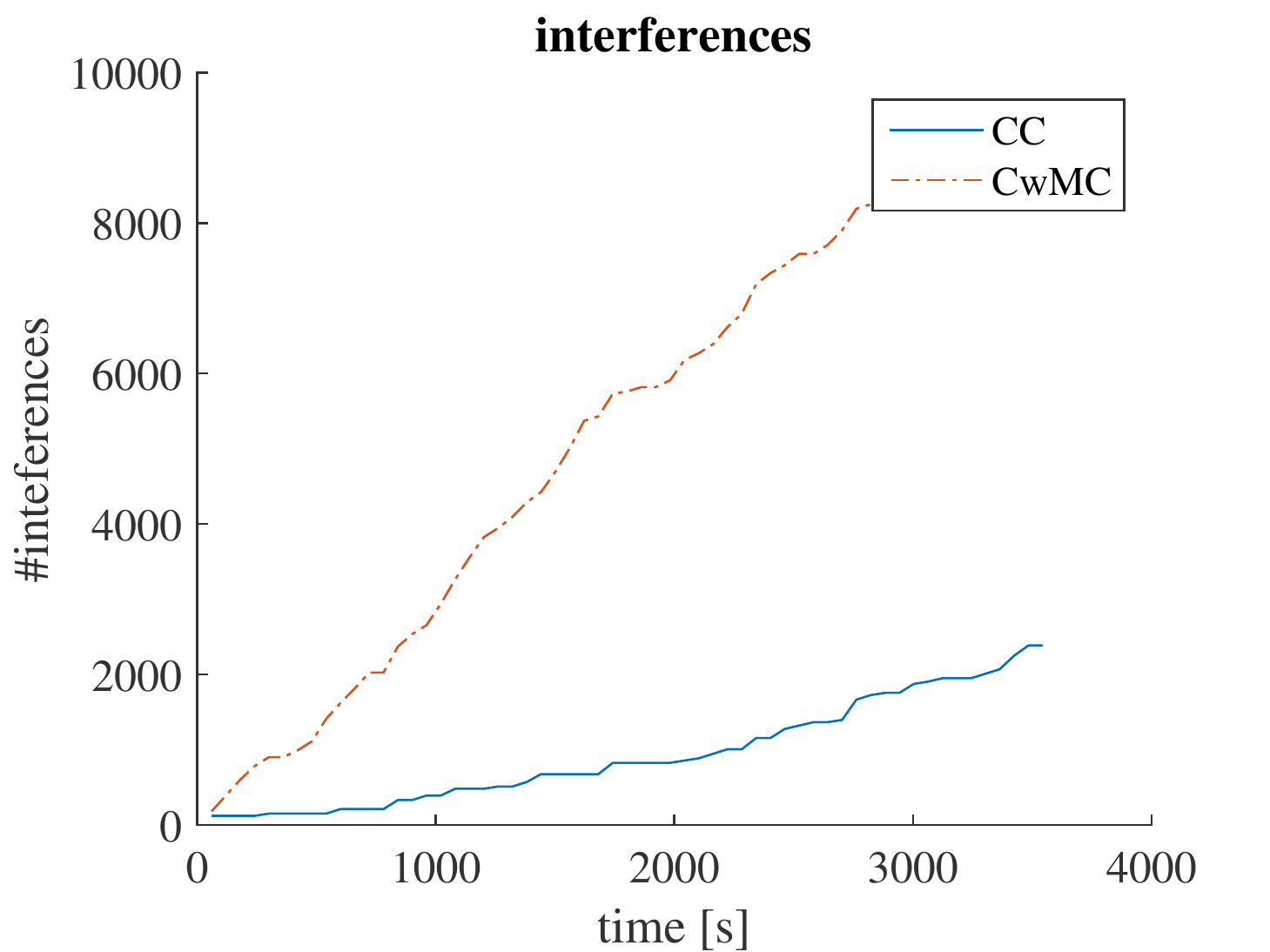}
}
\,
\subfloat[Crossroad \label{SubFig:performances:crossroad:cwc}]
{
\includegraphics[height=\PerformancesHeight,keepaspectratio]{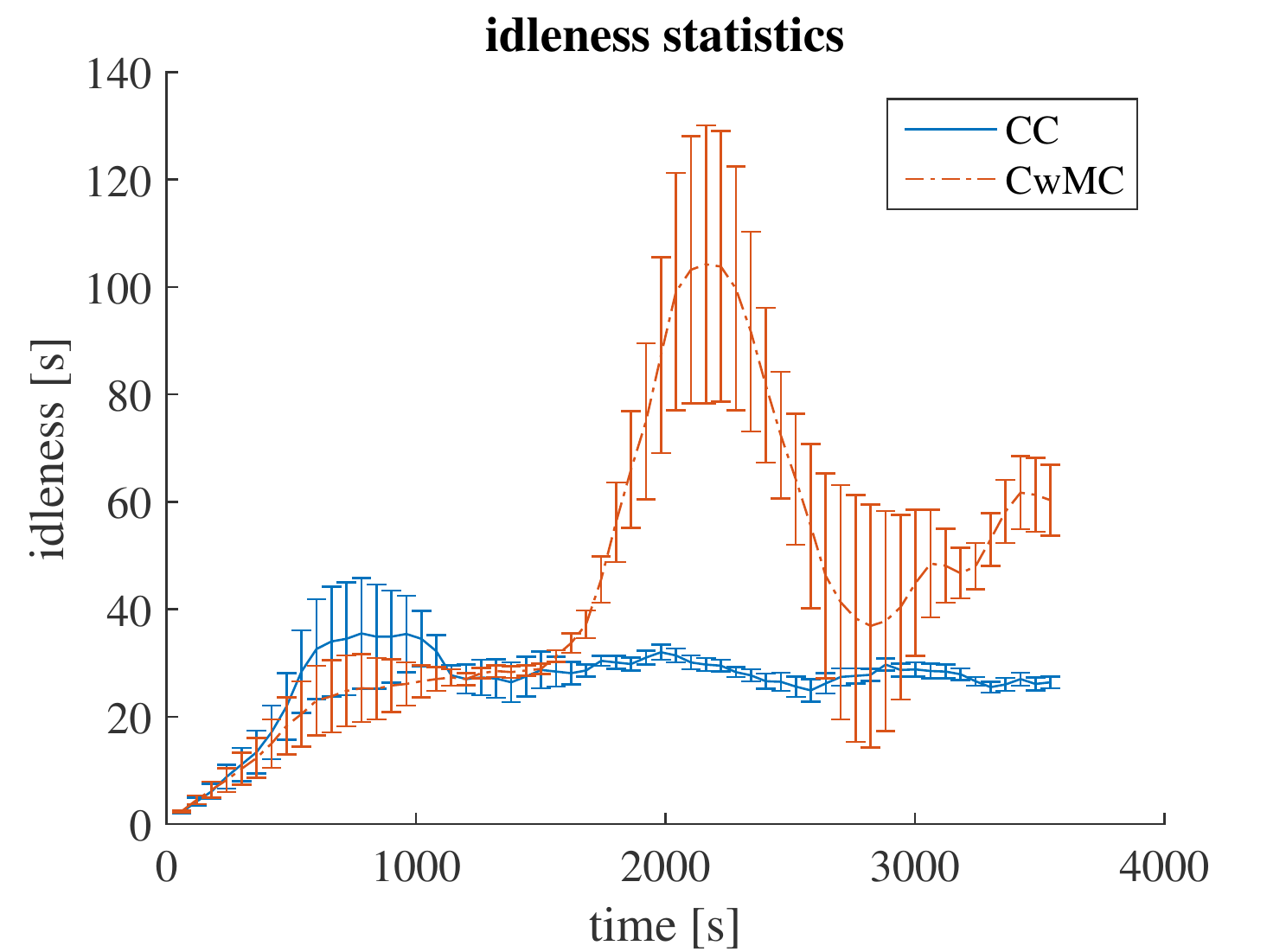}
\includegraphics[height=\PerformancesHeight,keepaspectratio]{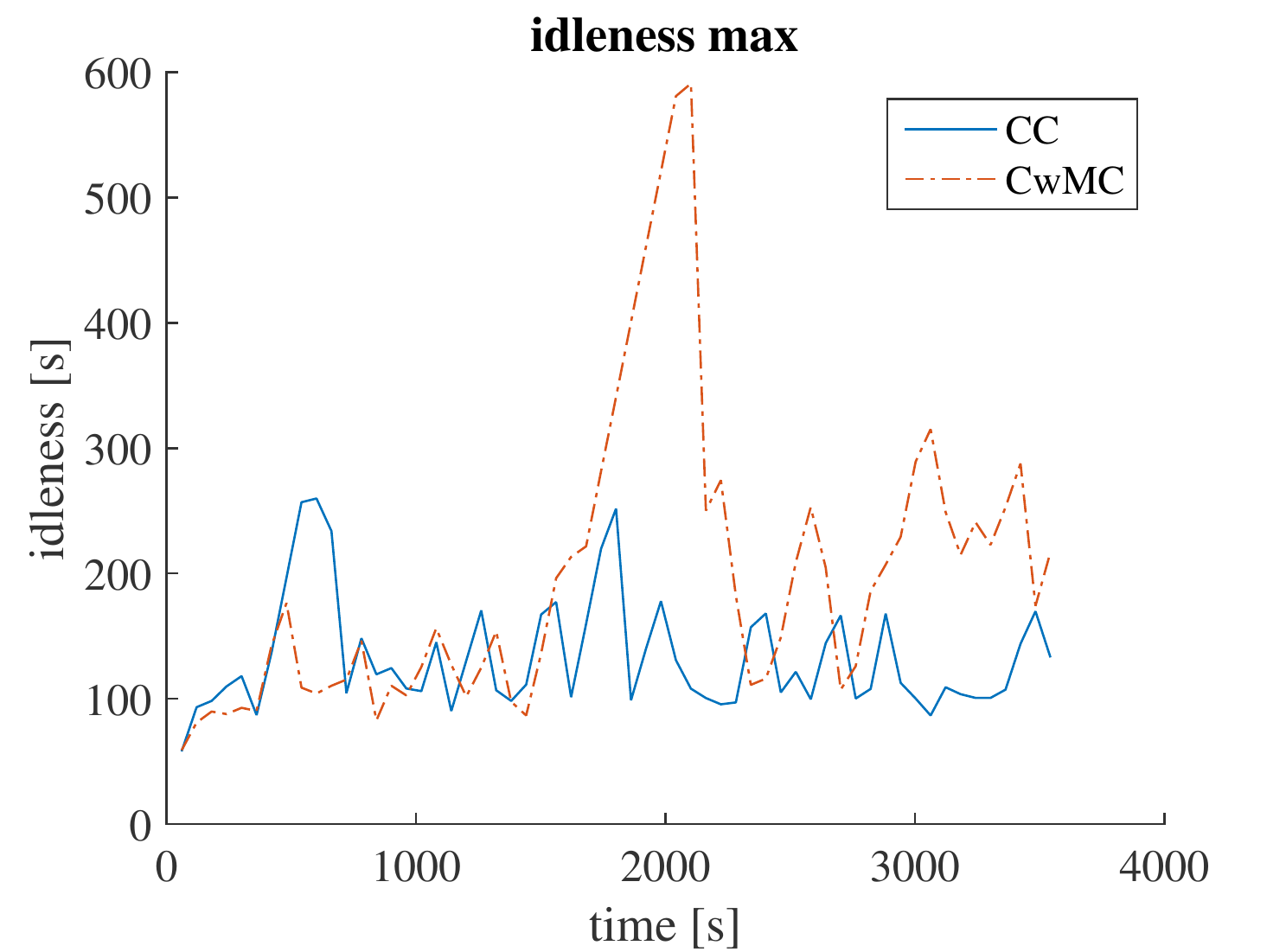}
\includegraphics[height=\PerformancesHeight,keepaspectratio]{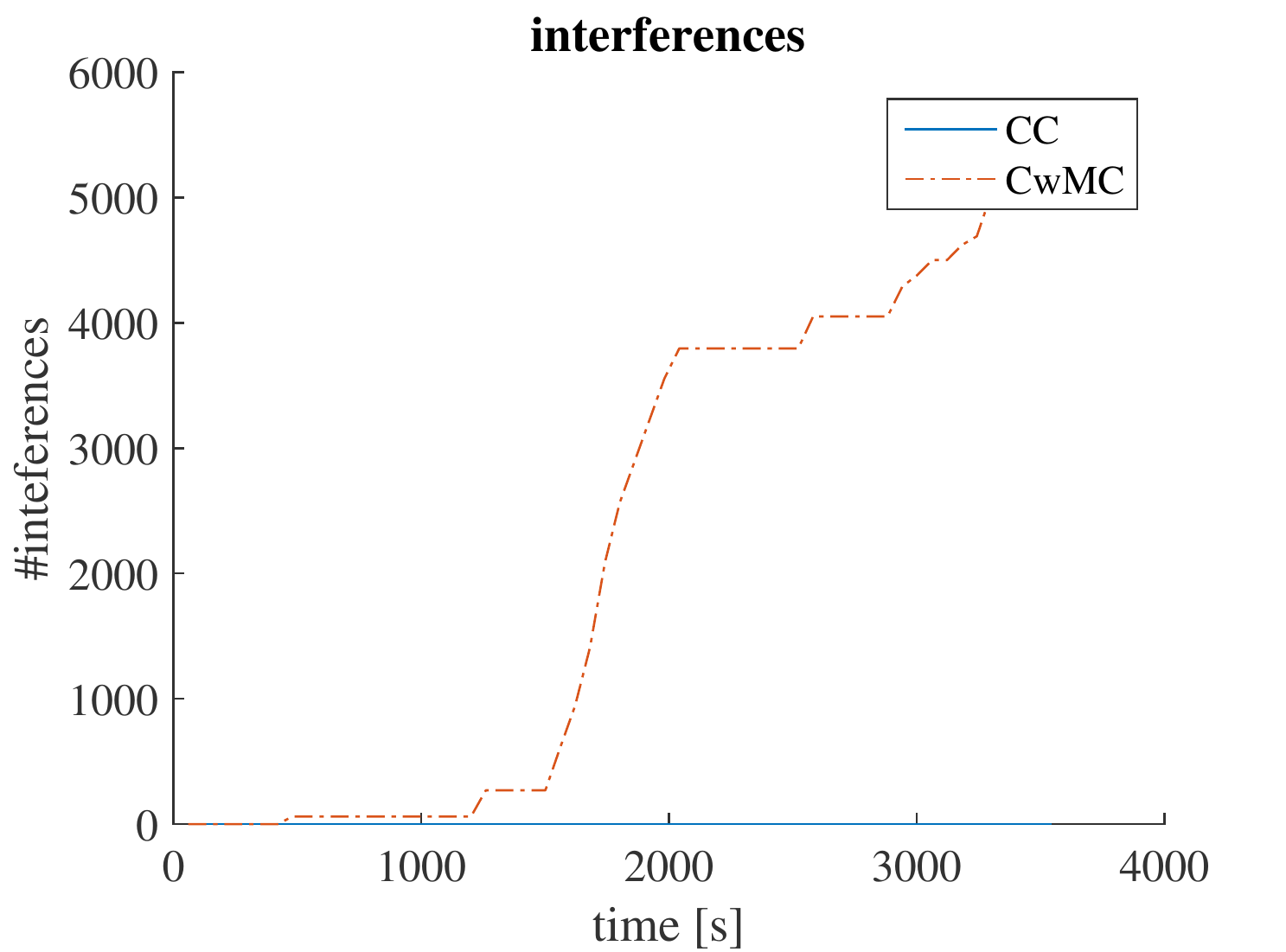}
}
\caption{Performance metrics obtained by comparing \emph{CC} with \emph{CwMC} in the three-ways and crossroad scenarios. \emph{Left}: a plot of the average idleness of the graph along with its standard deviation. Statistics are computed in a moving time-window of width $600s$. \emph{Center}: the maximum idleness observed in the moving time-window. \emph{Right}: the total number of observed interferences up to the current time. \emph{CC} strategy performances are reported in blue while \emph{CwMC} performances are depicted in red.}
\label{Fig:CwCPerformanceMetrics}
\end{center}
\end{figure*}

\begin{figure*}[!t]
\begin{center}
\subfloat[Multi-floor ramp \label{SubFig:performances:3d_ramp}]
{
\includegraphics[height=\PerformancesHeight,keepaspectratio]{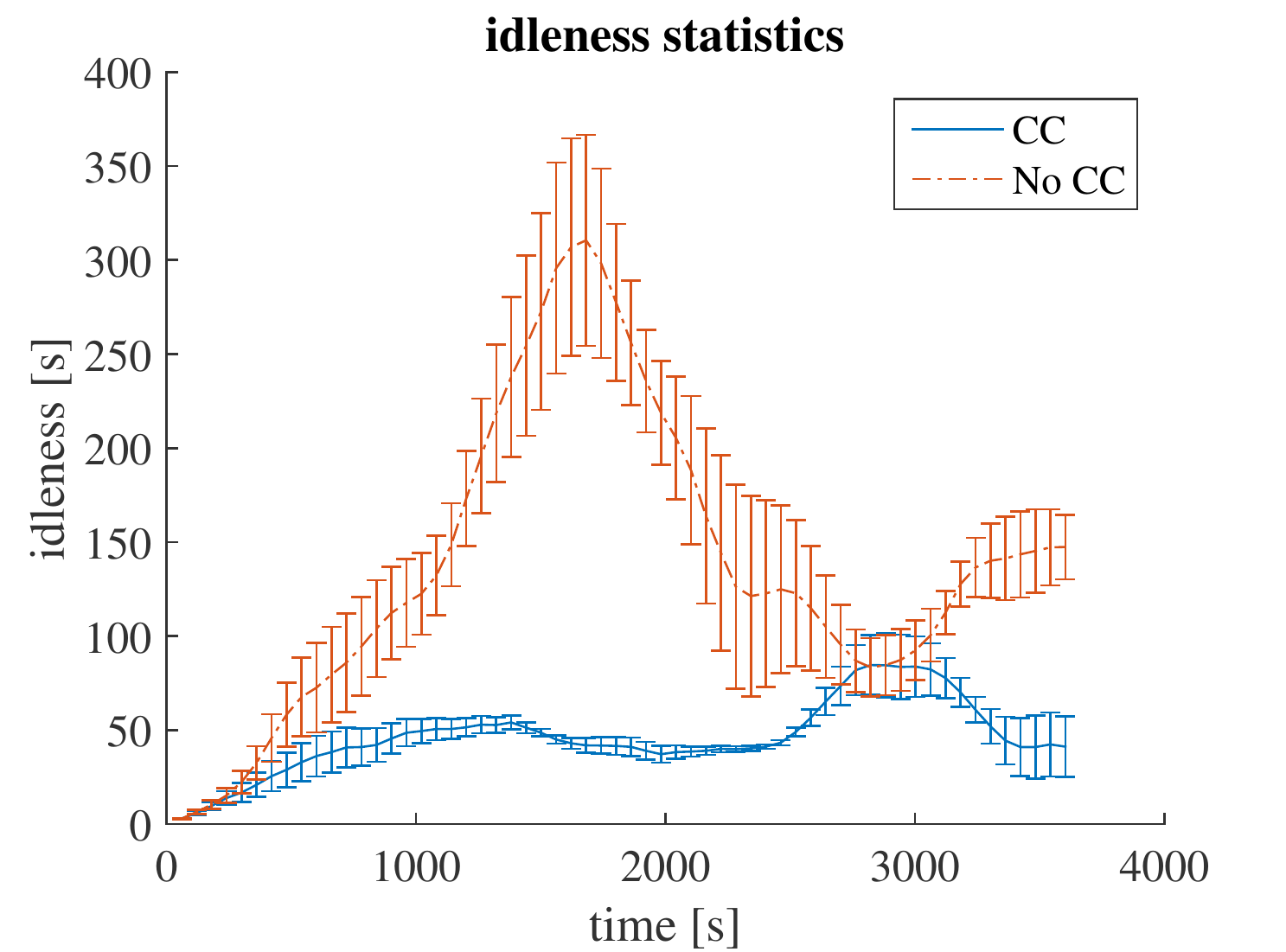}
\includegraphics[height=\PerformancesHeight,keepaspectratio]{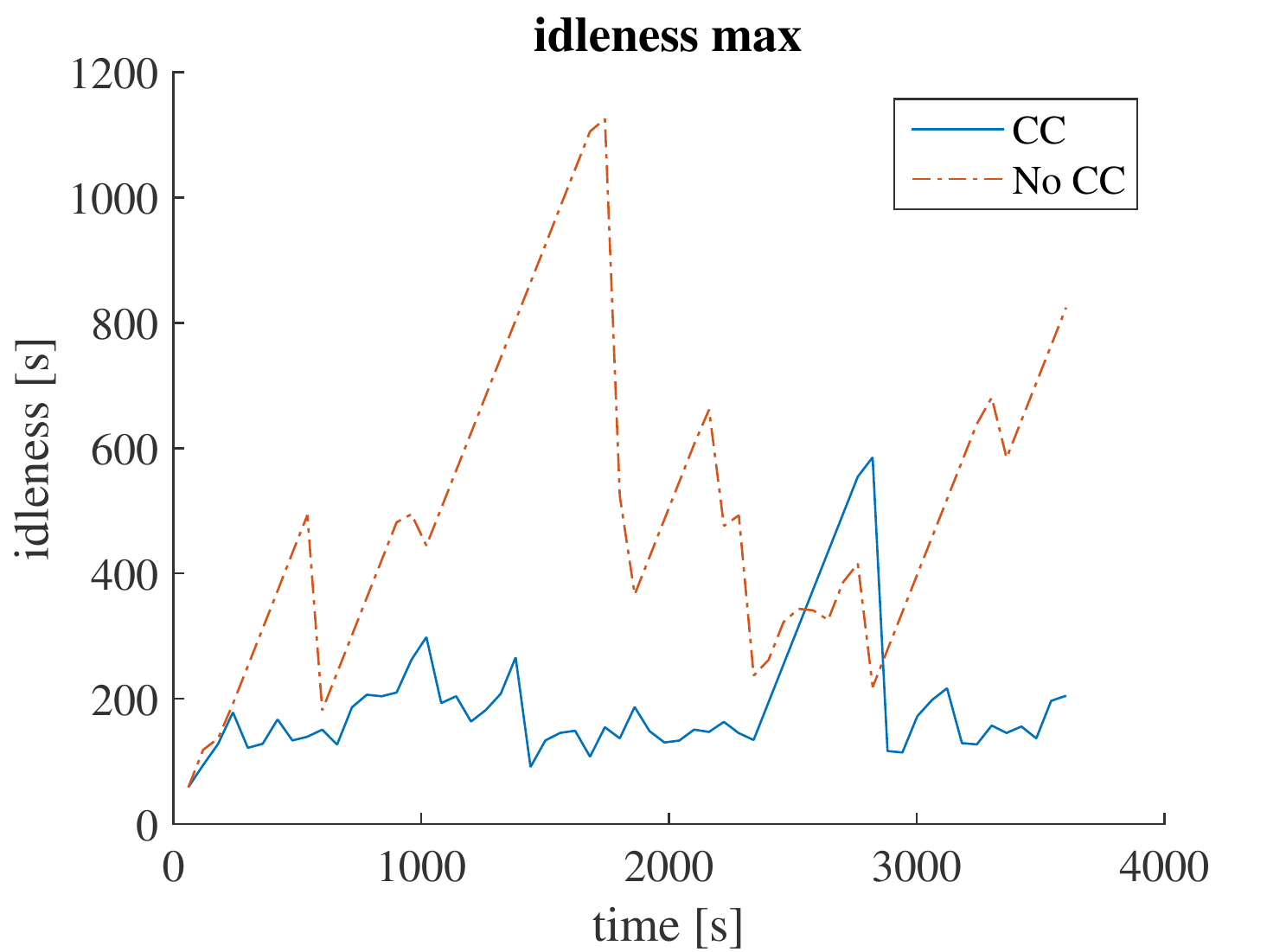}
\includegraphics[height=\PerformancesHeight,keepaspectratio]{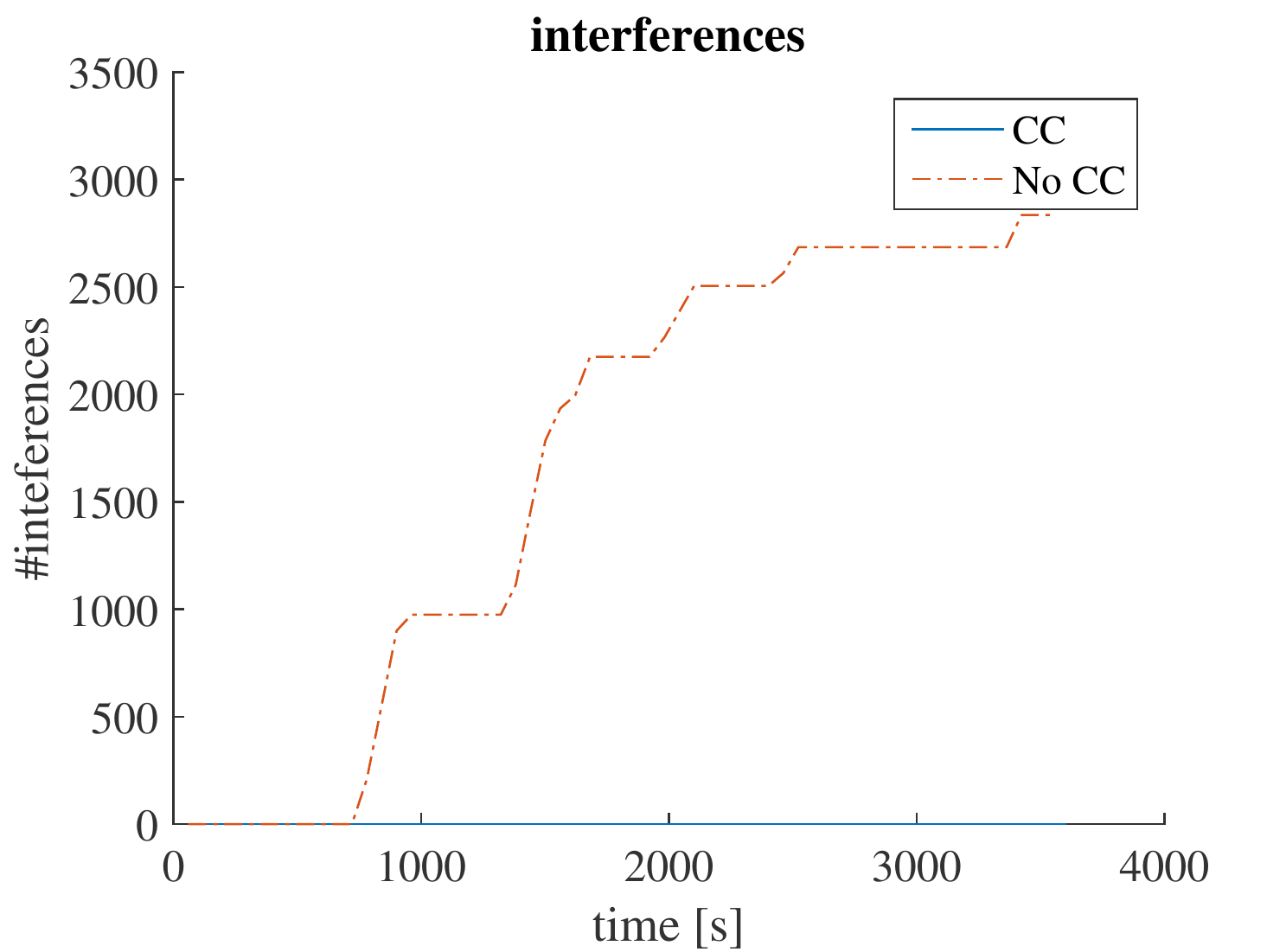}
}
\,
\subfloat[Two-floor ring \label{SubFig:performances:3d_ring}]
{
\includegraphics[height=\PerformancesHeight,keepaspectratio]{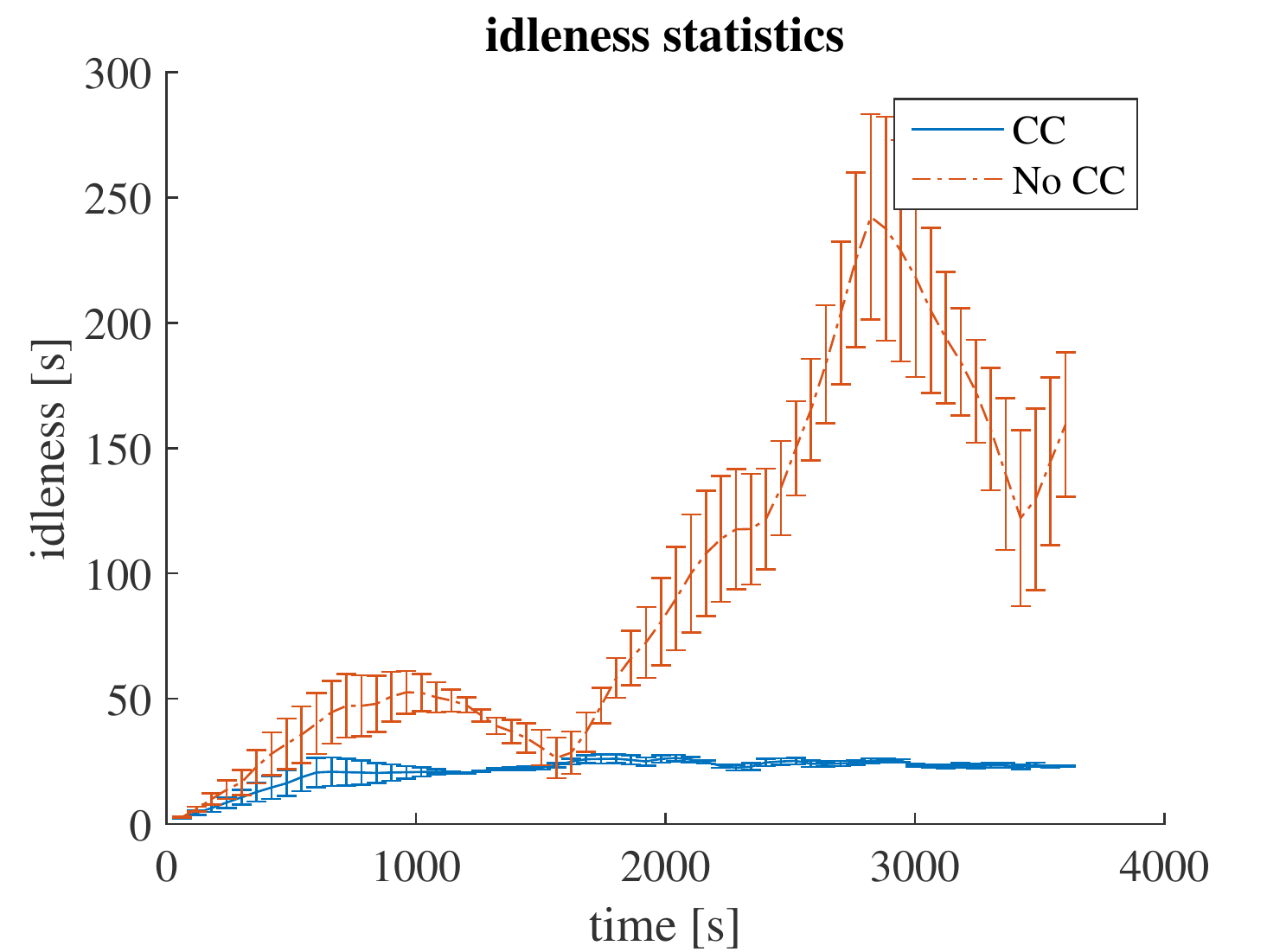}
\includegraphics[height=\PerformancesHeight,keepaspectratio]{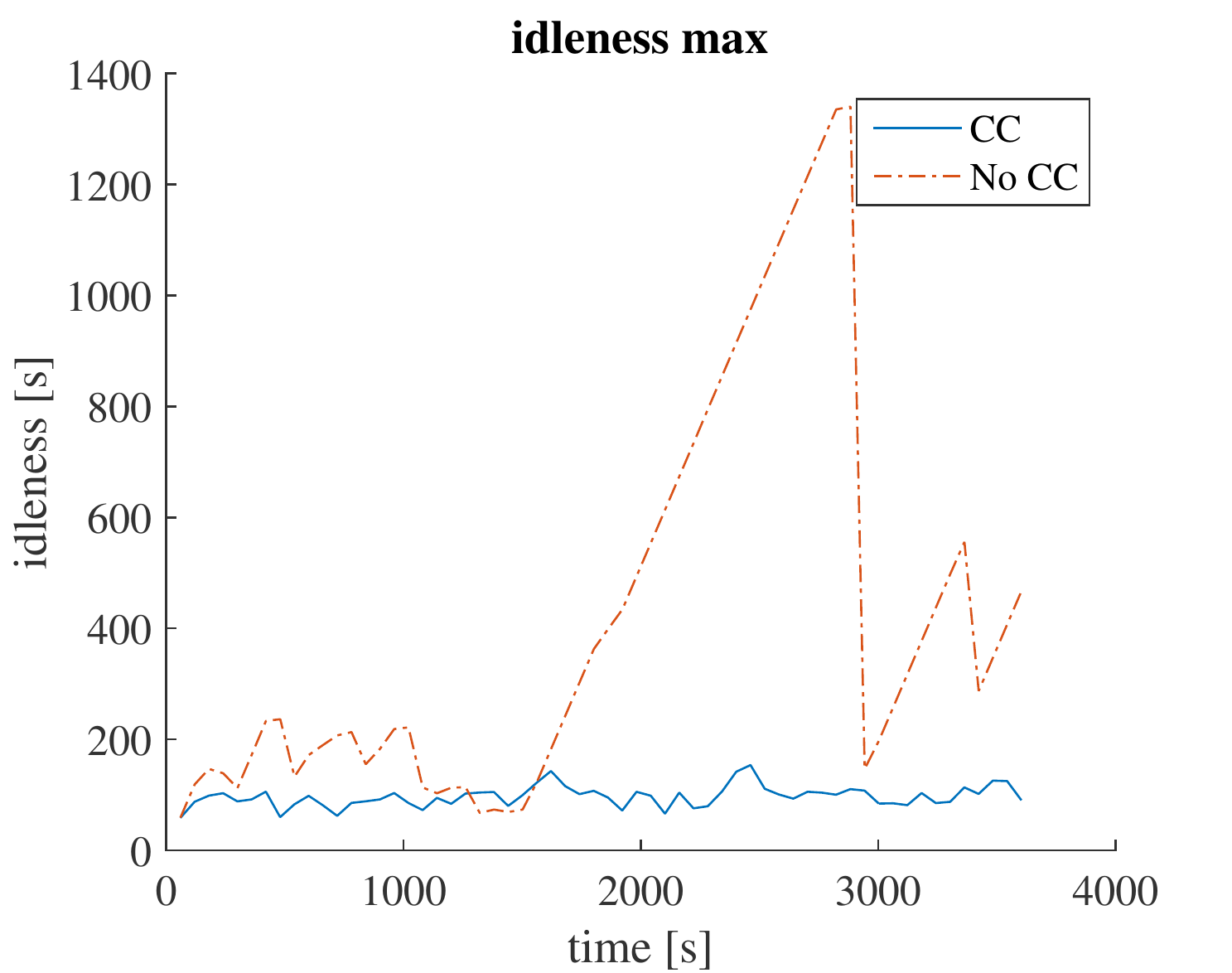}
\includegraphics[height=\PerformancesHeight,keepaspectratio]{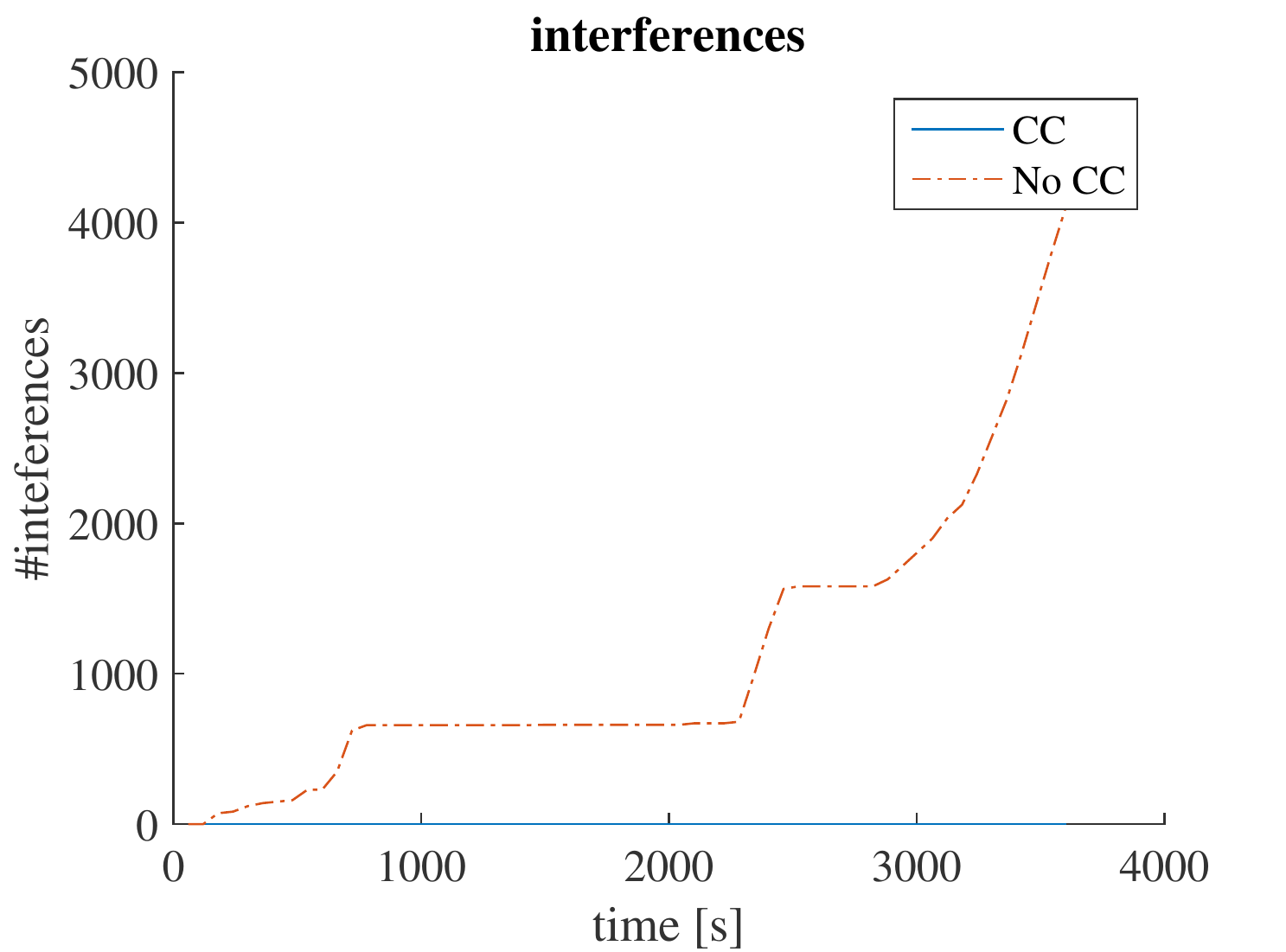}
}
\caption{Performance metrics obtained by comparing \emph{CC} with \emph{No-CC }in the multi-floor scenarios. \emph{Left}: a plot of the average idleness of the graph along with its standard deviation. Statistics are computed in a moving time-window of width $600s$. \emph{Center}: the maximum idleness observed in the moving time-window. \emph{Right}: the total number of observed interferences up to the current time. \emph{CC} strategy performances are reported in blue while \emph{No-CC} performances are depicted in red.}
\label{Fig:MultiFloorPerformanceMetrics}
\end{center}
\end{figure*}

\begin{figure*}[!t]
\begin{center}
\subfloat[Small crossroad with 3 robots \label{SubFig:performances:smallcrossroad3w3}]
{
\includegraphics[height=\PerformancesHeight,keepaspectratio]{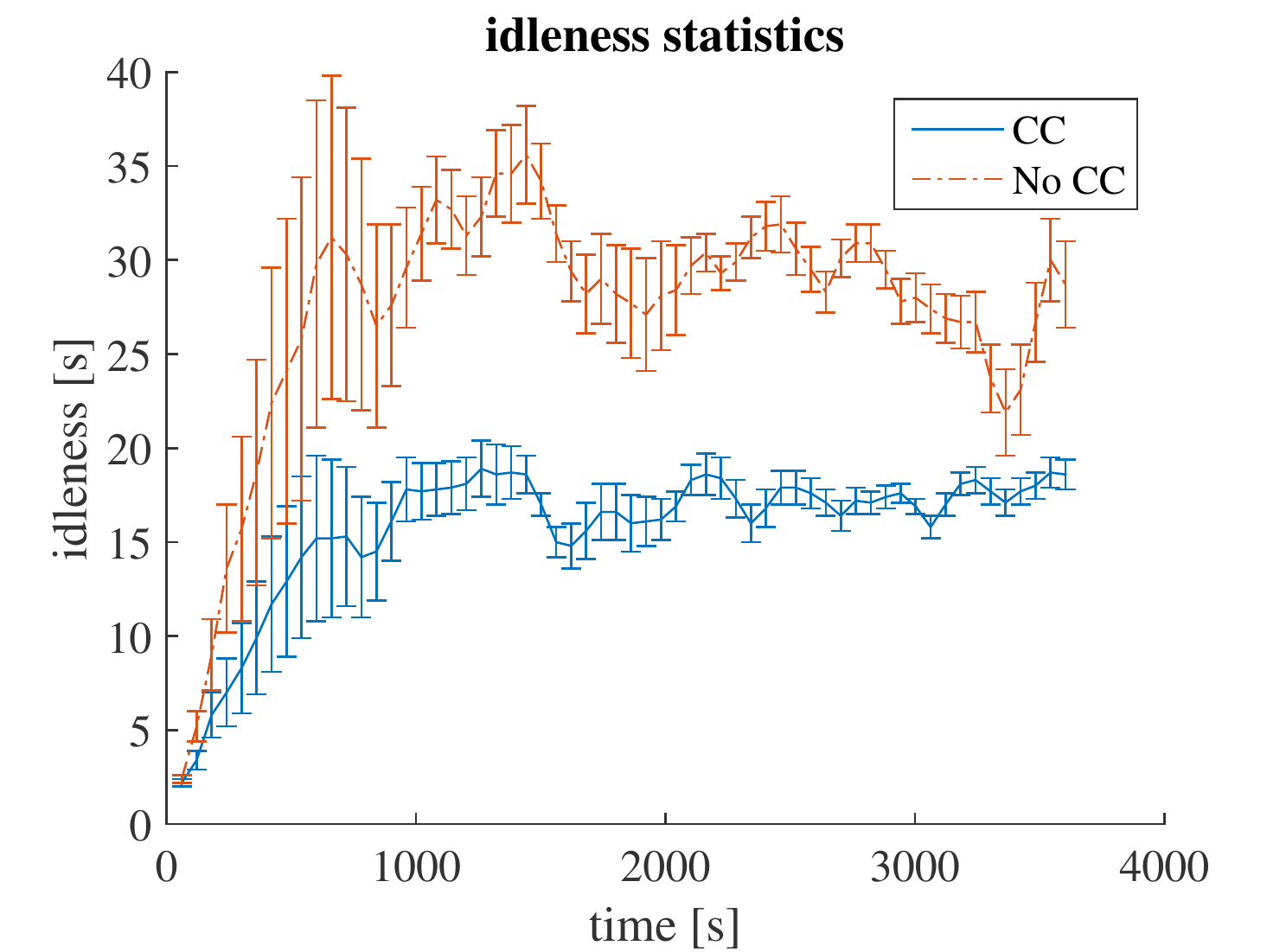}
\includegraphics[height=\PerformancesHeight,keepaspectratio]{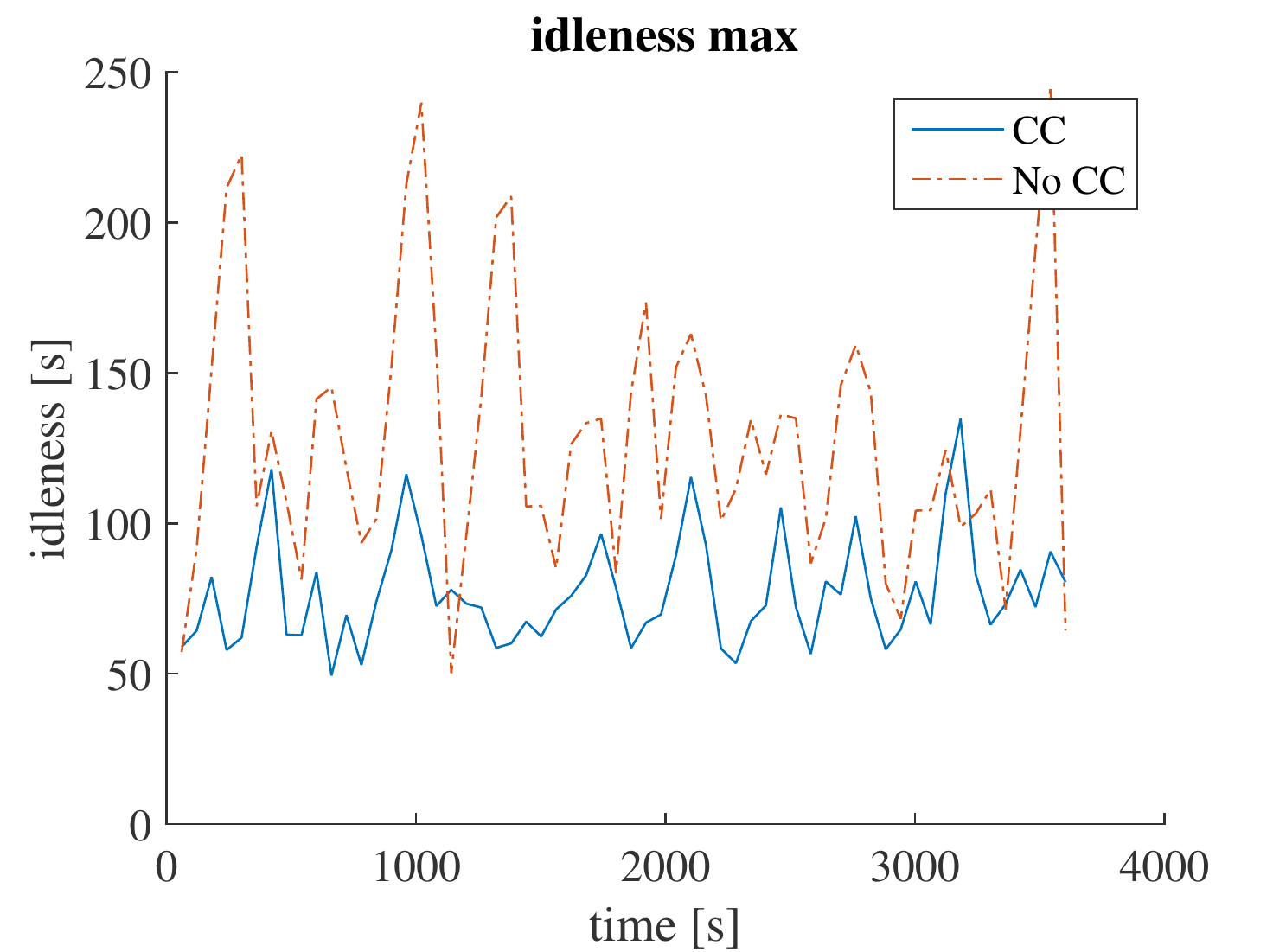}
\includegraphics[height=\PerformancesHeight,keepaspectratio]{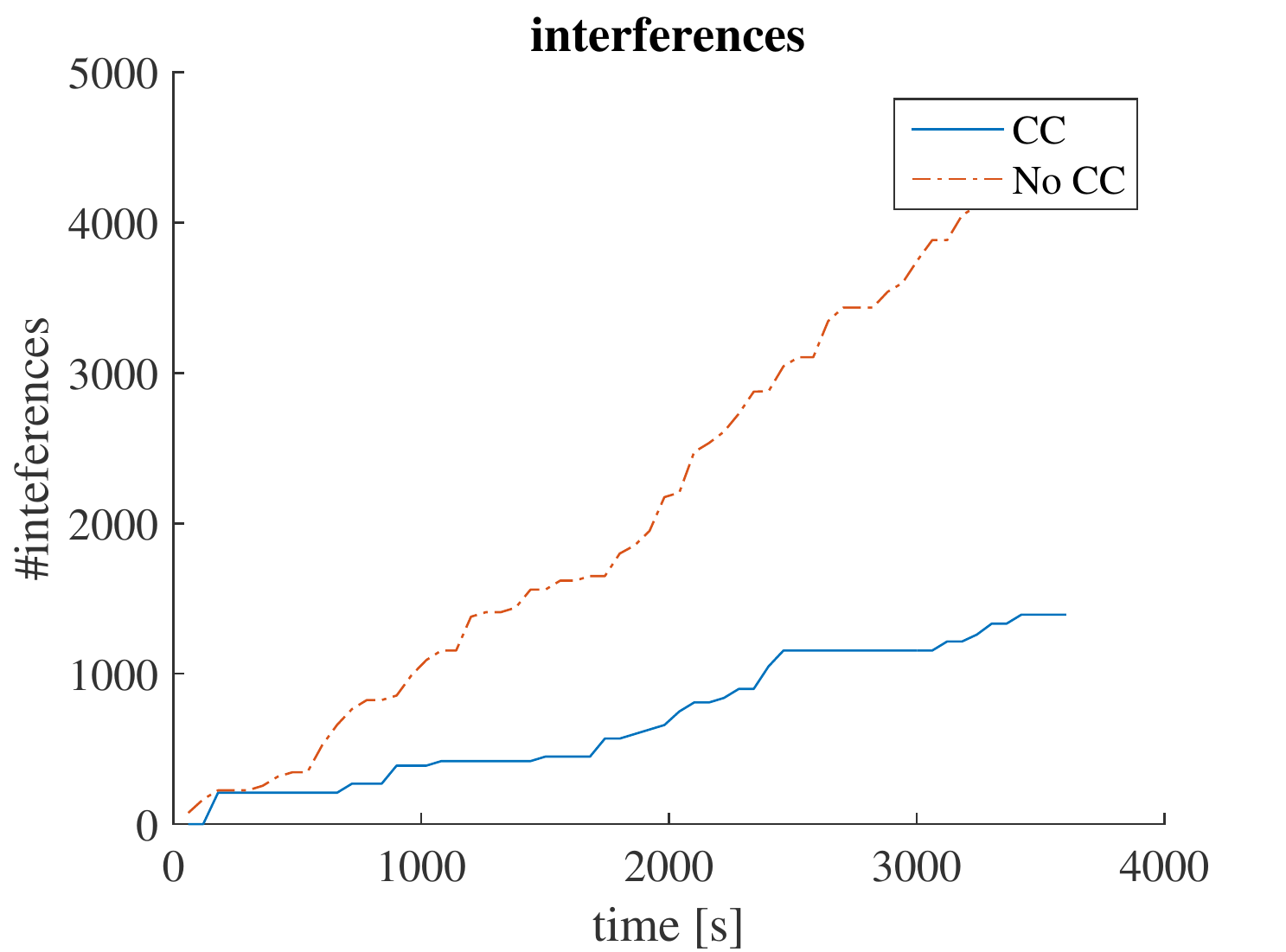}
}
\,
\subfloat[Small crossroad with 4 robots \label{SubFig:performances:smallcrossroad3w4}]
{
\includegraphics[height=\PerformancesHeight,keepaspectratio]{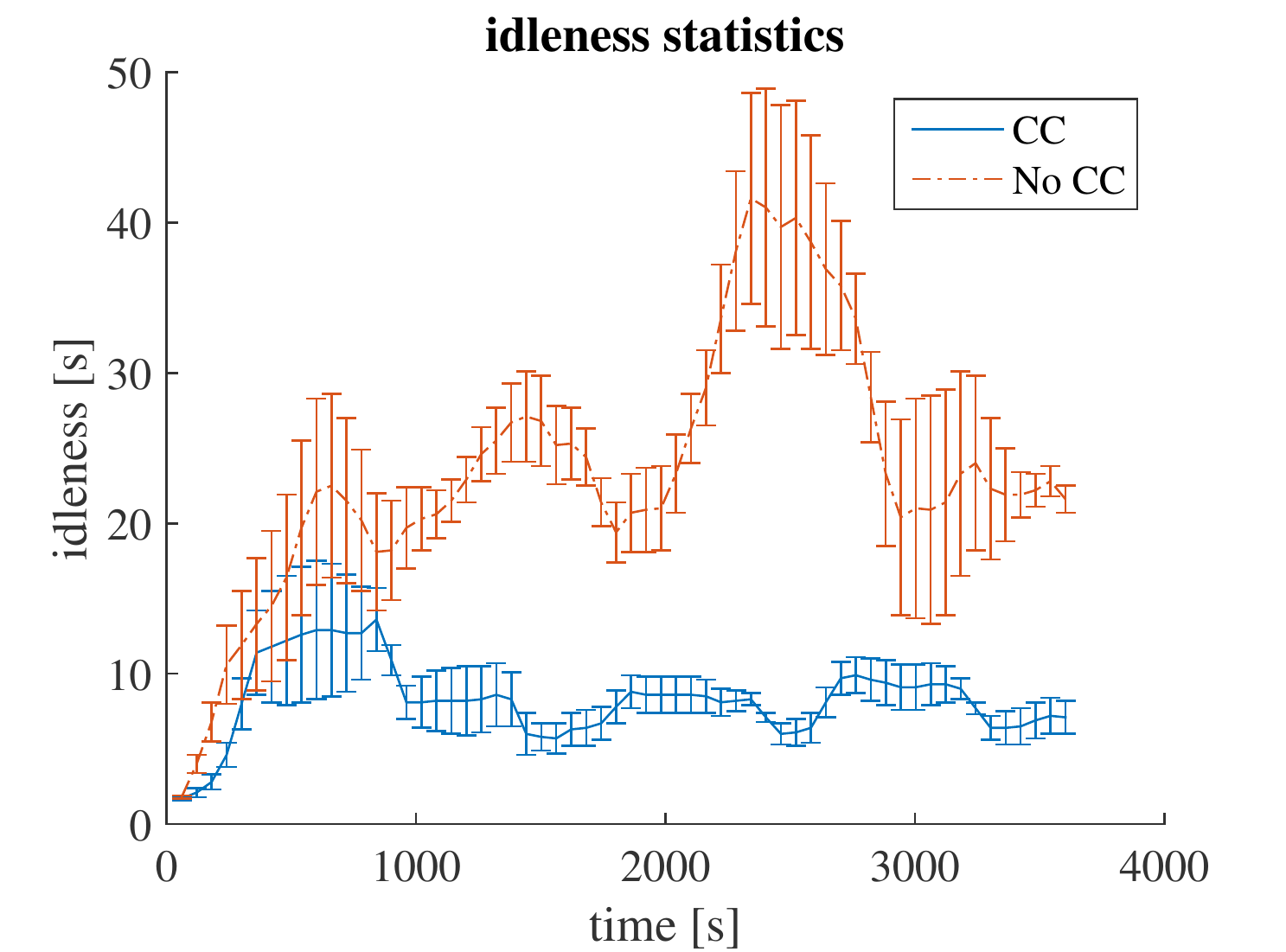}
\includegraphics[height=\PerformancesHeight,keepaspectratio]{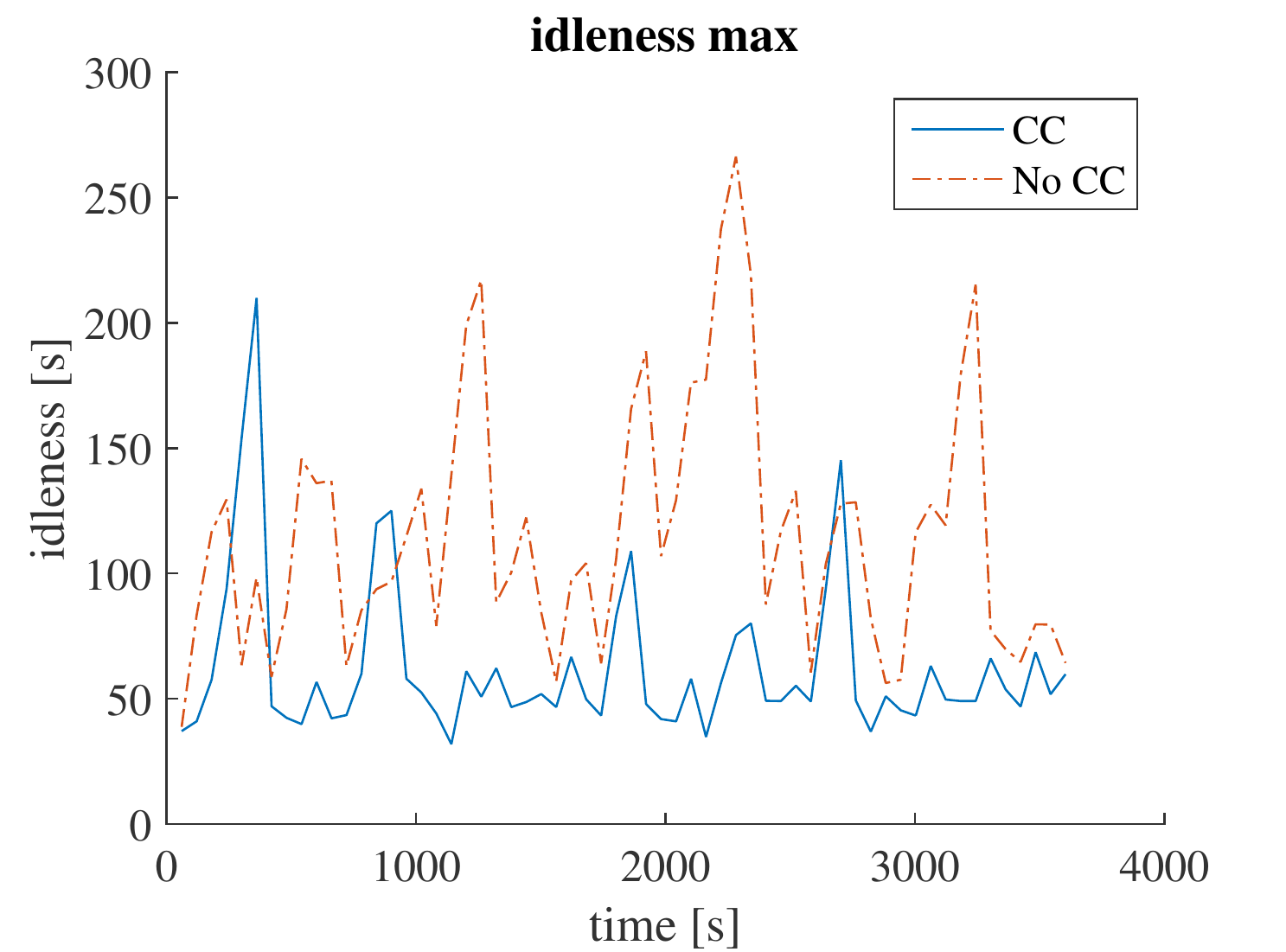}
\includegraphics[height=\PerformancesHeight,keepaspectratio]{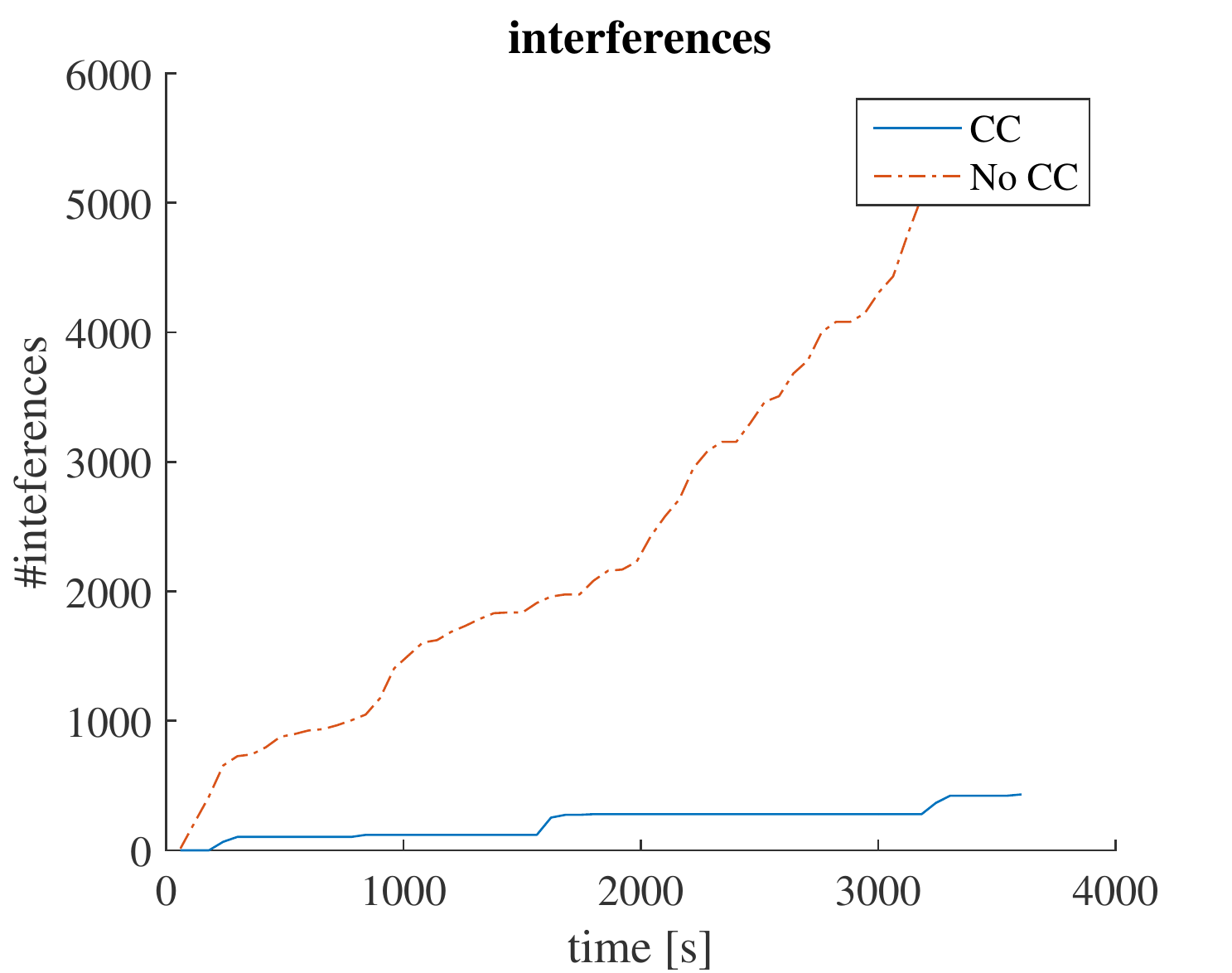}
}
\caption{Performance metrics obtained  by comparing \emph{CC} with \emph{No-CC} in the small crossroad scenarios with 3 and 4 robots. \emph{Left}: a plot of the average idleness of the graph along with its standard deviation. Statistics are computed in a moving time-window of width $600s$. \emph{Center}: the maximum idleness observed in the moving time-window. \emph{Right}: the total number of observed interferences up to the current time. \emph{CC} strategy performances are reported in blue while \emph{No-CC} performances in red.}
\label{Fig:SingleFloorPerformanceMetricsA}
\end{center}
\end{figure*}

\begin{figure*}[!t]
\begin{center}
\subfloat[Ring \label{SubFig:performances:2d_ring}]
{
\includegraphics[height=\PerformancesHeight,keepaspectratio]{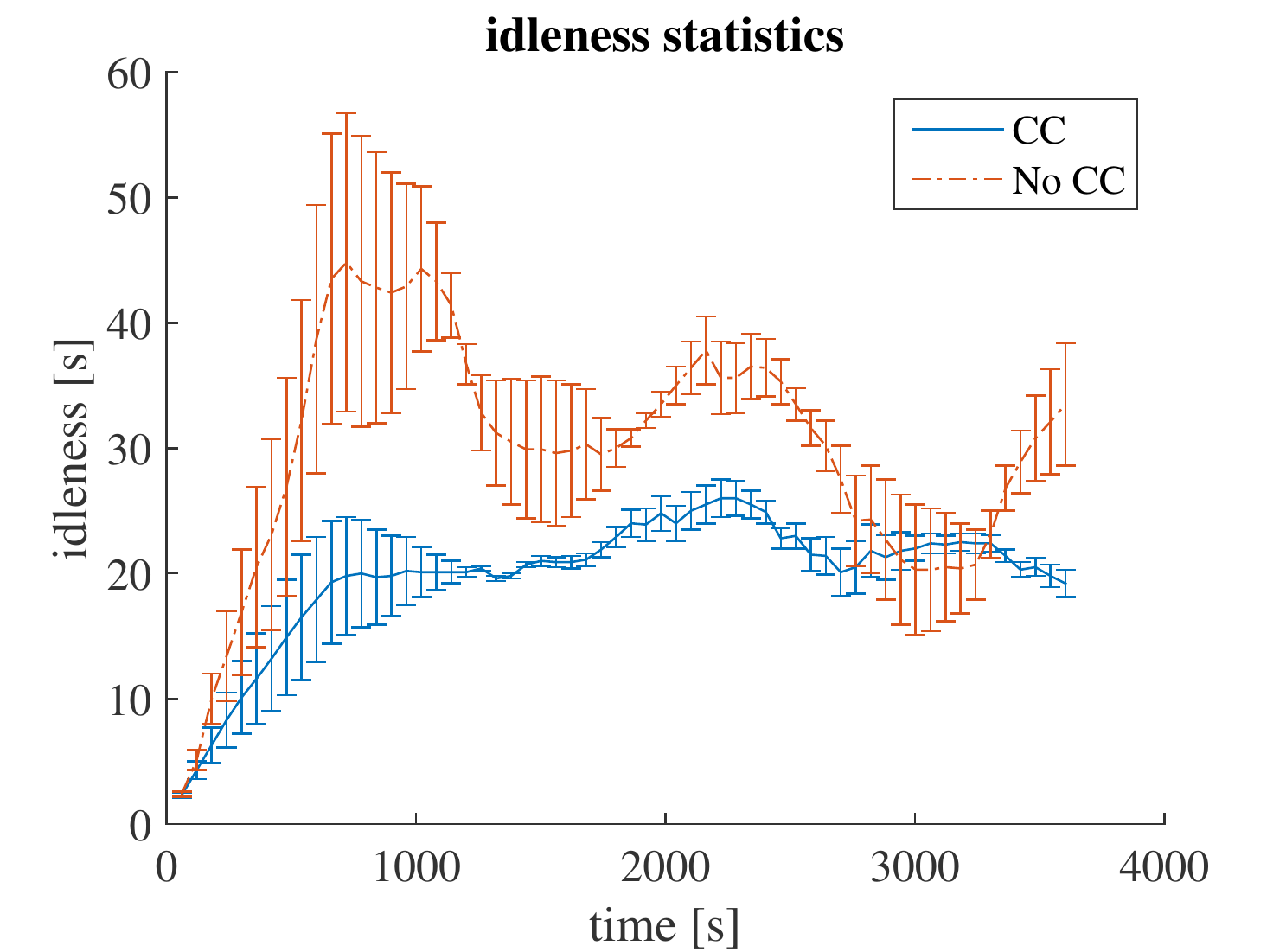}
\includegraphics[height=\PerformancesHeight,keepaspectratio]{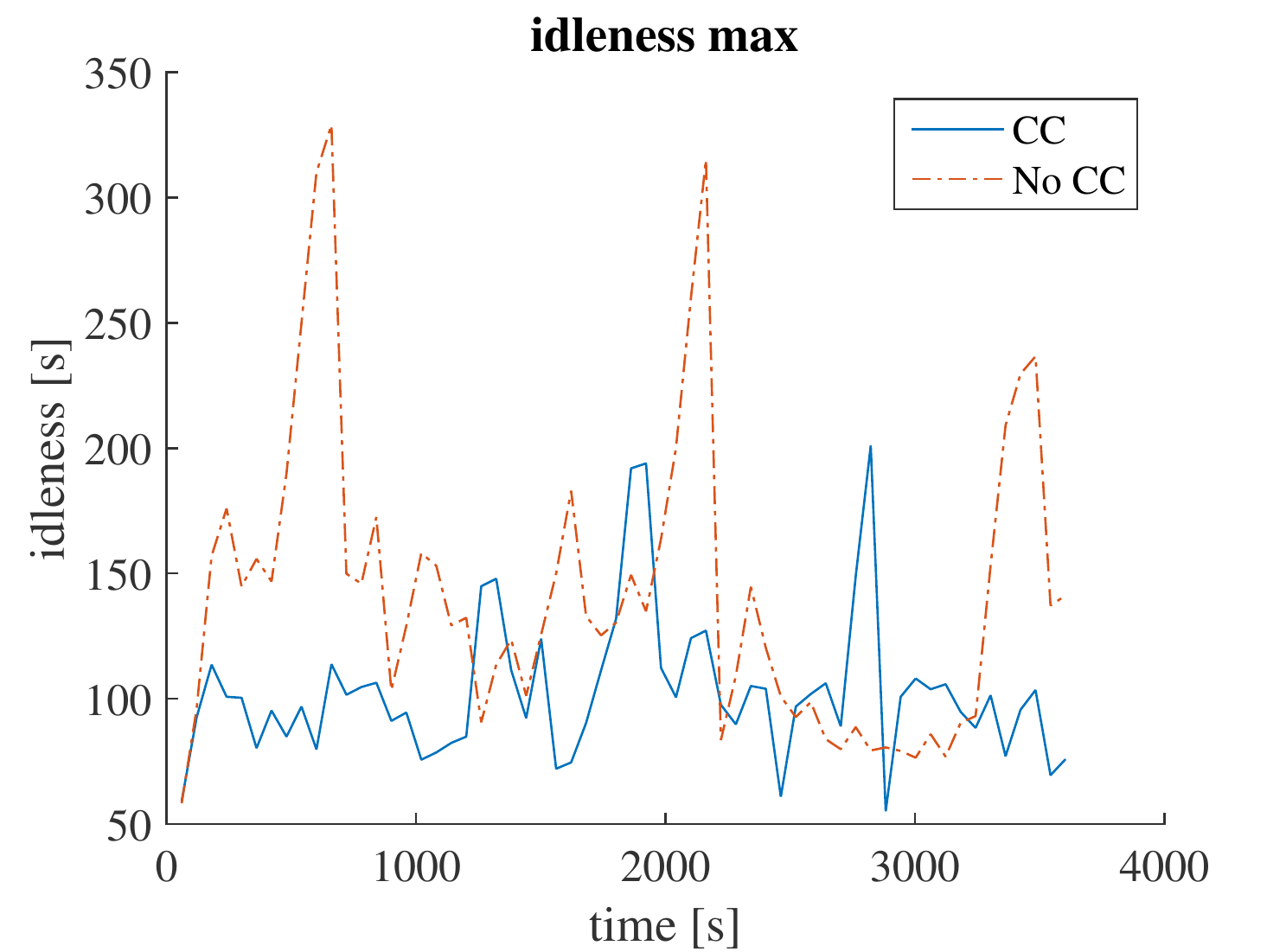}
\includegraphics[height=\PerformancesHeight,keepaspectratio]{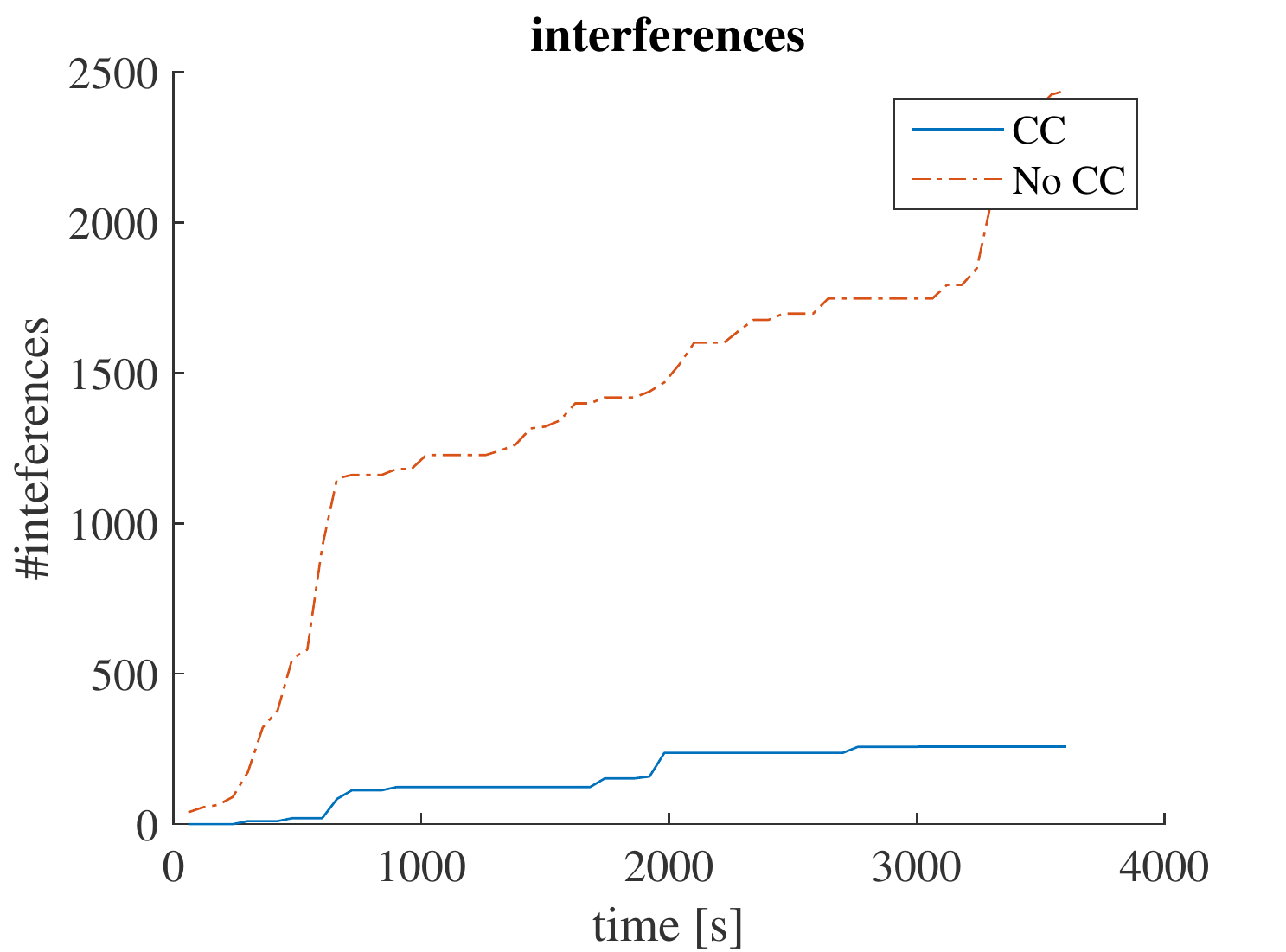}
}
\,
\subfloat[Fork \label{SubFig:performances:fork}]
{
\includegraphics[height=\PerformancesHeight,keepaspectratio]{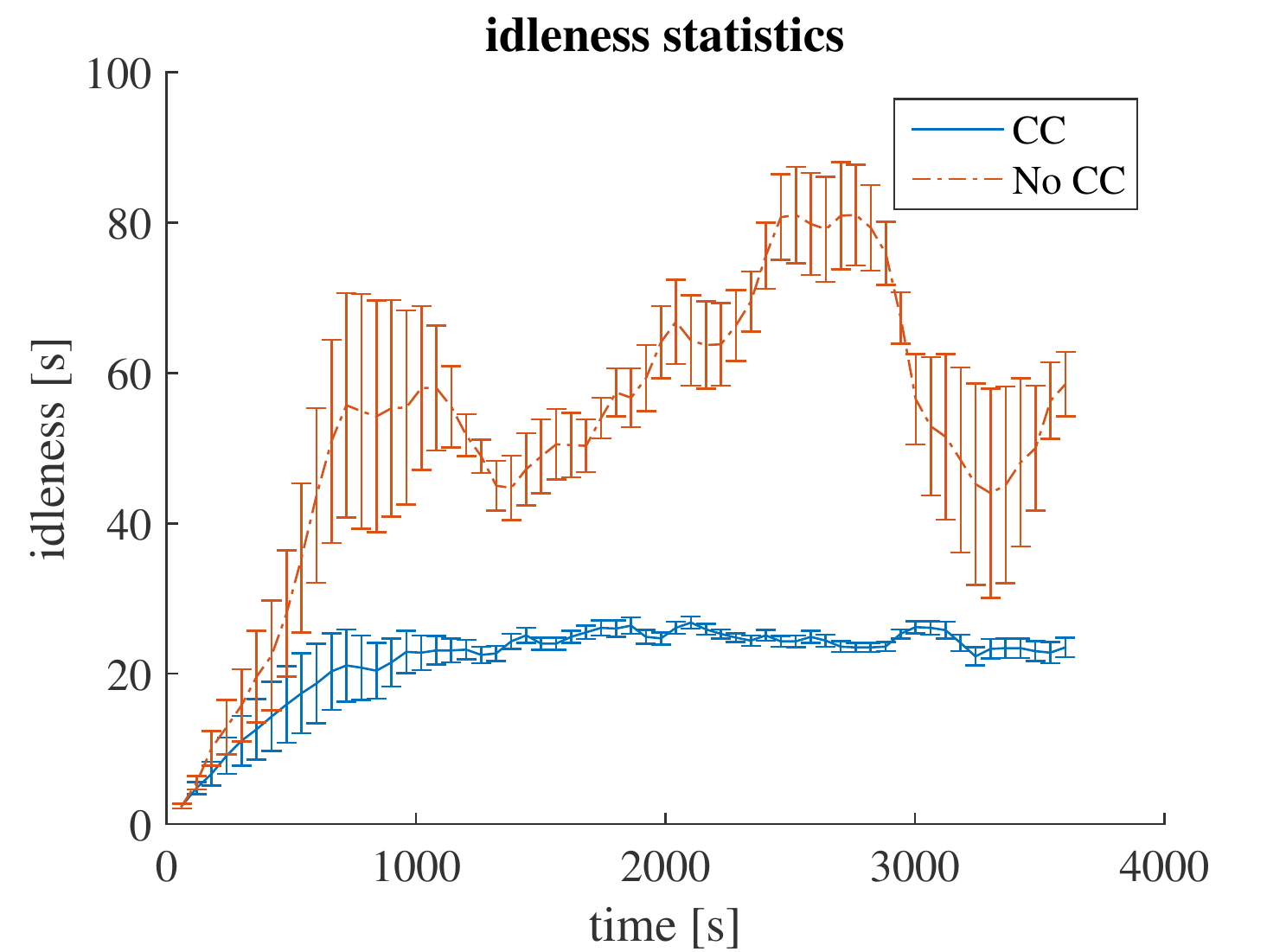}
\includegraphics[height=\PerformancesHeight,keepaspectratio]{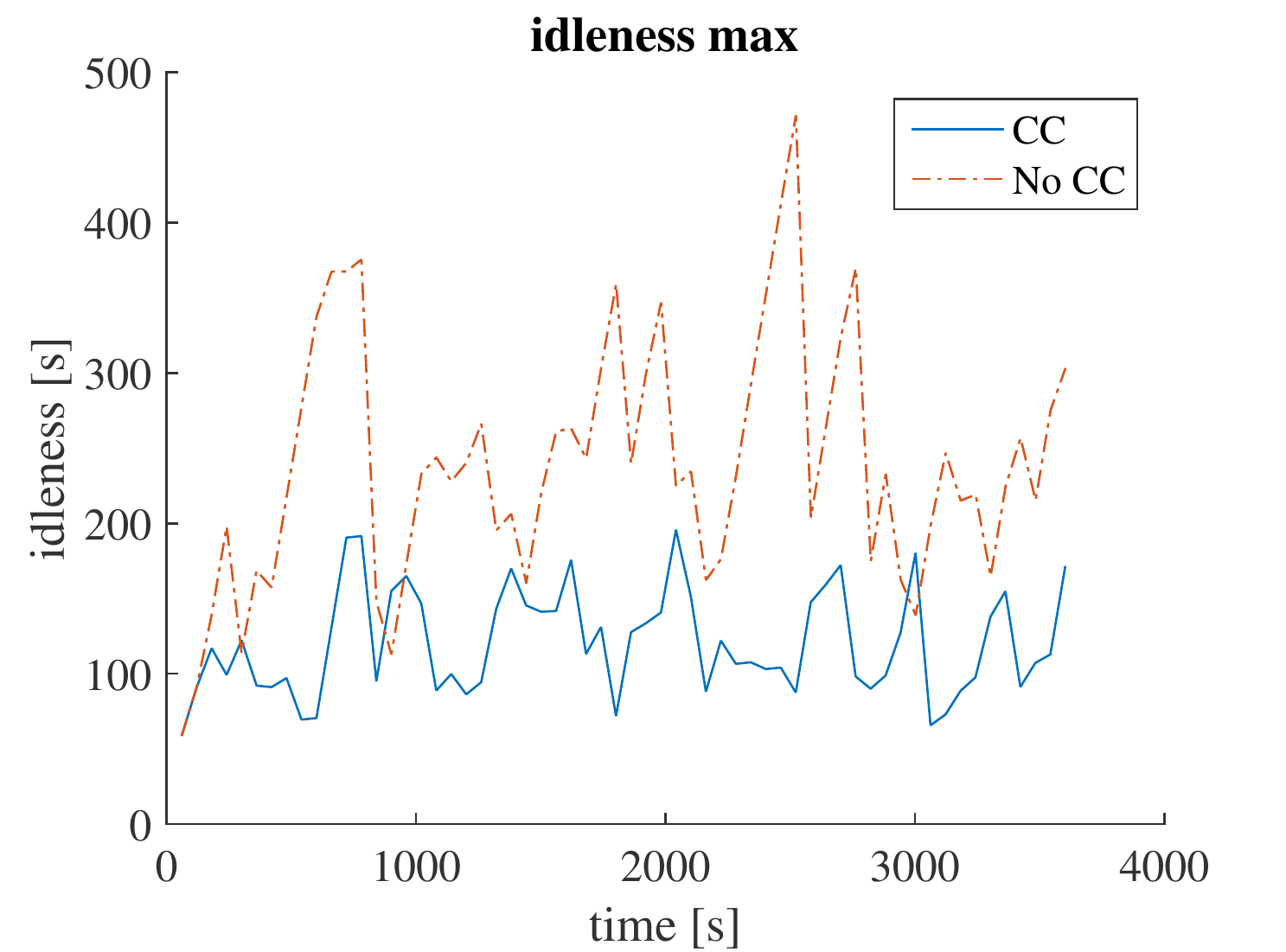}
\includegraphics[height=\PerformancesHeight,keepaspectratio]{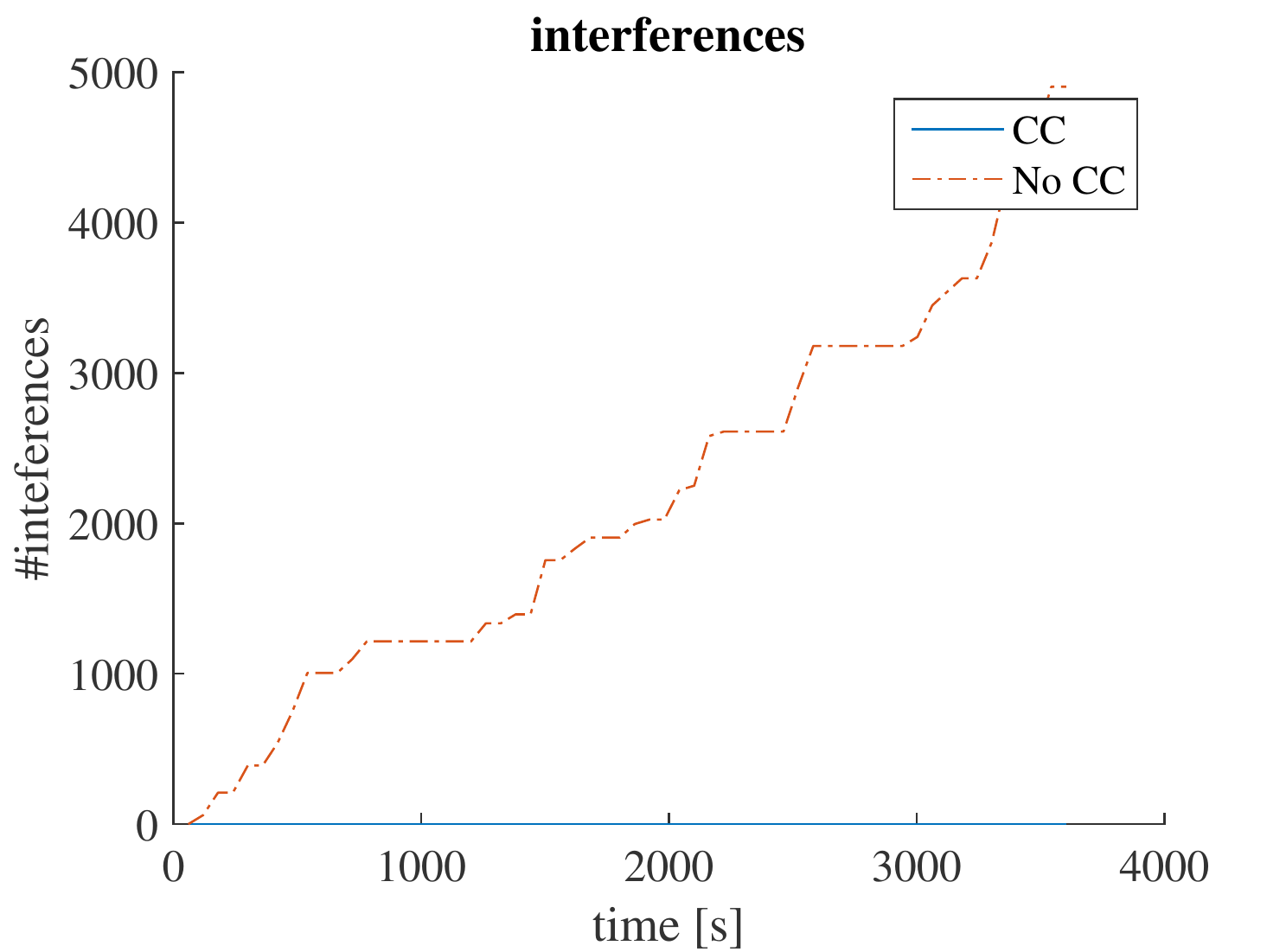}
}
\,
\subfloat[Corridor \label{SubFig:performances:longcorridor}]
{
\includegraphics[height=\PerformancesHeight,keepaspectratio]{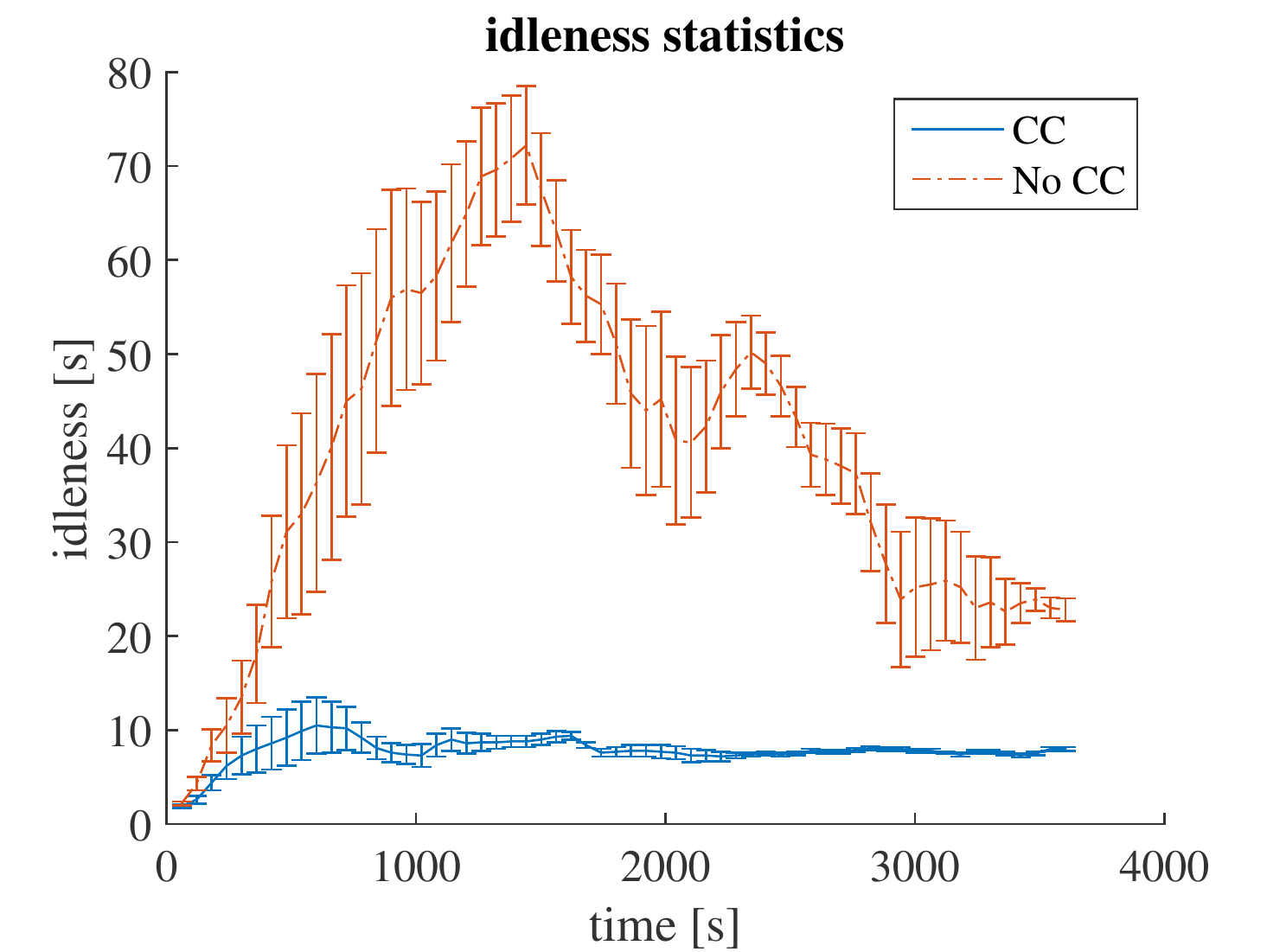}
\includegraphics[height=\PerformancesHeight,keepaspectratio]{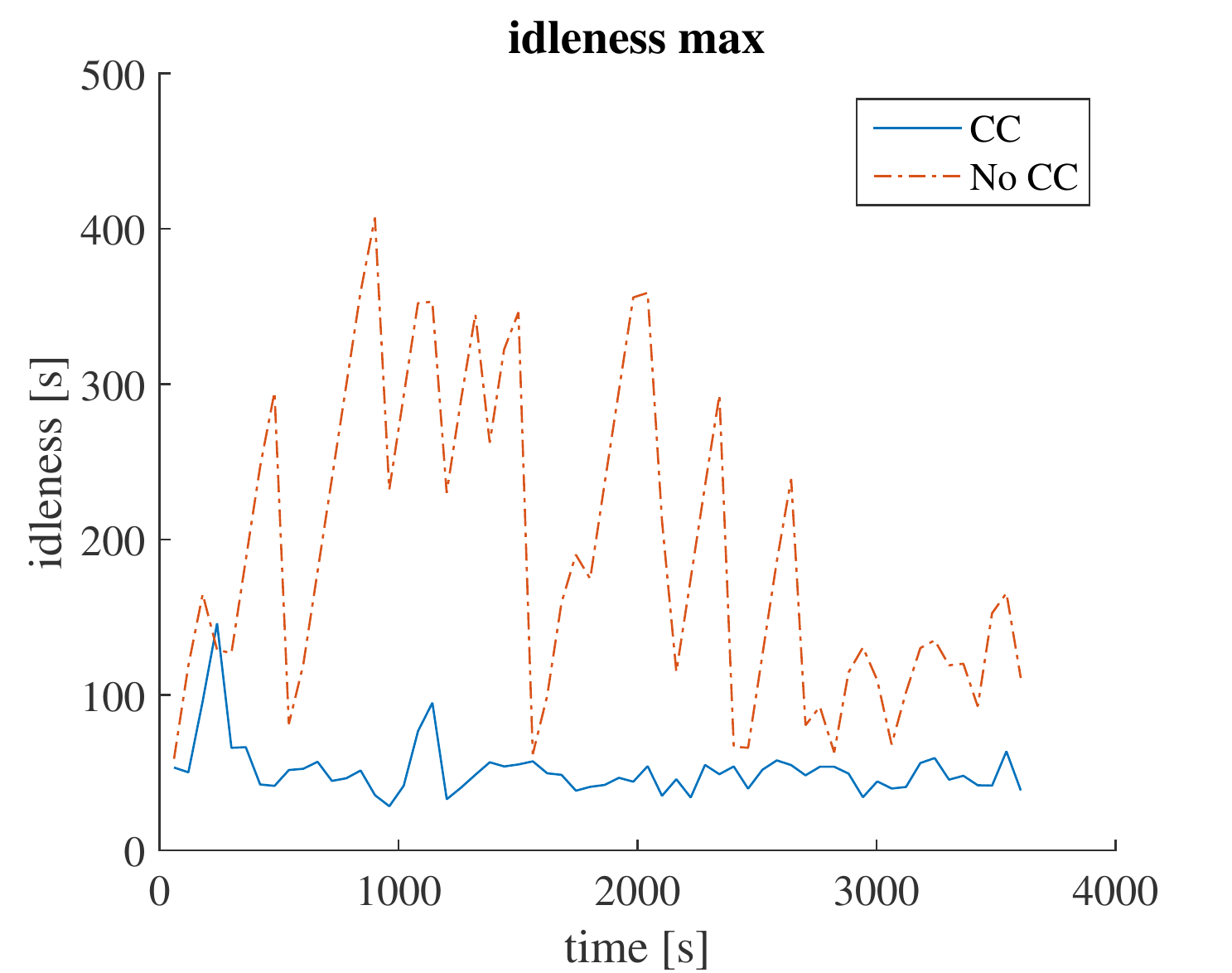}
\includegraphics[height=\PerformancesHeight,keepaspectratio]{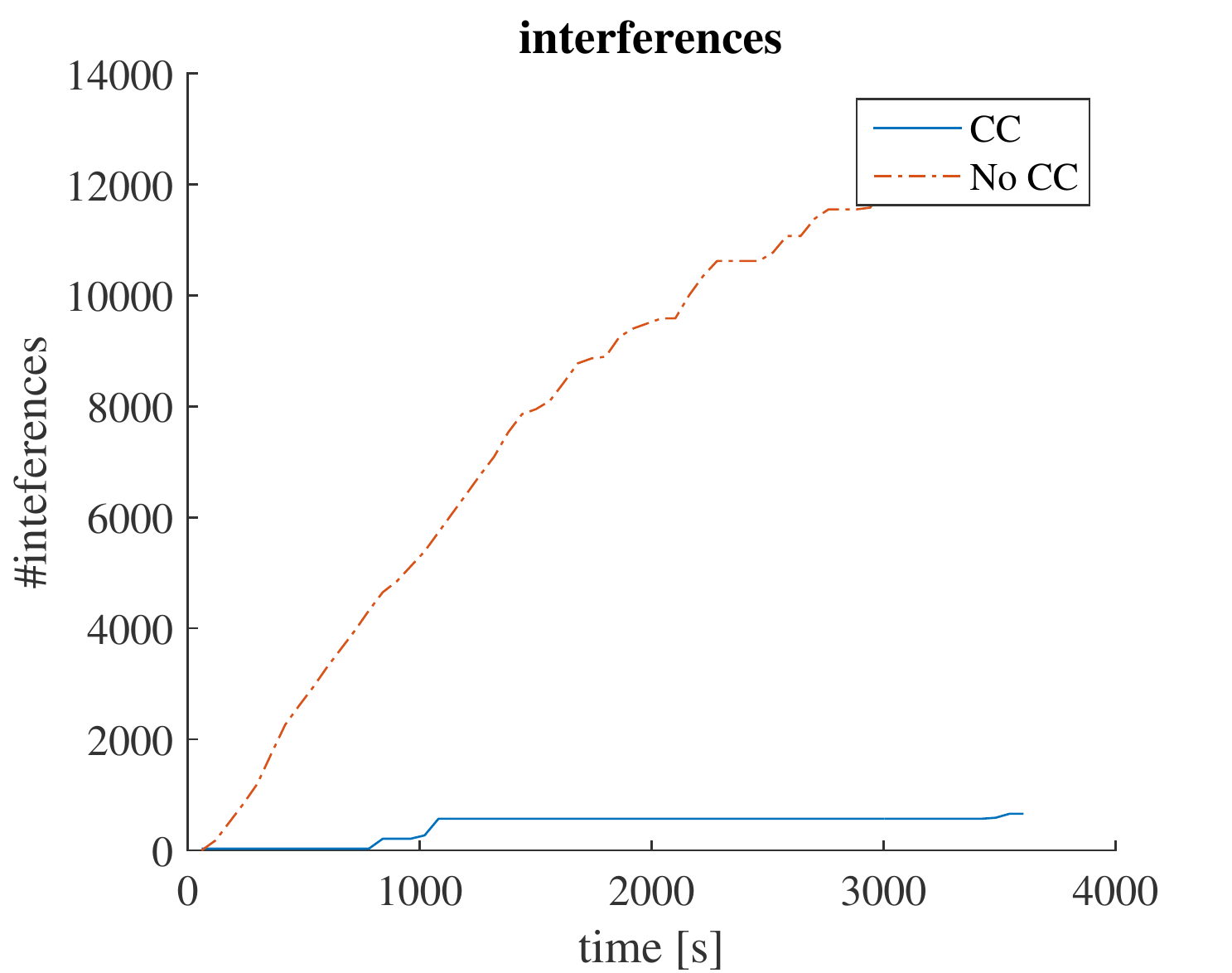}
}
\,
\subfloat[Crossroad \label{SubFig:performances:longcorridor}]
{
\includegraphics[height=\PerformancesHeight,keepaspectratio]{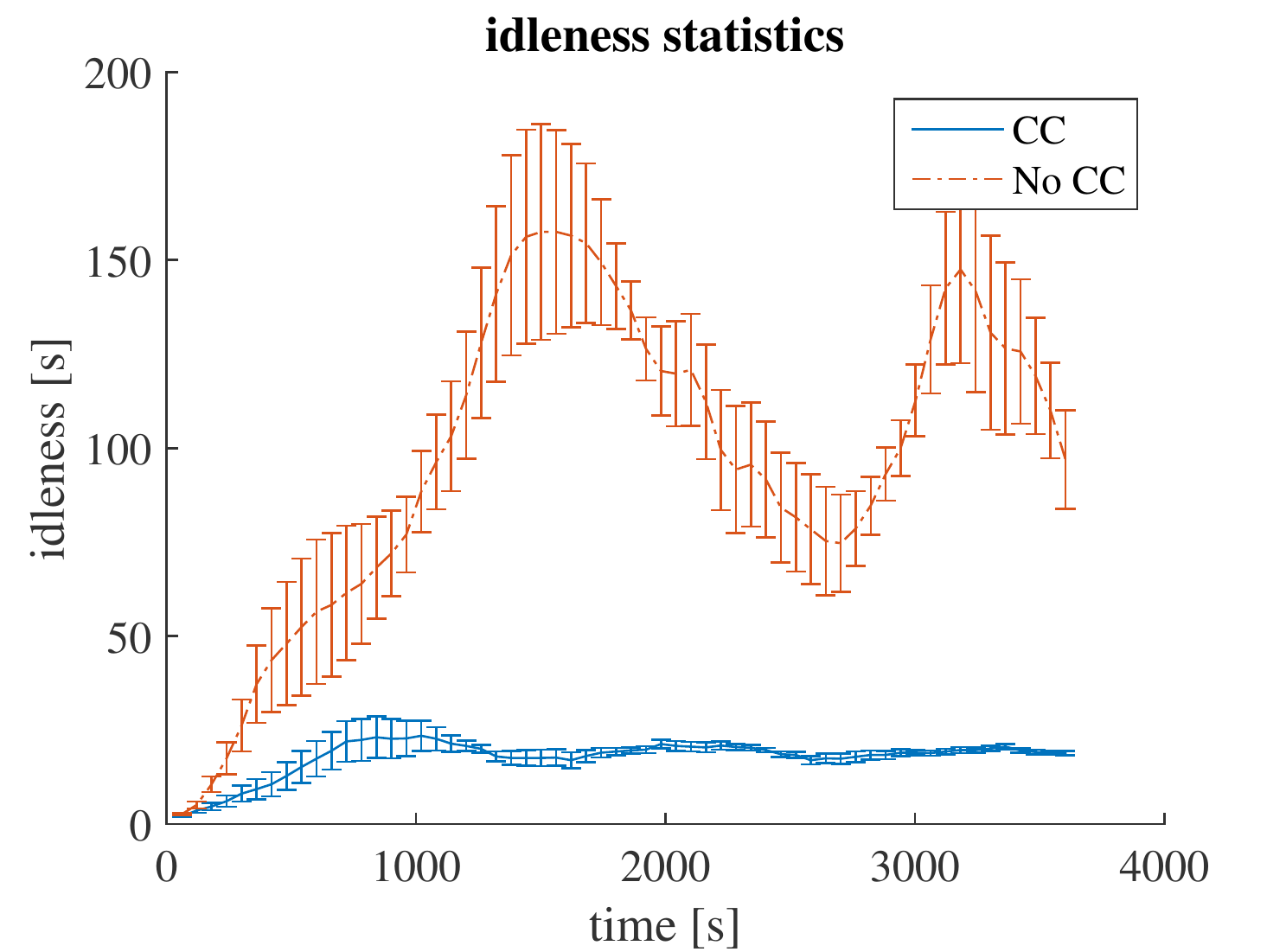}
\includegraphics[height=\PerformancesHeight,keepaspectratio]{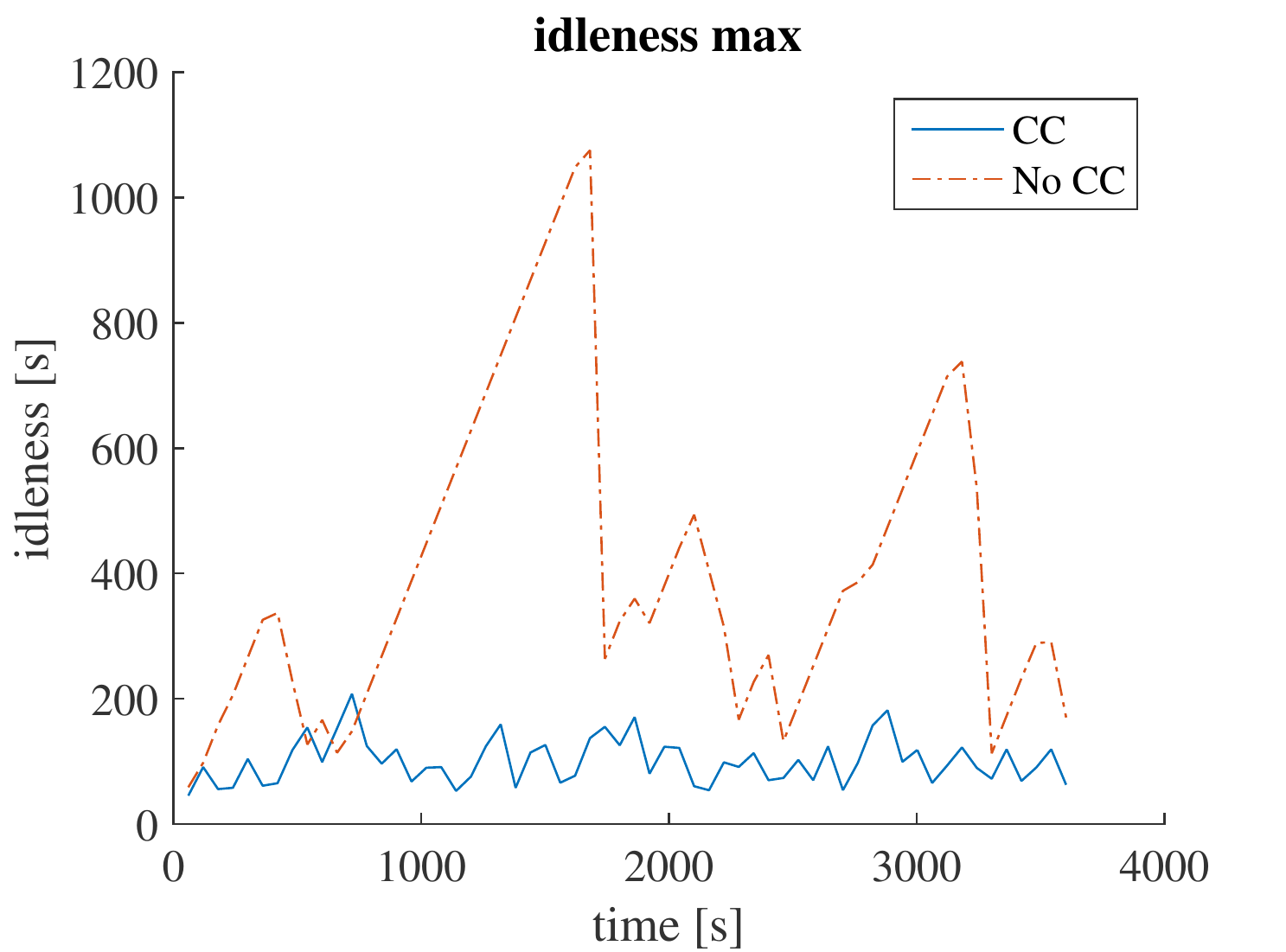}
\includegraphics[height=\PerformancesHeight,keepaspectratio]{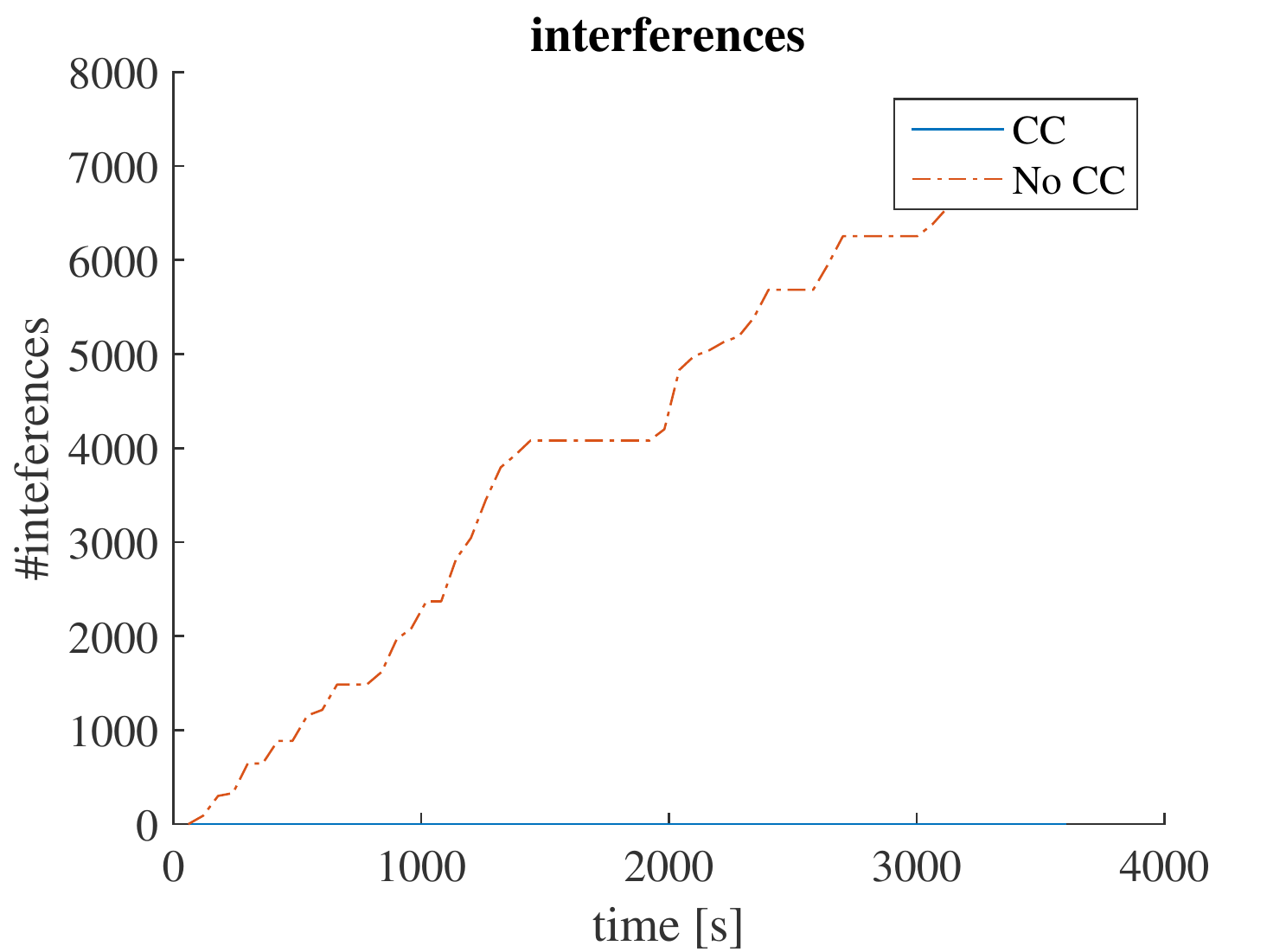}
}
\caption{Performance metrics obtained by comparing \emph{CC} with \emph{No-CC} in the single-floor scenarios ring, fork, corridor and crossroad. \emph{Left}: a plot of the average idleness of the graph along with its standard deviation. Statistics are computed in a moving time-window of width $600s$. \emph{Center}: the maximum idleness observed in the moving time-window. \emph{Right}: the total number of observed interferences up to the current time. \emph{CC} strategy performances are reported in blue while \emph{No-CC} performances in red.}
\label{Fig:SingleFloorPerformanceMetricsB}
\end{center}
\end{figure*}

For each simulated scenario, we report the results obtained with a simulation run lasting one hour. In all the runs, we used the same software deployment, i.e. we distributed ROS nodes and V-REP in the same way. 
It is worth noting that, in each scenario, we observed consistent results across simulation experiments started with different initial robot poses, as already reported in other works~\cite{Farinelli-2016}.

The obtained performance metrics are shown in Figures~\ref{Fig:CwCPerformanceMetrics}, \ref{Fig:MultiFloorPerformanceMetrics}, \ref{Fig:SingleFloorPerformanceMetricsA} and \ref{Fig:SingleFloorPerformanceMetricsB}.
In each sub-figure, we report (\emph{left}) a plot of the moving average idleness of the graph along with its standard deviation, (\emph{center}) the maximum idleness observed in the moving time-window and (\emph{right}) the total number of observed interferences up to the current time.

In particular, we compared the \emph{CwMC} and \emph{CC} strategies in the challenging scenarios three-ways (now using the patrolling graph in Fig.~\ref{SubFig:scenario:threeways:maps}) and crossroad. These simulations allow to highlight the performance improvements that can be provided by the multi-robot traversability when patrolling robots need to negotiate challenging space conflicts. 

As can be observed, the performance metrics of the \emph{CC} strategy overall present better trends in all the scenarios. 
In Fig.~\ref{Fig:CwCPerformanceMetrics}, results confirm the superiority of combining the multi-robot traversability with the path planner. In other scenarios, the comparisons between \emph{CC} and \emph{CwMC} returned small improvements or comparable idleness performances\footnote{Which we do not report here in order to reduce space.}. 
Notably, in the multi-floor scenarios, the number of interferences of \emph{CC} is constantly zero in the two-floor ring (Fig.~\ref{SubFig:performances:3d_ring}), while its value grows\footnote{This is not visible in the plot but it was observed by inspecting the recorded data.} to $90$ during the second part of the simulation in the multi-floor ramp (Fig.~\ref{SubFig:performances:3d_ramp}). %
In general, the big spikes which characterize the max idleness curves in Fig.~\ref{Fig:MultiFloorPerformanceMetrics} correspond to an inefficient team deployment over the graph or to the occurrence of challenging conflicts. In the latter case, the conflicts are constantly controlled and solved by the \emph{CC}, while they produce a big performance degradation in the case of the \emph{No-CC} strategy. Indeed, it is possible to observe a significant correlation between the maximum idleness and the average idleness which are shown in Fig.~\ref{Fig:MultiFloorPerformanceMetrics}.

Another important result can be observed on both the idleness statistics curves shown in Fig.~\ref{Fig:MultiFloorPerformanceMetrics}: The moving average idleness of the \emph{CC} is overall smaller and much less dispersed than the correspondent curve of \emph{No-CC}.  
Similar results are obtained in the case of single-floor scenarios (see Figures~\ref{Fig:SingleFloorPerformanceMetricsA} and \ref{Fig:SingleFloorPerformanceMetricsB}). We observed that the multi-floor ramp, the single-floor corridor and the crossroad are very challenging scenarios for the \emph{No-CC} strategy since the robots continuously obstruct each other while trying to reach the ends of the graph. On the other hand, the \emph{CC} strategy succeeds to avoid interference and direct negotiation of metric conflicts by mainly using node conflict management and shared idleness in order to properly redirect and redistribute robots over the graph. Clearly, in these challenging cases, all the encountered metric conflicts usually subject the engaged robot path planners to an high and useless computational load with a strong performance degradation.

It should be emphasized that no deadlocks occurred during all our simulation runs. The two-level strategy succeeded in safely governing the robot behaviour, arbitrating conflicts and suitably distributing the robots over the graph.

%==================================================================
%==================================================================

\subsection{Real-world Experiments}\label{Sect:Experiments}
%==================================================================
%==================================================================
%\input{experiments.tex}

\newcommand{\gridimageheightexp}{5.5cm}

\begin{figure*}
\begin{center}
\subfloat[DIAG ramp \label{SubFig:exp:diag_ramp}]
{
\begin{tabular}[b]{cc}
\includegraphics[height=\gridimageheightexp, keepaspectratio] {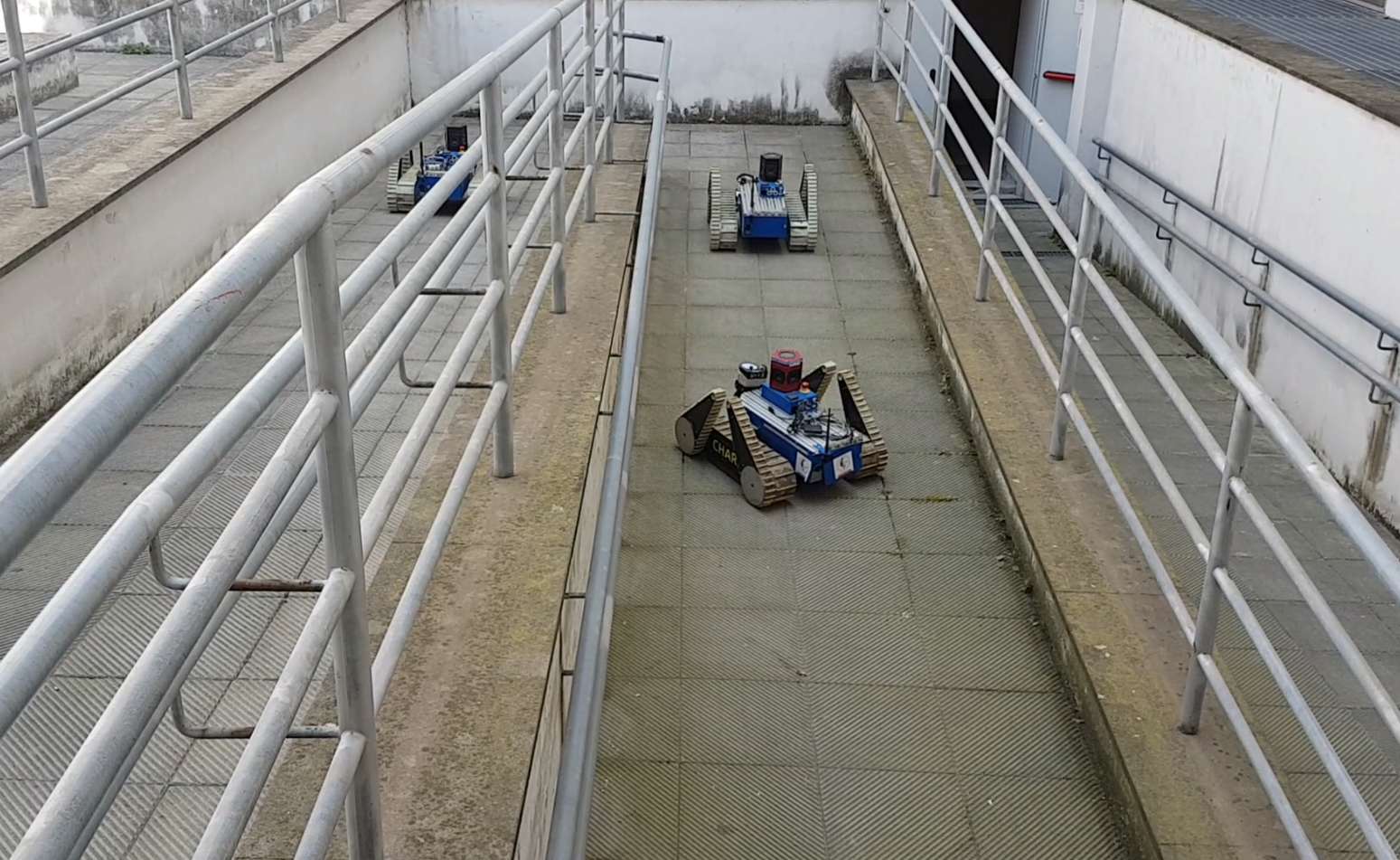}
\includegraphics[height=\gridimageheightexp, keepaspectratio]{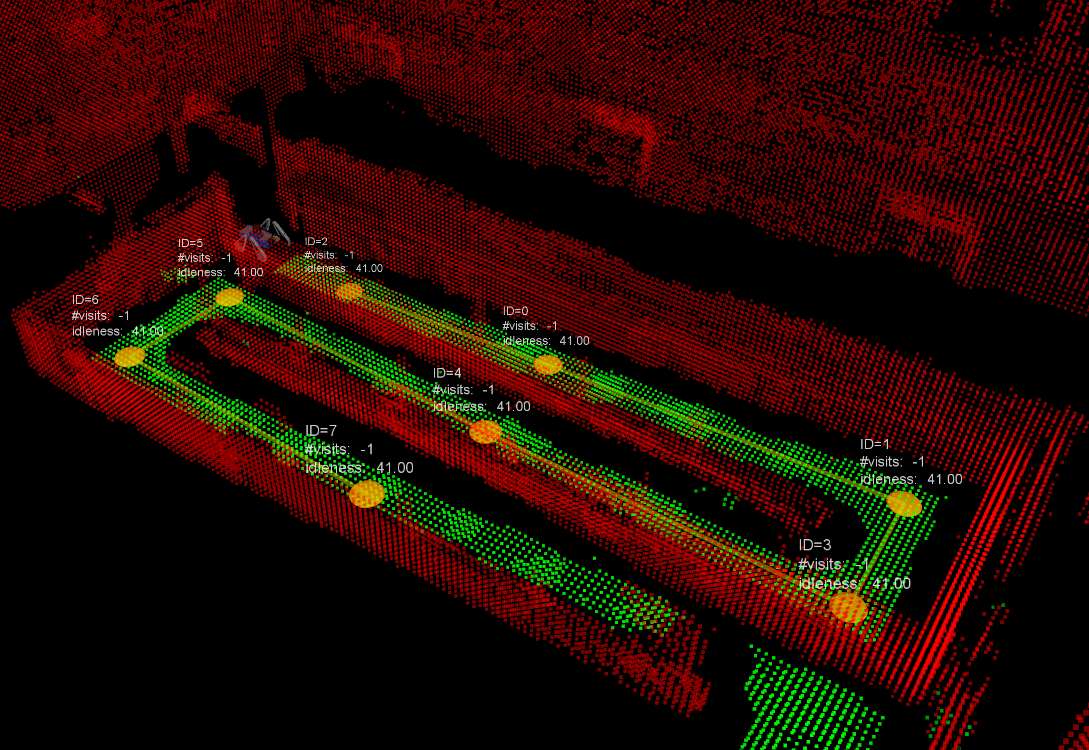}
%\\
%\includegraphics[height=\gridimageheightexp, keepaspectratio]{exp-diag-ramp-patrollin-graph}
\end{tabular}
}
\,
\subfloat[DIAG corridor \label{SubFig:exp:diag_corridor}]
{
\includegraphics[height=\gridimageheightexp, keepaspectratio]{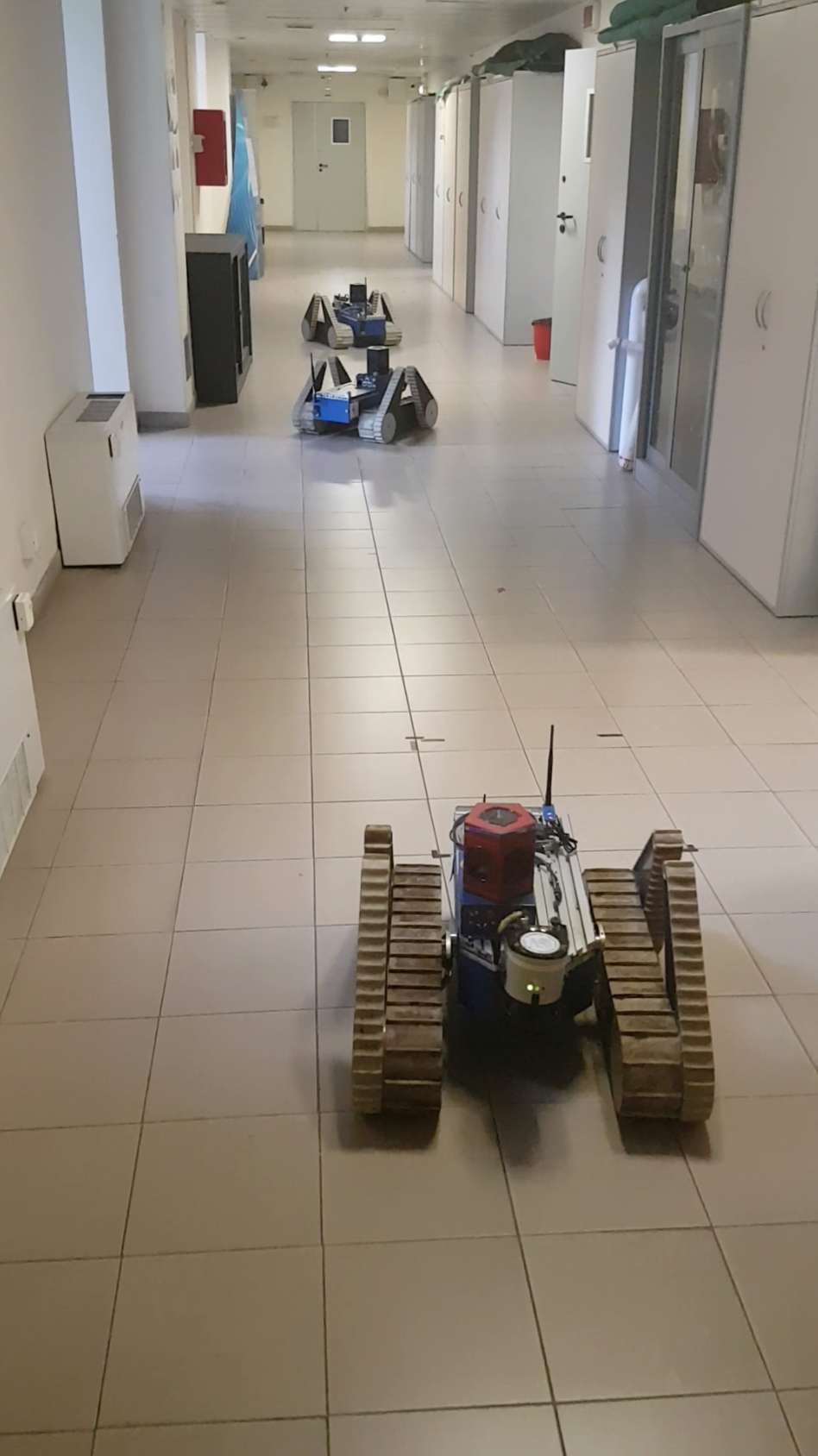}
\includegraphics[height=\gridimageheightexp,keepaspectratio]{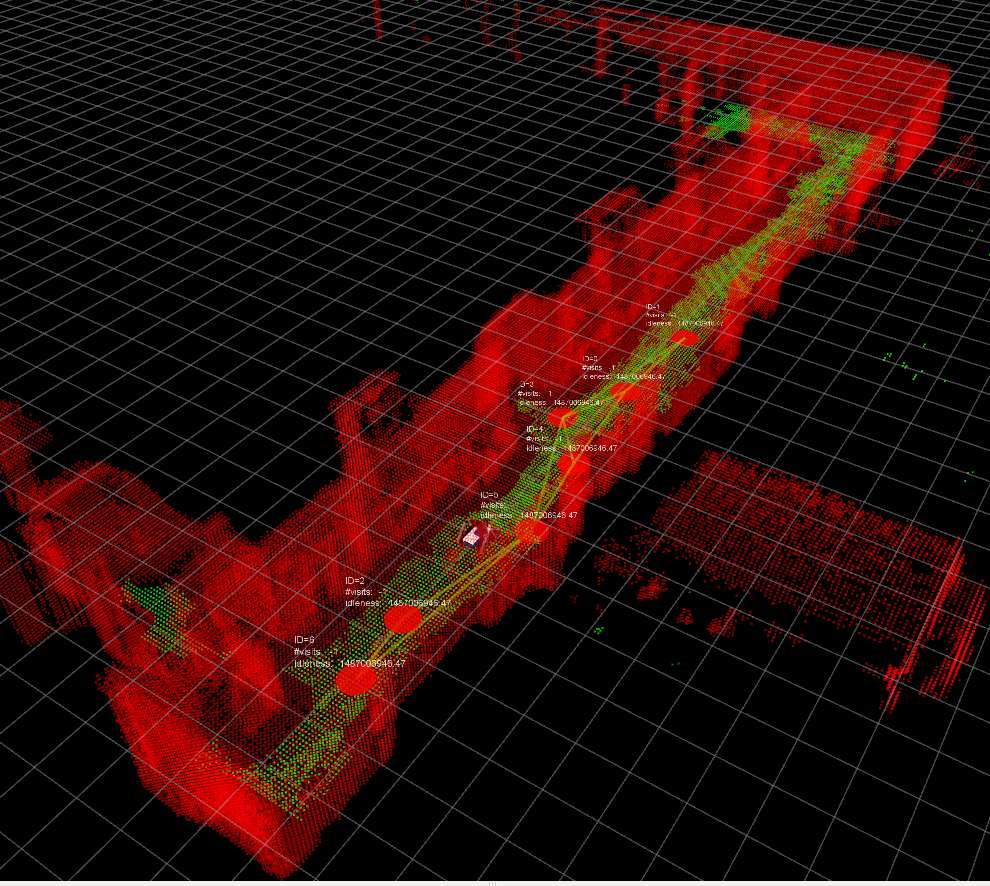}
%\hspace{-0.1cm}
\includegraphics[height=\gridimageheightexp,keepaspectratio]{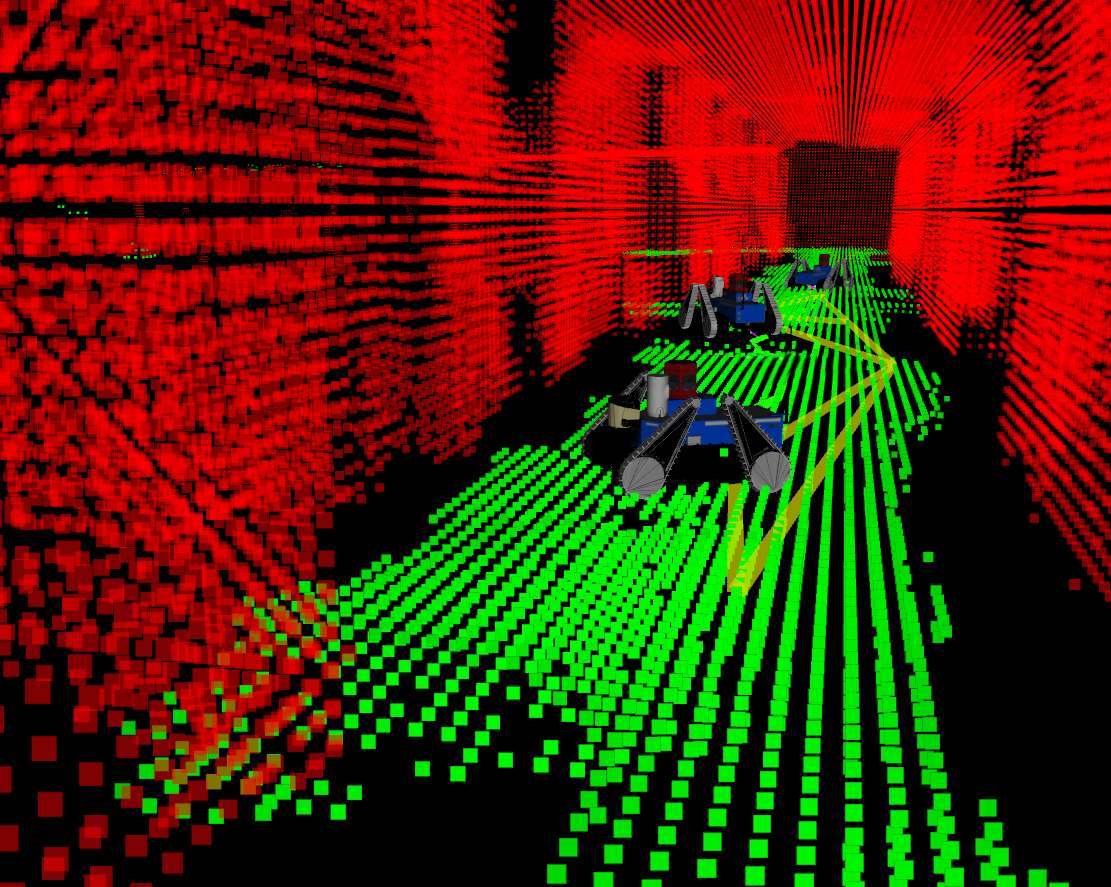}
}
%
%\subfloat[RDM Deltalinqs Training plant \label{SubFig:exp:deltalinq}]
%{
%\centering
%\begin{tabular}[b]{cc}
%\includegraphics[height=5.5cm, keepaspectratio] {deltalinqs.jpg}
%\includegraphics[height=5.5cm, keepaspectratio]{deltalinqs-patrolling2}\\
%%\includegraphics[height=4.5cm, keepaspectratio] {deltalinqs.jpg}
%%\includegraphics[height=4.5cm, keepaspectratio]{deltalinqs-patrolling2}
%\includegraphics[height=5.5cm, keepaspectratio]{deltalinqs-patrolling3}
%\end{tabular}
%} 
\caption{Two of the experimented scenarios with real UGVs.}
\label{Fig:ExperimentScenarios}
\end{center}
\end{figure*}

The real-world multi-robot system is implemented in ROS by using a multi-master architecture. In particular, \emph{nimbro\_network}~\cite{nimbro} is used for efficiently transporting ROS topics and services over a WIFI network. Indeed, \emph{nimbro\_network} allows to fully leverage UDP and TCP protocols in order to control bandwidth consumption and avoid network congestions. This capability along with a comparative testing of different ROS multi-master architectures made the TRADR consortium adopt \emph{nimbro\_network}~\cite{Kruijff-2015,worstdrDr63,worstdrDr64}.

%As already mentioned, 
We used the same C++ code in order to run both simulations and experiments (cfr. Sect.~\ref{Sect:CodeImplementation}). Only ROS launch scripts were adapted in order to specifically interface the modules with the actual multi-master nimbro\_network transport layer. 
%Moreover, we used the same parameters values both in simulations and experiments, as reported in Section~\ref{Sect:UsedParameters}. 

We performed patrolling experiments with real UGVs aiming at showing the applicability and portability of the developed software in the real 3D world. We tested our strategy with teams of two and three robots in different environments. Figure~\ref{Fig:ExperimentScenarios} shows two of the considered scenarios along with their maps and patrolling graphs. 
In particular, the \emph{CC} strategy described in Sect. \ref{Sect:NextNodeSelection} was tested on the TRADR UGVs and a satisfactory behaviour was achieved. 
Some videos of the performed experiments and further details are publicly available\footnote{ \scriptsize
\href{https://sites.google.com/a/dis.uniroma1.it/3d-cc-patrolling/}{ https://sites.google.com/a/dis.uniroma1.it/3d-cc-patrolling/}}.

Experiments confirmed that map visualization is the most demanding networking functionality of the system. This is only required on the 3D GUI of the central core if a user want to monitor patrolling actitivities (see Fig.~\ref{Fig:Architecture}). 
%As a matter of fact, published data includes full map updates, which are continuously sent to the central core (see Fig.~\ref{Fig:Architecture}). 
In this context, nimbro\_network transport layer was crucial for achieving almost smooth map data transfers. 
%Clearly, map visualization is not actually required by the distributed decision-making of the presented multi-robot strategy. 
Only the broadcast of compact coordination messages is required in order to implement the presented \emph{CC} strategy. %(see Section~\ref{Sect:PatrollingMessages}).   

During the experiments, we observed that some of the \emph{path} and \emph{selected} messages were delayed or lost. Such situations temporally provoked a patrolling performance drop, due to a locally degraded coordination. Nonetheless, nor the operation activity of the system was crucially affected, neither major congestions or deadlocks occurred. These aspects are further discussed in Sect.~\ref{Sect:Discussion}.  

It is worth noting that the windowed search strategy presented in Section~\ref{Sect:PathPlanningWindowedSearchStrategy} proved to work very well in practice. Most times, the path planner finds a path at the first attempt with the advantages of \emph{(i)} conveniently reducing the search space \footnote{In these cases, the path planner only considers the most interesting and useful part of the traversability map.} and \emph{(ii)} reducing on the average the computational load generated by the path planner.

%==================================================================
%==================================================================

%==================================================================
%==================================================================

\section{Discussion}\label{Sect:Discussion}
%==================================================================
%==================================================================
%\input{resilience_scalability.tex}

In this section, we shortly discuss the presented patrolling approach in terms of network resilience and scalability. Then, we present some lessons learnt in deploying our system in real-world scenarios.

\subsection{Network Resilience}\label{Sect:NetworkResilience}

The proposed multi-robot system is distributed and
avoids any centralized arbitration scheme, which would represent a critical point of failure. 

In the proposed strategy, the communication protocol was designed with redundant messages and an idleness synchronization scheme which support the shared knowledge representation (see Sect.~\ref{Sect:SharedKnowledge}).

In particular, at topological level, a \emph{selected} message is periodically broadcast (see Sect.~\ref{Sect:SharedKnowledge}). This redundancy adds robustness with respect to sporadic \emph{selected} message losses. In fact, if a single \emph{selected} message is lost, two robots may move towards the same node until new \emph{selected} messages arrive and allow them to resolve the node conflict. Clearly, if a significant amount of messages is lost, each robot plans its actions relying on an incomplete representation of the world state. In such case, idleness estimates are not cooperatively updated, moreover, coordination and cooperation smoothly degrade given the missed shared information and teammates decisions. 
When the network is completely down, each robot greedily performs an independent patrolling mission by avoiding teammates (see below) and solving critical path-planning failures with goal pre-emption and continuous re-planning.

At metric level, the path-planners continuously re-plan paths and correspondingly broadcast \emph{path} messages (see Table~\ref{Tab:BroadcastMessages}). In this way, each multi-robot traversability map is continuously updated. If many \emph{path} messages are lost, robots will not stop but will independently proceed towards their goals, avoiding each other thanks to the combination of the continuous re-planning with a low-level proximity checker\footnote{This laser proximity checker inhibits forward velocity commands when a close front obstacle is detected by the laser.}. It is worth noting that the metric coordination enforced by the multi-robot traversability is locally bound by the radius $R_t \geq R_c$ (see Sect.~\ref{Sect:TravCost}). This implies that a correct multi-robot traversability could be computed even if robots were only able to exchange path messages within a limited communication range $R_t$.

Additionally, even if not presented in this work, it is worth mentioning that the system can make use of the com\-mu\-ni\-ca\-tion-aware path planner presented in~\cite{caccamo2017rcamp}. This drives each robot towards better WIFI connectivity regions while planning a path towards the designated goal.

\subsection{Scalability}\label{Sect:Scalability}

In our experiments, the number of robots was limited by V-REP\footnote{In our setup, V-REP is not able to stably simulate more than four robots under realistic conditions (cfr. Sect.~\ref{Sect:Simulations}).} and the real TRADR UGVs available. %
 Nonetheless, we observed that increasing the number of robots tends to improve the patrolling performance even in challenging situations, as shown for instance by the average graph idleness curves in Fig.~\ref{Fig:SingleFloorPerformanceMetricsA}.

Additionally, we observed that, under some conditions, the robot team tends to create dynamic regions where agents patrol more often. This is a nice behaviour already observed in other works~\cite{Portugal-2016}, without recurring to an explicit space decomposition and allocation. In our case, this behaviour is induced by an explicit management of interference and conflicts (topological and metric coordination).
%, which makes the difference in narrow spaces.
%, and also partially favoured by the combination with the topological and metric characteristics of the considered 3D scenarios.

In terms of network bandwidth consumption, our approach is not demanding and could be scaled up to many robots. In fact, the data size of the messages \emph{reached}, \emph{visited}, \emph{planned}, \emph{selected}, and \emph{aborted} is very contained. On the other hand, even if \emph{path} and \emph{idleness} messages convey vector data\footnote{The \emph{path} and \emph{idleness} message sizes actually depends on the number of patrolling graph nodes.}, their broadcast frequencies are lower. In particular, \emph{path} messages are broadcast on path planning updates, which typically occur at time-varying frequencies higher than $1 Hz$. Moreover, \emph{idleness} messages are broadcast according to a pre-fixed frequency $1/T_{idln}$. If required, \emph{selected} messages could also be broadcast at a pre-fixed frequency. In this regard, the user can control such broadcast frequencies and trade-off between bandwidth consumption and system robustness.

As the number of robots grows, local high densities of robots may form. In this case, the number of coordination ``interactions" may increase in a large group of close robots facing a challenging space conflict (e.g. a narrow crossroad). 
Specifically, such robots 
%in such situations, if the Plan\-Next\-Goal($\cdot$) function does not suitably take into account team cooperation, close
may need to exchange more coordination messages in order to resolve node conflicts and converge in the negotiation of new goals. 
We already observed such challenging situations in the experimented 3D scenarios. Nonetheless, the robots always succeeded in nicely redistributing over the patrolling graph in a reasonable amount of time. In this regard, we would like to note that both metric coordination and topological coordination tend to prevent the formation of local high densities of robots.

\subsection{Lesson Learnt in Real World Deployment}\label{Sect:LessonsLearnt}

During this research and the TRADR experience~\cite{Kruijff-2015}, we learnt the following main lessons through numerous real world deployments. 

First, a robust 3D SLAM was required in order to enable multi-UGV operations in 3D dynamic environments over long-term missions. In fact, an accurate multi-robot localization is crucial to enable consistent spatially-registered cooperation and coordination. 
%Continuous robot bumps (due to wreckages or debris) may generate SLAM failures. 
%In order to improve SLAM and better support traversability analysis,
%The sole use of laser data hardly enable fast and efficient re-localizations, loop closures and mutual robot localizations. 
In some situations, we experienced that our rotating laser system was not stiff enough and driving over rough terrain resulted in noisy point clouds.
Therefore, a dense RGBD mapping could open the way to a more accurate point cloud segmentation and traversability analysis. In this regard, 
the use of a multi-modal SLAM approach which processes both RGBD and laser information could be beneficial. 

Second, a distributed knowledge representation and a robust coordination
protocol is crucial in order to attain multi-robot collaboration over unreliable network infrastructures. Our framework achieves this through redundant messages and information synchronization mechanims. In this regard, we found some of the \emph{nimbro\_network} features (e.g. forward
error correction, adaptive image compression rate and current network quality visualization) to be highly beneficial \cite{worstdrDr63,worstdrDr64}.

Third, we discovered that interferences and conflicts are very likely in disaster scenarios. In order to effectively cope with these problems, high-level decision making and low-level path planning must be tightly coupled. This is implemented in our two-level coordination strategy. In this context, metric coordination and topological coordination favour each other in a virtuous circle.  In fact, when robots ``reserve" their motion space by laying down prospective paths over the multi-robot traversability (metric coordination), teammates part away and, therefore, node conflicts are often prevented. On the other hand, when node conflicts are resolved (topological coordination), robots are redistributed over the patrolling graph and, therefore, generally pushed away from each other (preventing interferences).

%==================================================================
%==================================================================

\section{Main Characteristics of the Strategy}\label{Sect:MainCharacteristics}
%==================================================================
%==================================================================
%\input{main_characteristics.tex}

Before presenting our conclusions, we summarize the main characteristics of the presented strategy.

\noindent {\bf Coordination} (avoid conflicts):
\begin{itemize}
[noitemsep,topsep=0pt,parsep=0pt,partopsep=0pt]
\item[$\circ$] The proposed patrolling strategy is distributed.
\item[$\circ$] Interferences and conflicts are explicitly managed.
\item[$\circ$] Metric conflicts are managed by the path planner by continuously replanning over the multi-robot traversability. This mechanism implements a prioritized path planning~\cite{Lavalle-2006} which takes into account prospective robot interactions.%; ``Wait" and ``go" phases are reactively generated at need.
\item[$\circ$] Topological node conflicts are detected and resolved by the patrolling agent. 
\item[$\circ$] Metric coordination and topological coordination \linebreak favour each other in a virtuous circle (see Sect.~\ref{Sect:LessonsLearnt}). 
%\item[$\circ$]  robot priority on interference checking (based on robot ID);
%\item[$\circ$]  Task reallocation: this avoids that robots get stuck in situations where they block each other's ways while retrying to plan a path toward unreachable target positions;
\end{itemize}

%\noindent{
%\begin{compactitem} 
\noindent {\bf Cooperation} (avoid inefficient actions): a shared idleness representation supports any optimization strategy in the selection of the next node (see Sect.~\ref{Sect:SharedKnowledge}). This allows to avoid that a patrolling agent selects a goal node recently visited by a teammate.

\noindent {\bf Decision making}:
\begin{itemize}
[noitemsep,topsep=0pt,parsep=0pt,partopsep=0pt]
\item[$\circ$] Decision making relies on a tight coupling between the patrolling agent and the path planner. In particular, the patrolling agent continuously monitors the path planner and accomplishes goal pre-emption and replanning when critical conditions are detected (see Sect.~\ref{Sect:NextNodeSelection}). Additionally, path lengths computed by the path planner are used to negotiate conflicts.
%with the provision that the search set $\cal S$ must not contain the contended node in case of node conflict.
\item[$\circ$] A randomized goal selection strategy (line 2, Algorithm~\ref{Alg:SelectNexNode}) is used in order to escape from ``local minima" traps generated by critical conditions (see Sect.~\ref{Sect:NextNodeSelection}). For instance, these may be provoked by environment changes or teammates obstructions.
\item[$\circ$] Our strategy can be used as a base to develop any online patrolling solution. A wide range of user-defined strategies could be easily encoded in the best node selection (lines 5--6, Algorithm~\ref{Alg:SelectNexNode}).
%\item[$\circ$] In this work, we chose to focus on a reactive agent strategy: next best node is selected at depth 1.
\end{itemize}
%\item Communication:
%\begin{itemize}
%\itemsep0em
%\item[$\circ$] Communication is explicit by using broadcast messages.
%\end{itemize}
%\end{compactitem}
%}

\noindent {\bf Network}:
\begin{itemize}
[noitemsep,topsep=0pt,parsep=0pt,partopsep=0pt]
\item[$\circ$] Redundant messages and information synchronization mechanisms add robustness with respect to network failures (see Sect.~\ref{Sect:NetworkResilience}). 
\end{itemize}

%==================================================================
%==================================================================

\section{Conclusions}\label{Sect:Conclusions}
%==================================================================
%==================================================================
%\input{conclusions.tex}

This works presented a distributed approach for multi-robot patrolling. We focused on aspects that are typically overlooked in the literature,
such as avoiding conflicts and deadlocks in spaces shared by multiple UGVs, considering full 3D environments, traversability analysis, coordinated path planning, and real validation in 3D scenarios. 
Some of these aspects are summarized in Sect.~\ref{Sect:MainCharacteristics}. 

In particular, we developed a comprehensive framework for multi-robot patrolling dealing with all the inherent design aspects, from high-level cooperation and decision making, to low-level coordination and path planning. We improved upon the state-of-the-art methods by developing a two-level coordination strategy, which crucially takes into account the necessary tight coupling between topological and metric decision making. In this regard, both topological and metric coordination allow to explicitly minimize interference and conflicts, which crucially affect UGVs activity. We experienced that this approach allows to effectively cope with the typical challenges involved when a team of UGVs is deployed in a disaster scenario. 
%navigation over 3D terrains, dynamic environments, spatial conflicts, unreliable communication networks and long-term operations.

%
The presented two-way coordination strategy is general and
%on topological and metric level 
can be used as a base to develop new strategies for optimizing the patrolling graph \emph{idleness} 
%of nodes in a patrolling graph 
and ensuring space conflicts negotiation.
%collision-free operation by applying a staged avoidance scheme on several levels of severity.
%

Our multi-robot patrolling algorithm is fully integrated with a 3D SLAM algorithm, traversability analysis and coordinated path planning. This enables our system of ground robots to operate in 3D. 

We demonstrate competitive performance in both simulation and real world experiments, enabling robots to simultaneously operate in realistic simulation and in real world experiments.
The obtained results show that the \emph{Coordination plus Cooperation} strategy was superior than our baseline throughout all performance measures, i.e., mean idleness, max idleness, spread of idleness and inference events. Notably, when using the \emph{CC} strategy, no deadlocks were observed during our experiments and the number of interferences was always significantly reduced (or zeroed in some cases). Moreover, we observed that the multi-robot traversability is able to improve the patrolling team behaviour in the most challenging scenarios, where space conflicts crucially affect robot activities.
As discussed in Sect.~\ref{Sect:Scalability}, our approach offers good scalability properties both in terms of network bandwidth consumption and performance (the latter to be further validated with larger robot fleets). 

%We will publish the source code of the presented approach upon this paper acceptance, with the aim of providing a useful tool for researchers in the Robotics Community.
We publish the source code of the presented approach  with the aim of providing a useful tool for researchers in the Robotics Community.

In the future, we plan to increase the number of robots simultaneously operating in real world experiments\footnote{Recurring to simpler and more affordable robotic platforms is required.}.
Furthermore, we wish to investigate on patrolling prioritization with heterogeneous robot fleets. In this context, exploration of ``unknown" environments given a topological prior (i.e., a topological map used as a patrolling graph) seems a promising research direction.
Furthermore, it would be beneficial to integrate explicit dynamic updates of the patrolling graph.
Finally, integration of a multi-robot SLAM algorithm, enabling map-sharing and map-persistence over the whole operation is a promising avenue for scaling the real-world operation to larger areas.

%==================================================================
%==================================================================

\begin{acknowledgements}
This work was supported by the European Union's Seventh Framework Programme for research, technological development and demonstration under the \linebreak TRADR project No. FP7-ICT-609763.
\end{acknowledgements}

\section{Appendix}

%==================================================================
%==================================================================
%\input{implementation_details.tex}

\subsection{Code Implementation}\label{Sect:CodeImplementation}

For the implementation of the patrolling agent algorithm, we used the C++ ROS package \emph{patrolling\_sim} as a starting point~\cite{patrollingsim,Portugal-2016}. This is specifically designed for 2D patrolling tasks. It was used as a starting skeleton architecture providing core functionalities (such as graph management utilities). We significantly modified the core of this package in order to manage 3D data, implement our new patrolling agent algorithm, interface the agent module more tightly with the path planner and the 3D GUI in our network architecture.  

%An open source implementation of our framework will be available upon acceptance\footnote{\href{https://gitlab.com/luigifreda/3dpatrolling}{ https://gitlab.com/luigifreda/3dpatrolling}.}. 
An open source implementation of our framework is available\footnote{\href{https://gitlab.com/luigifreda/3dpatrolling}{ https://gitlab.com/luigifreda/3dpatrolling}.}.

\subsection{Software Design}\label{Sect:FunctionalArchitecture}

A functional diagram of the presented multi-robot system is reported in Fig.~\ref{Fig:Architecture}. The main blocks are listed below.

\noindent {\bf The robots}, each one with its own ID~$\in \{1,...,m\}$, have the same internal architecture and host the on-board functionalities which concern decision and processing aspects both at topological level and  at metric level. 
According to Sect.~\ref{Sect:SharedKnowledge}, each robot maintains and updates an instance of the patrolling graph and of the metric map in its internal memory.  

\noindent {\bf The core services}, hosted in the main central computer,
 manage the multi-robot system persistence data\-base and allow specific modules to load/save map, trajectories and patrolling graphs from/into the central database (for re-using relevant data along different missions). 

\noindent {\bf The core modules}, also hosted in central computer, include the patrolling graph builder and the patrolling monitor. The first builds a patrolling graph from a user assigned set of waypoints or from a saved history of robot trajectories. %The latter continuously checks the current status of the patrolling activities and of records relevant data.
The built patrolling graph is then distributed to all the robots and saved in the central persistence database. The patrolling monitor continuously checks the current status of the patrolling activities and records relevant data for monitoring and benchmarking.

\noindent {\bf The multi-robot 3D GUI}, hosted on one OCU (Operator Control Unit), is based on RVIZ and allows the user \emph{(i)} to select multiple waypoints which can be fed to the path planners or to the patrolling graph builder \emph{(ii)} to visualize relevant point cloud data, maps, and robot models \emph{(iii)} to stop/restart robots when needed \emph{(iv)} to trigger the loading/saving of maps and robot trajectories \emph{(v)} to realign the current map of a selected robot to a loaded map.

The architecture is fully distributed without centralized coordination mechanisms. In particular, each robot hosts an instance of the patrolling agent and of the path-planner.

As shown in Fig.~\ref{Fig:Architecture}, the various modules in the architecture exchange different kind of messages. These are grouped in the following types.
\begin{itemize}
[noitemsep,topsep=0pt,parsep=0pt,partopsep=0pt]
\item \emph{Coordination messages}: these are exchanged amongst robots for sharing knowledge and decisions, in order to attain cooperation and coordination. For convenience, the patrol monitor records an history of these messages. 
\item \emph{GUI messages}: these are exchanged with the 3D GUI and include both control messages and visualization data.
\item \emph{Load/save messages}: these are exchanged with the core services and contain both loaded and saved data.
%Clearly, each coordination message includes the emitting robot ID in its header.
\end{itemize}

%==================================================================
%==================================================================

\bibliographystyle{spmpsci}      % mathematics and physical sciences
%\bibliography{bibliography}   % name your BibTeX data base

%==================================================================
%==================================================================
% for the bbl 

%==================================================================
%==================================================================

%\begin{thebibliography}{}
%\end{thebibliography}

\begin{acronym}
\acro{TRADR}{``Long-Term Human-Robot Teaming for Robots Assisted Disaster Response''}
\acro{TJEx}{TRADR Joint Exercise}
\acro{ICP}{Iterative Closest Point}
\acro{OCU}{Operational Control Unit}
\acro{MAP}{Maximum A Posteriori}
\acro{SLAM}{Simultaneous Localization and Mapping}
\acro{UGV}{Unmanned Ground Vehicle}
\acro{SaR}{Search and Rescue}
\acro{AI}{Artificial Intelligence}
\acro{POIs}{Points Of Interest}
\end{acronym}

%\newpage
%\tableofcontents

\end{document}